\documentclass[10pt,twocolumn,letterpaper]{article}

\usepackage[pagenumbers]{cvpr}

\definecolor{cvprblue}{rgb}{0.21,0.49,0.74}
\usepackage[T1]{fontenc} %
\usepackage{booktabs}
\usepackage{xspace}
\usepackage{multirow}
\usepackage{pifont} %
\usepackage{array}
\usepackage{tikz}
\usetikzlibrary{arrows.meta,positioning,fit,calc,shapes.geometric,shapes.misc,backgrounds}
\newcommand{\exFilm}{Wildlife\xspace}    %
\newcommand{\exN}{143}                   %
\newcommand{\exE}{77}                    %
\newcommand{\exL}{112}                   %
\newcommand{\exGap}{35}                  %
\newcommand{\exShots}{243}               %
\newcommand{\exIntro}{11}                %
\newcommand{\exW}{100}                   %
\newcommand{\exNotches}{13,23,26,28,33,35,39,44,45,74,75,91,92,100}
\newcommand{\exFactE}{chevy\_wagon.at = warren\_miller\_house}
\newcommand{\exFactL}{chevy\_wagon.at = brinson\_yard}
\newcommand{\exKept}{26}

\newcommand{\exStory}{In \emph{Wildlife}, the screenplay names the family's car, a 1956 Chevrolet wagon, only in scenes~11 and~13. In scene~77 Jeanette pulls ``the car'' up outside Warren Miller's house; in scene~112 she gets into ``her car'' and drives away. A film must show the same wagon both times.\xspace}

\newcommand{\method}{Continuo\xspace}              %
\newcommand{\ledger}{story ledger\xspace}          %

\newcommand{\subgraph}{dependency sub-graph\xspace} %
\newcommand{\reqeval}{request-space evaluation\xspace} %
\newcommand{\Reqeval}{Request-space evaluation\xspace}
\newcommand{\spe}{SPE-Scenes\xspace}  %
\newcommand{\speall}{SPE\xspace}         %
\newcommand{\spelong}{Scene-Pair Eval\xspace}
\newcommand{\planaudit}{plan audit\xspace}         %
\newcommand{\Planaudit}{Plan audit\xspace}

\newcommand{\ceiling}{control ceiling\xspace}      %
\newcommand{\Ceiling}{Control ceiling\xspace}
\newcommand{\speobj}{SPE-Objects\xspace}            %
\makeatletter
\newcommand{\Rraw}[1]{\@ifundefined{res@#1}{\PackageError{preamble}{Result key `#1' is not defined in numbers.tex}{}}{\expandafter\expandafter\expandafter\R@sign\csname res@#1\endcsname\relax\@nil}}
\def\R@sign#1#2\@nil{\ifx-#1\ensuremath{-}#2\else#1#2\fi}%
\newcommand{\R}[1]{\ifmmode\textnormal{\Rraw{#1}}\else\Rraw{#1}\fi}%
\newcommand{\setR}[2]{\expandafter\def\csname res@#1\endcsname{#2}}
\makeatother
\setR{agree.spe.kappa}{.87}
\setR{agree.spe.noopus.pct}{95}
\setR{agree.spe.opus-argon.kappa}{.85}
\setR{agree.spe.opus-argon.pct}{89.4}
\setR{agree.spe.opus-sol.kappa}{.87}
\setR{agree.spe.opus-sol.pct}{90.5}
\setR{agree.spe.opus-sonnet.kappa}{.81}
\setR{agree.spe.opus-sonnet.pct}{86.2}
\setR{agree.spe.pct}{91}
\setR{agree.spe.rank}{4}
\setR{agree.spe.sol-argon.kappa}{.88}
\setR{agree.spe.sol-argon.pct}{91.2}
\setR{agree.spe.sonnet-argon.kappa}{.80}
\setR{agree.spe.sonnet-argon.pct}{85.7}
\setR{agree.spe.sonnet-sol.kappa}{.80}
\setR{agree.spe.sonnet-sol.pct}{85.6}
\setR{agree.spe.tau}{.90}
\setR{agree.spo.kappa}{.75}
\setR{agree.spo.noopus.pct}{91}
\setR{agree.spo.opus-argon.kappa}{.72}
\setR{agree.spo.opus-argon.pct}{82.0}
\setR{agree.spo.opus-sol.kappa}{.63}
\setR{agree.spo.opus-sol.pct}{76.1}
\setR{agree.spo.opus-sonnet.kappa}{.71}
\setR{agree.spo.opus-sonnet.pct}{81.0}
\setR{agree.spo.pct}{84}
\setR{agree.spo.rank}{4}
\setR{agree.spo.sol-argon.kappa}{.69}
\setR{agree.spo.sol-argon.pct}{80.7}
\setR{agree.spo.sonnet-argon.kappa}{.69}
\setR{agree.spo.sonnet-argon.pct}{80.0}
\setR{agree.spo.sonnet-sol.kappa}{.60}
\setR{agree.spo.sonnet-sol.pct}{74.7}
\setR{agree.spo.tau}{1.00}
\setR{aud.caus.best}{.83}
\setR{aud.caus.cc}{.69}
\setR{aud.caus.cont}{.82}
\setR{aud.caus.se}{.77}
\setR{aud.caus.vimax}{.83}
\setR{aud.caus.vmem}{.59}
\setR{aud.crit.caus.cc}{5.44}
\setR{aud.crit.caus.cont}{6.59}
\setR{aud.crit.caus.se}{4.33}
\setR{aud.crit.caus.vimax}{4.44}
\setR{aud.crit.caus.vmem}{4.56}
\setR{aud.crit.col.cc}{4.67}
\setR{aud.crit.col.cont}{8.52}
\setR{aud.crit.col.se}{5.89}
\setR{aud.crit.col.vimax}{7.11}
\setR{aud.crit.col.vmem}{7.11}
\setR{aud.crit.cons.cc}{3.33}
\setR{aud.crit.cons.cont}{5.59}
\setR{aud.crit.cons.se}{3.11}
\setR{aud.crit.cons.vimax}{2.89}
\setR{aud.crit.cons.vmem}{4.11}
\setR{aud.crit.cov.cc}{7.11}
\setR{aud.crit.cov.cont}{9.41}
\setR{aud.crit.cov.se}{4.89}
\setR{aud.crit.cov.vimax}{5.67}
\setR{aud.crit.cov.vmem}{5.22}
\setR{aud.crit.fid.cc}{6.22}
\setR{aud.crit.fid.cont}{7.59}
\setR{aud.crit.fid.se}{6.11}
\setR{aud.crit.fid.vimax}{4.67}
\setR{aud.crit.fid.vmem}{4.89}
\setR{aud.crit.film.cc}{5.44}
\setR{aud.crit.film.cont}{8.41}
\setR{aud.crit.film.se}{7.22}
\setR{aud.crit.film.vimax}{7.56}
\setR{aud.crit.film.vmem}{7.44}
\setR{aud.crit.int.cc}{6.44}
\setR{aud.crit.int.cont}{8.93}
\setR{aud.crit.int.se}{6.22}
\setR{aud.crit.int.vimax}{6.56}
\setR{aud.crit.int.vmem}{6.11}
\setR{aud.crit.rel.cc}{5.33}
\setR{aud.crit.rel.cont}{7.59}
\setR{aud.crit.rel.se}{4.67}
\setR{aud.crit.rel.vimax}{4.67}
\setR{aud.crit.rel.vmem}{5.44}
\setR{aud.crit.ver.cc}{4.33}
\setR{aud.crit.ver.cont}{6.07}
\setR{aud.crit.ver.se}{3.56}
\setR{aud.crit.ver.vimax}{3.67}
\setR{aud.crit.ver.vmem}{4.44}
\setR{aud.events.cc}{.84}
\setR{aud.events.cont}{.94}
\setR{aud.events.se}{.81}
\setR{aud.events.vimax}{.89}
\setR{aud.events.vmem}{.75}
\setR{aud.film.bta.read.cc}{6.33}
\setR{aud.film.bta.read.cont}{7.19}
\setR{aud.film.bta.read.se}{4.44}
\setR{aud.film.bta.read.vimax}{4.56}
\setR{aud.film.bta.read.vmem}{5.00}
\setR{aud.film.civildead.read.cc}{6.33}
\setR{aud.film.civildead.read.cont}{7.74}
\setR{aud.film.civildead.read.se}{6.33}
\setR{aud.film.civildead.read.vimax}{6.11}
\setR{aud.film.civildead.read.vmem}{4.44}
\setR{aud.film.detail.read.cc}{5.89}
\setR{aud.film.detail.read.cont}{7.33}
\setR{aud.film.detail.read.se}{3.89}
\setR{aud.film.detail.read.vimax}{4.56}
\setR{aud.film.detail.read.vmem}{5.22}
\setR{aud.film.fastcolor.read.cc}{5.78}
\setR{aud.film.fastcolor.read.cont}{7.89}
\setR{aud.film.fastcolor.read.se}{4.67}
\setR{aud.film.fastcolor.read.vimax}{4.67}
\setR{aud.film.fastcolor.read.vmem}{3.89}
\setR{aud.film.jockey.read.cc}{5.56}
\setR{aud.film.jockey.read.cont}{7.96}
\setR{aud.film.jockey.read.se}{6.89}
\setR{aud.film.jockey.read.vimax}{6.56}
\setR{aud.film.jockey.read.vmem}{6.56}
\setR{aud.film.kin.read.cc}{6.22}
\setR{aud.film.kin.read.cont}{7.85}
\setR{aud.film.kin.read.se}{5.44}
\setR{aud.film.kin.read.vimax}{4.00}
\setR{aud.film.kin.read.vmem}{6.33}
\setR{aud.film.smalltime.read.cc}{3.89}
\setR{aud.film.smalltime.read.cont}{7.33}
\setR{aud.film.smalltime.read.se}{5.11}
\setR{aud.film.smalltime.read.vimax}{5.00}
\setR{aud.film.smalltime.read.vmem}{4.56}
\setR{aud.film.stepback.read.cc}{3.33}
\setR{aud.film.stepback.read.cont}{7.52}
\setR{aud.film.stepback.read.se}{4.22}
\setR{aud.film.stepback.read.vimax}{5.89}
\setR{aud.film.stepback.read.vmem}{7.22}
\setR{aud.film.wildlife.read.cc}{5.00}
\setR{aud.film.wildlife.read.cont}{7.89}
\setR{aud.film.wildlife.read.se}{5.00}
\setR{aud.film.wildlife.read.vimax}{5.89}
\setR{aud.film.wildlife.read.vmem}{6.11}
\setR{aud.pres.best}{.75}
\setR{aud.pres.cc}{.64}
\setR{aud.pres.cont}{.76}
\setR{aud.pres.se}{.75}
\setR{aud.pres.vimax}{.75}
\setR{aud.pres.vmem}{.63}
\setR{aud.read.best}{5.48}
\setR{aud.read.cc}{5.37}
\setR{aud.read.cont}{7.63}
\setR{aud.read.sd}{0.27}
\setR{aud.read.se}{5.11}
\setR{aud.read.vimax}{5.25}
\setR{aud.read.vmem}{5.48}
\setR{aud.rel.best}{.80}
\setR{aud.rel.cc}{.66}
\setR{aud.rel.cont}{.70}
\setR{aud.rel.se}{.72}
\setR{aud.rel.vimax}{.80}
\setR{aud.rel.vmem}{.56}
\setR{ceil.spe.stated}{625}
\setR{ceil.spe.stated.char}{137}
\setR{ceil.spe.stated.col}{84}
\setR{ceil.spe.stated.emo}{140}
\setR{ceil.spe.stated.state}{264}
\setR{ceil.spo.miss.cause.jury}{3}
\setR{ceil.spo.stated}{198}
\setR{ceil.spo.stated.cause}{18}
\setR{ceil.spo.stated.diff}{62}
\setR{ceil.spo.stated.same}{118}
\setR{noise.acc.text}{.75}
\setR{noise.acc.video}{.82}
\setR{noise.chance}{.46}
\setR{noise.flip.text}{.15}
\setR{noise.flip.video}{.14}
\setR{noise.items}{10}
\setR{noise.mirror.text}{.93}
\setR{noise.mirror.video}{.99}
\setR{run.calls.act}{103}
\setR{run.calls.draft}{4{,}733}
\setR{run.calls.draft.ahead}{2{,}873}
\setR{run.calls.draft.first}{3{,}392}
\setR{run.calls.draft.redraft}{1{,}341}
\setR{run.calls.extract}{3{,}309}
\setR{run.calls.film}{14}
\setR{run.calls.notice}{45}
\setR{run.calls.propose}{95}
\setR{run.calls.rewrite}{111}
\setR{run.chk.absent}{229}
\setR{run.chk.camera}{23}
\setR{run.chk.endstate}{140}
\setR{run.chk.forward.cont}{9}
\setR{run.chk.laterneed}{64}
\setR{run.chk.quote}{89}
\setR{run.chk.speech}{13}
\setR{run.clean.cont}{1{,}290}
\setR{run.contdiffers.cont}{14}
\setR{run.controws.cont}{4{,}218}
\setR{run.ctx.first}{10{,}508}
\setR{run.ctx.last}{14{,}784}
\setR{run.ctx.ratio}{1.41}
\setR{run.drafts.cont}{1.41}
\setR{run.extract.findings.cont}{363}
\setR{run.extract.ops.cont}{170}
\setR{run.extract.windows.cont}{144}
\setR{run.facts}{11{,}029}
\setR{run.facts.bible}{174}
\setR{run.factswithreason.pct}{88}
\setR{run.factswriter.pct}{98}
\setR{run.failed.cont}{0}
\setR{run.film.bta.drafts.cont}{1.30}
\setR{run.film.bta.firstpass.cont}{74\%}
\setR{run.film.bta.minutes.cont}{91}
\setR{run.film.bta.open.cont}{3}
\setR{run.film.bta.shots.cont}{890}
\setR{run.film.bta.windows.cont}{152}
\setR{run.film.civildead.drafts.cont}{1.47}
\setR{run.film.civildead.firstpass.cont}{66\%}
\setR{run.film.civildead.minutes.cont}{76}
\setR{run.film.civildead.open.cont}{9}
\setR{run.film.civildead.shots.cont}{768}
\setR{run.film.civildead.windows.cont}{148}
\setR{run.film.detail.drafts.cont}{1.47}
\setR{run.film.detail.firstpass.cont}{68\%}
\setR{run.film.detail.minutes.cont}{79}
\setR{run.film.detail.open.cont}{3}
\setR{run.film.detail.shots.cont}{703}
\setR{run.film.detail.windows.cont}{113}
\setR{run.film.fastcolor.drafts.cont}{1.44}
\setR{run.film.fastcolor.firstpass.cont}{69\%}
\setR{run.film.fastcolor.minutes.cont}{83}
\setR{run.film.fastcolor.open.cont}{5}
\setR{run.film.fastcolor.shots.cont}{992}
\setR{run.film.fastcolor.windows.cont}{212}
\setR{run.film.jockey.drafts.cont}{1.32}
\setR{run.film.jockey.firstpass.cont}{71\%}
\setR{run.film.jockey.minutes.cont}{69}
\setR{run.film.jockey.open.cont}{0}
\setR{run.film.jockey.shots.cont}{639}
\setR{run.film.jockey.windows.cont}{90}
\setR{run.film.kin.drafts.cont}{1.45}
\setR{run.film.kin.firstpass.cont}{64\%}
\setR{run.film.kin.minutes.cont}{100}
\setR{run.film.kin.open.cont}{2}
\setR{run.film.kin.shots.cont}{1{,}137}
\setR{run.film.kin.windows.cont}{220}
\setR{run.film.smalltime.drafts.cont}{1.38}
\setR{run.film.smalltime.firstpass.cont}{72\%}
\setR{run.film.smalltime.minutes.cont}{83}
\setR{run.film.smalltime.open.cont}{0}
\setR{run.film.smalltime.shots.cont}{703}
\setR{run.film.smalltime.windows.cont}{120}
\setR{run.film.stepback.drafts.cont}{1.29}
\setR{run.film.stepback.firstpass.cont}{80\%}
\setR{run.film.stepback.minutes.cont}{89}
\setR{run.film.stepback.open.cont}{2}
\setR{run.film.stepback.shots.cont}{828}
\setR{run.film.stepback.windows.cont}{97}
\setR{run.film.wildlife.drafts.cont}{1.45}
\setR{run.film.wildlife.firstpass.cont}{64\%}
\setR{run.film.wildlife.minutes.cont}{90}
\setR{run.film.wildlife.open.cont}{6}
\setR{run.film.wildlife.shots.cont}{975}
\setR{run.film.wildlife.windows.cont}{168}
\setR{run.firstpass.cont}{69\%}
\setR{run.hours.max}{3.6}
\setR{run.keeper.i}{278}
\setR{run.keeper.ii}{193}
\setR{run.keeper.iii}{9}
\setR{run.keeper.iv}{182}
\setR{run.keeper.other}{1}
\setR{run.keeper.total}{1{,}014}
\setR{run.keeper.v}{58}
\setR{run.keeper.vi}{268}
\setR{run.keeper.vii}{23}
\setR{run.keeper.viii}{2}
\setR{run.minutes.cont}{758}
\setR{run.notices.cont}{45}
\setR{run.notices.failedkeeper.cont}{6}
\setR{run.notices.persist.cont}{2}
\setR{run.notices.rejected.cont}{31}
\setR{run.notices.reverted.cont}{1}
\setR{run.noticescleared.cont}{5}
\setR{run.openissues.cont}{30}
\setR{run.outlinefaults}{31}
\setR{run.outlinefaults.cont}{31}
\setR{run.outlineleft}{0}
\setR{run.outlineleft.cont}{0}
\setR{run.plan.max}{606{,}406}
\setR{run.req.act}{53k}
\setR{run.req.draft}{14k}
\setR{run.req.draft.first}{12.6k}
\setR{run.req.draft.redraft}{16.9k}
\setR{run.req.extract}{7k}
\setR{run.req.film}{35k}
\setR{run.req.notice}{16k}
\setR{run.req.propose}{1k}
\setR{run.req.rewrite}{8k}
\setR{run.rounds.cont}{1{,}860}
\setR{run.scenes}{1{,}016}
\setR{run.shots.cont}{7{,}635}
\setR{run.shots.max}{1{,}137}
\setR{run.shots.min}{639}
\setR{run.shots20}{150}
\setR{run.shotsperscene}{7.5}
\setR{run.spec.hit}{502}
\setR{run.spec.miss}{309}
\setR{run.spec.none}{509}
\setR{run.violations}{1{,}014}
\setR{run.violwindows}{410}
\setR{run.windows.cont}{1{,}320}
\setR{sheet.agree.n}{19{,}098}
\setR{sheet.agree.pct}{65.3}
\setR{spe.all.best.argon}{.76}
\setR{spe.all.best.jury}{.75}
\setR{spe.all.best.noopus}{.75}
\setR{spe.all.cc.argon}{.76}
\setR{spe.all.cc.jury}{.74}
\setR{spe.all.cc.noopus}{.74}
\setR{spe.all.cc.opus}{.74}
\setR{spe.all.cc.sol}{.74}
\setR{spe.all.cc.sonnet}{.73}
\setR{spe.all.ceil.argon}{.98}
\setR{spe.all.ceil.jury}{.98}
\setR{spe.all.ceil.opus}{.97}
\setR{spe.all.ceil.sol}{.99}
\setR{spe.all.ceil.sonnet}{.96}
\setR{spe.all.cont.argon}{.89}
\setR{spe.all.cont.jury}{.86}
\setR{spe.all.cont.noopus}{.88}
\setR{spe.all.cont.opus}{.86}
\setR{spe.all.cont.sol}{.86}
\setR{spe.all.cont.sonnet}{.84}
\setR{spe.all.se.argon}{.76}
\setR{spe.all.se.jury}{.75}
\setR{spe.all.se.noopus}{.75}
\setR{spe.all.se.opus}{.74}
\setR{spe.all.se.sol}{.76}
\setR{spe.all.se.sonnet}{.74}
\setR{spe.all.vimax.argon}{.74}
\setR{spe.all.vimax.jury}{.68}
\setR{spe.all.vimax.noopus}{.70}
\setR{spe.all.vimax.opus}{.69}
\setR{spe.all.vimax.sol}{.69}
\setR{spe.all.vimax.sonnet}{.68}
\setR{spe.all.vmem.argon}{.57}
\setR{spe.all.vmem.jury}{.54}
\setR{spe.all.vmem.noopus}{.56}
\setR{spe.all.vmem.opus}{.56}
\setR{spe.all.vmem.sol}{.55}
\setR{spe.all.vmem.sonnet}{.54}
\setR{spe.char.cc.argon}{.88}
\setR{spe.char.cc.jury}{.81}
\setR{spe.char.cc.noopus}{.84}
\setR{spe.char.ceil.argon}{1.00}
\setR{spe.char.ceil.jury}{1.00}
\setR{spe.char.cont.argon}{.99}
\setR{spe.char.cont.jury}{.94}
\setR{spe.char.cont.noopus}{.96}
\setR{spe.char.se.argon}{.85}
\setR{spe.char.se.jury}{.83}
\setR{spe.char.se.noopus}{.83}
\setR{spe.char.vimax.argon}{.94}
\setR{spe.char.vimax.jury}{.88}
\setR{spe.char.vimax.noopus}{.89}
\setR{spe.char.vmem.argon}{.60}
\setR{spe.char.vmem.jury}{.58}
\setR{spe.char.vmem.noopus}{.61}
\setR{spe.col.cc.argon}{.77}
\setR{spe.col.cc.jury}{.83}
\setR{spe.col.cc.noopus}{.81}
\setR{spe.col.ceil.argon}{1.00}
\setR{spe.col.ceil.jury}{1.00}
\setR{spe.col.cont.argon}{.70}
\setR{spe.col.cont.jury}{.62}
\setR{spe.col.cont.noopus}{.69}
\setR{spe.col.se.argon}{.73}
\setR{spe.col.se.jury}{.70}
\setR{spe.col.se.noopus}{.69}
\setR{spe.col.vimax.argon}{.36}
\setR{spe.col.vimax.jury}{.38}
\setR{spe.col.vimax.noopus}{.38}
\setR{spe.col.vmem.argon}{.65}
\setR{spe.col.vmem.jury}{.61}
\setR{spe.col.vmem.noopus}{.67}
\setR{spe.cov.cc}{605}
\setR{spe.cov.se}{502}
\setR{spe.cov.vimax}{566}
\setR{spe.cov.vmem}{434}
\setR{spe.covscore.cc.argon}{.76}
\setR{spe.covscore.cc.jury}{.75}
\setR{spe.covscore.char.cc.argon}{.90}
\setR{spe.covscore.char.cc.jury}{.83}
\setR{spe.covscore.char.cont.vscc.argon}{.98}
\setR{spe.covscore.char.cont.vscc.jury}{.95}
\setR{spe.covscore.char.cont.vsse.argon}{.98}
\setR{spe.covscore.char.cont.vsse.jury}{.95}
\setR{spe.covscore.char.cont.vsvimax.argon}{.98}
\setR{spe.covscore.char.cont.vsvimax.jury}{.94}
\setR{spe.covscore.char.cont.vsvmem.argon}{.99}
\setR{spe.covscore.char.cont.vsvmem.jury}{.95}
\setR{spe.covscore.char.se.argon}{.96}
\setR{spe.covscore.char.se.jury}{.94}
\setR{spe.covscore.char.vimax.argon}{.97}
\setR{spe.covscore.char.vimax.jury}{.90}
\setR{spe.covscore.char.vmem.argon}{.87}
\setR{spe.covscore.char.vmem.jury}{.86}
\setR{spe.covscore.col.cc.argon}{.79}
\setR{spe.covscore.col.cc.jury}{.85}
\setR{spe.covscore.col.cont.vscc.argon}{.70}
\setR{spe.covscore.col.cont.vscc.jury}{.65}
\setR{spe.covscore.col.cont.vsse.argon}{.70}
\setR{spe.covscore.col.cont.vsse.jury}{.65}
\setR{spe.covscore.col.cont.vsvimax.argon}{.69}
\setR{spe.covscore.col.cont.vsvimax.jury}{.59}
\setR{spe.covscore.col.cont.vsvmem.argon}{.65}
\setR{spe.covscore.col.cont.vsvmem.jury}{.61}
\setR{spe.covscore.col.se.argon}{.87}
\setR{spe.covscore.col.se.jury}{.84}
\setR{spe.covscore.col.vimax.argon}{.40}
\setR{spe.covscore.col.vimax.jury}{.43}
\setR{spe.covscore.col.vmem.argon}{.68}
\setR{spe.covscore.col.vmem.jury}{.64}
\setR{spe.covscore.cont.vscc.argon}{.89}
\setR{spe.covscore.cont.vscc.jury}{.87}
\setR{spe.covscore.cont.vsse.argon}{.89}
\setR{spe.covscore.cont.vsse.jury}{.87}
\setR{spe.covscore.cont.vsvimax.argon}{.90}
\setR{spe.covscore.cont.vsvimax.jury}{.87}
\setR{spe.covscore.cont.vsvmem.argon}{.88}
\setR{spe.covscore.cont.vsvmem.jury}{.85}
\setR{spe.covscore.emo.cc.argon}{.64}
\setR{spe.covscore.emo.cc.jury}{.66}
\setR{spe.covscore.emo.cont.vscc.argon}{.78}
\setR{spe.covscore.emo.cont.vscc.jury}{.83}
\setR{spe.covscore.emo.cont.vsse.argon}{.77}
\setR{spe.covscore.emo.cont.vsse.jury}{.83}
\setR{spe.covscore.emo.cont.vsvimax.argon}{.76}
\setR{spe.covscore.emo.cont.vsvimax.jury}{.82}
\setR{spe.covscore.emo.cont.vsvmem.argon}{.77}
\setR{spe.covscore.emo.cont.vsvmem.jury}{.82}
\setR{spe.covscore.emo.se.argon}{.77}
\setR{spe.covscore.emo.se.jury}{.78}
\setR{spe.covscore.emo.vimax.argon}{.66}
\setR{spe.covscore.emo.vimax.jury}{.70}
\setR{spe.covscore.emo.vmem.argon}{.69}
\setR{spe.covscore.emo.vmem.jury}{.69}
\setR{spe.covscore.se.argon}{.85}
\setR{spe.covscore.se.jury}{.84}
\setR{spe.covscore.state.cc.argon}{.75}
\setR{spe.covscore.state.cc.jury}{.72}
\setR{spe.covscore.state.cont.vscc.argon}{.97}
\setR{spe.covscore.state.cont.vscc.jury}{.92}
\setR{spe.covscore.state.cont.vsse.argon}{.96}
\setR{spe.covscore.state.cont.vsse.jury}{.92}
\setR{spe.covscore.state.cont.vsvimax.argon}{.98}
\setR{spe.covscore.state.cont.vsvimax.jury}{.94}
\setR{spe.covscore.state.cont.vsvmem.argon}{.97}
\setR{spe.covscore.state.cont.vsvmem.jury}{.91}
\setR{spe.covscore.state.se.argon}{.83}
\setR{spe.covscore.state.se.jury}{.82}
\setR{spe.covscore.state.vimax.argon}{.88}
\setR{spe.covscore.state.vimax.jury}{.74}
\setR{spe.covscore.state.vmem.argon}{.61}
\setR{spe.covscore.state.vmem.jury}{.57}
\setR{spe.covscore.vimax.argon}{.79}
\setR{spe.covscore.vimax.jury}{.73}
\setR{spe.covscore.vmem.argon}{.69}
\setR{spe.covscore.vmem.jury}{.66}
\setR{spe.emo.cc.argon}{.65}
\setR{spe.emo.cc.jury}{.66}
\setR{spe.emo.cc.noopus}{.64}
\setR{spe.emo.ceil.argon}{1.00}
\setR{spe.emo.ceil.jury}{1.00}
\setR{spe.emo.cont.argon}{.78}
\setR{spe.emo.cont.jury}{.83}
\setR{spe.emo.cont.noopus}{.79}
\setR{spe.emo.se.argon}{.70}
\setR{spe.emo.se.jury}{.71}
\setR{spe.emo.se.noopus}{.72}
\setR{spe.emo.vimax.argon}{.60}
\setR{spe.emo.vimax.jury}{.64}
\setR{spe.emo.vimax.noopus}{.61}
\setR{spe.emo.vmem.argon}{.55}
\setR{spe.emo.vmem.jury}{.55}
\setR{spe.emo.vmem.noopus}{.54}
\setR{spe.film.bta.cc.argon}{.79}
\setR{spe.film.bta.cc.jury}{.79}
\setR{spe.film.bta.ceil.argon}{.97}
\setR{spe.film.bta.ceil.jury}{.93}
\setR{spe.film.bta.cont.argon}{.89}
\setR{spe.film.bta.cont.jury}{.85}
\setR{spe.film.bta.n}{71}
\setR{spe.film.bta.se.argon}{.83}
\setR{spe.film.bta.se.jury}{.83}
\setR{spe.film.bta.vimax.argon}{.56}
\setR{spe.film.bta.vimax.jury}{.54}
\setR{spe.film.bta.vmem.argon}{.61}
\setR{spe.film.bta.vmem.jury}{.62}
\setR{spe.film.civildead.cc.argon}{.83}
\setR{spe.film.civildead.cc.jury}{.82}
\setR{spe.film.civildead.ceil.argon}{.97}
\setR{spe.film.civildead.ceil.jury}{.98}
\setR{spe.film.civildead.cont.argon}{.91}
\setR{spe.film.civildead.cont.jury}{.89}
\setR{spe.film.civildead.n}{65}
\setR{spe.film.civildead.se.argon}{.88}
\setR{spe.film.civildead.se.jury}{.88}
\setR{spe.film.civildead.vimax.argon}{.86}
\setR{spe.film.civildead.vimax.jury}{.75}
\setR{spe.film.civildead.vmem.argon}{.60}
\setR{spe.film.civildead.vmem.jury}{.54}
\setR{spe.film.detail.cc.argon}{.79}
\setR{spe.film.detail.cc.jury}{.78}
\setR{spe.film.detail.ceil.argon}{.98}
\setR{spe.film.detail.ceil.jury}{.98}
\setR{spe.film.detail.cont.argon}{.88}
\setR{spe.film.detail.cont.jury}{.86}
\setR{spe.film.detail.n}{58}
\setR{spe.film.detail.se.argon}{.90}
\setR{spe.film.detail.se.jury}{.81}
\setR{spe.film.detail.vimax.argon}{.74}
\setR{spe.film.detail.vimax.jury}{.71}
\setR{spe.film.detail.vmem.argon}{.47}
\setR{spe.film.detail.vmem.jury}{.45}
\setR{spe.film.fastcolor.cc.argon}{.78}
\setR{spe.film.fastcolor.cc.jury}{.79}
\setR{spe.film.fastcolor.ceil.argon}{1.00}
\setR{spe.film.fastcolor.ceil.jury}{1.00}
\setR{spe.film.fastcolor.cont.argon}{.90}
\setR{spe.film.fastcolor.cont.jury}{.87}
\setR{spe.film.fastcolor.n}{63}
\setR{spe.film.fastcolor.se.argon}{.62}
\setR{spe.film.fastcolor.se.jury}{.62}
\setR{spe.film.fastcolor.vimax.argon}{.84}
\setR{spe.film.fastcolor.vimax.jury}{.78}
\setR{spe.film.fastcolor.vmem.argon}{.70}
\setR{spe.film.fastcolor.vmem.jury}{.71}
\setR{spe.film.jockey.cc.argon}{.76}
\setR{spe.film.jockey.cc.jury}{.69}
\setR{spe.film.jockey.ceil.argon}{.98}
\setR{spe.film.jockey.ceil.jury}{.95}
\setR{spe.film.jockey.cont.argon}{.93}
\setR{spe.film.jockey.cont.jury}{.90}
\setR{spe.film.jockey.n}{88}
\setR{spe.film.jockey.se.argon}{.91}
\setR{spe.film.jockey.se.jury}{.89}
\setR{spe.film.jockey.vimax.argon}{.83}
\setR{spe.film.jockey.vimax.jury}{.72}
\setR{spe.film.jockey.vmem.argon}{.73}
\setR{spe.film.jockey.vmem.jury}{.67}
\setR{spe.film.kin.cc.argon}{.74}
\setR{spe.film.kin.cc.jury}{.76}
\setR{spe.film.kin.ceil.argon}{.99}
\setR{spe.film.kin.ceil.jury}{1.00}
\setR{spe.film.kin.cont.argon}{.88}
\setR{spe.film.kin.cont.jury}{.84}
\setR{spe.film.kin.n}{80}
\setR{spe.film.kin.se.argon}{.66}
\setR{spe.film.kin.se.jury}{.66}
\setR{spe.film.kin.vimax.argon}{.68}
\setR{spe.film.kin.vimax.jury}{.68}
\setR{spe.film.kin.vmem.argon}{.55}
\setR{spe.film.kin.vmem.jury}{.56}
\setR{spe.film.smalltime.cc.argon}{.80}
\setR{spe.film.smalltime.cc.jury}{.70}
\setR{spe.film.smalltime.ceil.argon}{1.00}
\setR{spe.film.smalltime.ceil.jury}{1.00}
\setR{spe.film.smalltime.cont.argon}{.93}
\setR{spe.film.smalltime.cont.jury}{.91}
\setR{spe.film.smalltime.n}{46}
\setR{spe.film.smalltime.se.argon}{.91}
\setR{spe.film.smalltime.se.jury}{.85}
\setR{spe.film.smalltime.vimax.argon}{.85}
\setR{spe.film.smalltime.vimax.jury}{.74}
\setR{spe.film.smalltime.vmem.argon}{.24}
\setR{spe.film.smalltime.vmem.jury}{.24}
\setR{spe.film.stepback.cc.argon}{.69}
\setR{spe.film.stepback.cc.jury}{.69}
\setR{spe.film.stepback.ceil.argon}{.99}
\setR{spe.film.stepback.ceil.jury}{.96}
\setR{spe.film.stepback.cont.argon}{.88}
\setR{spe.film.stepback.cont.jury}{.77}
\setR{spe.film.stepback.n}{75}
\setR{spe.film.stepback.se.argon}{.41}
\setR{spe.film.stepback.se.jury}{.41}
\setR{spe.film.stepback.vimax.argon}{.76}
\setR{spe.film.stepback.vimax.jury}{.65}
\setR{spe.film.stepback.vmem.argon}{.71}
\setR{spe.film.stepback.vmem.jury}{.68}
\setR{spe.film.wildlife.cc.argon}{.66}
\setR{spe.film.wildlife.cc.jury}{.67}
\setR{spe.film.wildlife.ceil.argon}{.99}
\setR{spe.film.wildlife.ceil.jury}{1.00}
\setR{spe.film.wildlife.cont.argon}{.86}
\setR{spe.film.wildlife.cont.jury}{.90}
\setR{spe.film.wildlife.n}{79}
\setR{spe.film.wildlife.se.argon}{.77}
\setR{spe.film.wildlife.se.jury}{.80}
\setR{spe.film.wildlife.vimax.argon}{.62}
\setR{spe.film.wildlife.vimax.jury}{.61}
\setR{spe.film.wildlife.vmem.argon}{.38}
\setR{spe.film.wildlife.vmem.jury}{.30}
\setR{spe.first.argon}{8}
\setR{spe.first.jury}{9}
\setR{spe.gap.max}{125}
\setR{spe.gap.median}{34}
\setR{spe.gap.min}{20}
\setR{spe.lead.all.argon}{14}
\setR{spe.lead.all.argon.cl.hi}{26}
\setR{spe.lead.all.argon.cl.lo}{4}
\setR{spe.lead.all.jury}{12}
\setR{spe.lead.all.jury.cl.hi}{22}
\setR{spe.lead.all.jury.cl.lo}{4}
\setR{spe.lead.all.jury.mde}{13}
\setR{spe.lead.all.noopus}{13}
\setR{spe.lead.all.noopus.cl.hi}{24}
\setR{spe.lead.all.noopus.cl.lo}{4}
\setR{spe.lead.char.jury}{7}
\setR{spe.lead.char.jury.cl.hi}{16}
\setR{spe.lead.char.jury.cl.lo}{-1}
\setR{spe.lead.char.noopus}{7}
\setR{spe.lead.char.noopus.cl.hi}{16}
\setR{spe.lead.char.noopus.cl.lo}{1}
\setR{spe.lead.col.jury}{-21}
\setR{spe.lead.col.jury.cl.hi}{-1}
\setR{spe.lead.col.jury.cl.lo}{-42}
\setR{spe.lead.col.noopus}{-12}
\setR{spe.lead.col.noopus.cl.hi}{4}
\setR{spe.lead.col.noopus.cl.lo}{-29}
\setR{spe.lead.emo.jury}{12}
\setR{spe.lead.emo.jury.cl.hi}{24}
\setR{spe.lead.emo.jury.cl.lo}{2}
\setR{spe.lead.emo.noopus}{6}
\setR{spe.lead.emo.noopus.cl.hi}{19}
\setR{spe.lead.emo.noopus.cl.lo}{-4}
\setR{spe.lead.state.jury}{19}
\setR{spe.lead.state.jury.cl.hi}{31}
\setR{spe.lead.state.jury.cl.lo}{8}
\setR{spe.lead.state.noopus}{19}
\setR{spe.lead.state.noopus.cl.hi}{27}
\setR{spe.lead.state.noopus.cl.lo}{10}
\setR{spe.n.all}{625}
\setR{spe.n.char}{137}
\setR{spe.n.col}{84}
\setR{spe.n.emo}{140}
\setR{spe.n.state}{264}
\setR{spe.nodev.all.best.jury}{.75}
\setR{spe.nodev.all.cc.argon}{.75}
\setR{spe.nodev.all.cc.jury}{.75}
\setR{spe.nodev.all.cont.argon}{.88}
\setR{spe.nodev.all.cont.jury}{.85}
\setR{spe.nodev.all.se.argon}{.72}
\setR{spe.nodev.all.se.jury}{.71}
\setR{spe.nodev.all.vimax.argon}{.72}
\setR{spe.nodev.all.vimax.jury}{.67}
\setR{spe.nodev.all.vmem.argon}{.57}
\setR{spe.nodev.all.vmem.jury}{.55}
\setR{spe.nodev.char.cc.argon}{.87}
\setR{spe.nodev.char.cc.jury}{.82}
\setR{spe.nodev.char.cont.argon}{.98}
\setR{spe.nodev.char.cont.jury}{.95}
\setR{spe.nodev.char.se.argon}{.79}
\setR{spe.nodev.char.se.jury}{.79}
\setR{spe.nodev.char.vimax.argon}{.92}
\setR{spe.nodev.char.vimax.jury}{.88}
\setR{spe.nodev.char.vmem.argon}{.64}
\setR{spe.nodev.char.vmem.jury}{.62}
\setR{spe.nodev.col.cc.argon}{.74}
\setR{spe.nodev.col.cc.jury}{.81}
\setR{spe.nodev.col.cont.argon}{.68}
\setR{spe.nodev.col.cont.jury}{.58}
\setR{spe.nodev.col.se.argon}{.69}
\setR{spe.nodev.col.se.jury}{.66}
\setR{spe.nodev.col.vimax.argon}{.32}
\setR{spe.nodev.col.vimax.jury}{.35}
\setR{spe.nodev.col.vmem.argon}{.61}
\setR{spe.nodev.col.vmem.jury}{.57}
\setR{spe.nodev.emo.cc.argon}{.65}
\setR{spe.nodev.emo.cc.jury}{.70}
\setR{spe.nodev.emo.cont.argon}{.78}
\setR{spe.nodev.emo.cont.jury}{.83}
\setR{spe.nodev.emo.se.argon}{.68}
\setR{spe.nodev.emo.se.jury}{.70}
\setR{spe.nodev.emo.vimax.argon}{.58}
\setR{spe.nodev.emo.vimax.jury}{.64}
\setR{spe.nodev.emo.vmem.argon}{.52}
\setR{spe.nodev.emo.vmem.jury}{.53}
\setR{spe.nodev.lead.all.argon}{13}
\setR{spe.nodev.lead.all.argon.cl.hi}{19}
\setR{spe.nodev.lead.all.argon.cl.lo}{8}
\setR{spe.nodev.lead.all.jury}{10}
\setR{spe.nodev.lead.all.jury.cl.hi}{16}
\setR{spe.nodev.lead.all.jury.cl.lo}{5}
\setR{spe.nodev.lead.char.argon}{6}
\setR{spe.nodev.lead.char.argon.cl.hi}{19}
\setR{spe.nodev.lead.char.argon.cl.lo}{-3}
\setR{spe.nodev.lead.char.jury}{6}
\setR{spe.nodev.lead.char.jury.cl.hi}{19}
\setR{spe.nodev.lead.char.jury.cl.lo}{-4}
\setR{spe.nodev.lead.col.argon}{-7}
\setR{spe.nodev.lead.col.argon.cl.hi}{11}
\setR{spe.nodev.lead.col.argon.cl.lo}{-27}
\setR{spe.nodev.lead.col.jury}{-23}
\setR{spe.nodev.lead.col.jury.cl.hi}{0}
\setR{spe.nodev.lead.col.jury.cl.lo}{-46}
\setR{spe.nodev.lead.emo.argon}{9}
\setR{spe.nodev.lead.emo.argon.cl.hi}{27}
\setR{spe.nodev.lead.emo.argon.cl.lo}{-4}
\setR{spe.nodev.lead.emo.jury}{13}
\setR{spe.nodev.lead.emo.jury.cl.hi}{28}
\setR{spe.nodev.lead.emo.jury.cl.lo}{0}
\setR{spe.nodev.lead.state.argon}{13}
\setR{spe.nodev.lead.state.argon.cl.hi}{23}
\setR{spe.nodev.lead.state.argon.cl.lo}{5}
\setR{spe.nodev.lead.state.jury}{19}
\setR{spe.nodev.lead.state.jury.cl.hi}{28}
\setR{spe.nodev.lead.state.jury.cl.lo}{9}
\setR{spe.nodev.n}{491}
\setR{spe.nodev.pairs}{27}
\setR{spe.nodev.state.cc.argon}{.74}
\setR{spe.nodev.state.cc.jury}{.73}
\setR{spe.nodev.state.cont.argon}{.97}
\setR{spe.nodev.state.cont.jury}{.92}
\setR{spe.nodev.state.se.argon}{.71}
\setR{spe.nodev.state.se.jury}{.70}
\setR{spe.nodev.state.vimax.argon}{.83}
\setR{spe.nodev.state.vimax.jury}{.70}
\setR{spe.nodev.state.vmem.argon}{.55}
\setR{spe.nodev.state.vmem.jury}{.52}
\setR{spe.pairs}{33}
\setR{spe.state.cc.argon}{.74}
\setR{spe.state.cc.jury}{.72}
\setR{spe.state.cc.noopus}{.72}
\setR{spe.state.ceil.argon}{.96}
\setR{spe.state.ceil.jury}{.95}
\setR{spe.state.cont.argon}{.97}
\setR{spe.state.cont.jury}{.92}
\setR{spe.state.cont.noopus}{.94}
\setR{spe.state.se.argon}{.75}
\setR{spe.state.se.jury}{.73}
\setR{spe.state.se.noopus}{.75}
\setR{spe.state.vimax.argon}{.84}
\setR{spe.state.vimax.jury}{.70}
\setR{spe.state.vimax.noopus}{.76}
\setR{spe.state.vmem.argon}{.53}
\setR{spe.state.vmem.jury}{.50}
\setR{spe.state.vmem.noopus}{.52}
\setR{spo.all.best.argon}{.44}
\setR{spo.all.best.jury}{.40}
\setR{spo.all.best.noopus}{.42}
\setR{spo.all.cc.argon}{.34}
\setR{spo.all.cc.jury}{.33}
\setR{spo.all.cc.noopus}{.34}
\setR{spo.all.cc.opus}{.37}
\setR{spo.all.cc.sol}{.31}
\setR{spo.all.cc.sonnet}{.35}
\setR{spo.all.ceil.argon}{.97}
\setR{spo.all.ceil.jury}{.98}
\setR{spo.all.ceil.opus}{.98}
\setR{spo.all.ceil.sol}{.98}
\setR{spo.all.ceil.sonnet}{.95}
\setR{spo.all.cont.argon}{.65}
\setR{spo.all.cont.jury}{.59}
\setR{spo.all.cont.noopus}{.59}
\setR{spo.all.cont.opus}{.65}
\setR{spo.all.cont.sol}{.49}
\setR{spo.all.cont.sonnet}{.57}
\setR{spo.all.se.argon}{.44}
\setR{spo.all.se.jury}{.40}
\setR{spo.all.se.noopus}{.42}
\setR{spo.all.se.opus}{.44}
\setR{spo.all.se.sol}{.37}
\setR{spo.all.se.sonnet}{.41}
\setR{spo.all.vimax.argon}{.18}
\setR{spo.all.vimax.jury}{.14}
\setR{spo.all.vimax.noopus}{.14}
\setR{spo.all.vimax.opus}{.19}
\setR{spo.all.vimax.sol}{.11}
\setR{spo.all.vimax.sonnet}{.15}
\setR{spo.all.vmem.argon}{.06}
\setR{spo.all.vmem.jury}{.05}
\setR{spo.all.vmem.noopus}{.04}
\setR{spo.all.vmem.opus}{.08}
\setR{spo.all.vmem.sol}{.04}
\setR{spo.all.vmem.sonnet}{.06}
\setR{spo.bal.best.argon}{.45}
\setR{spo.bal.best.jury}{.42}
\setR{spo.bal.cont.argon}{.62}
\setR{spo.bal.cont.jury}{.59}
\setR{spo.carry.cc.16to39.n}{2}
\setR{spo.carry.cc.n}{37}
\setR{spo.carry.cc.noid.right.argon}{.07}
\setR{spo.carry.cc.noid.right.jury}{.09}
\setR{spo.carry.cc.right.argon}{1.00}
\setR{spo.carry.cc.right.jury}{.97}
\setR{spo.carry.cont.16to39.n}{10}
\setR{spo.carry.cont.n}{71}
\setR{spo.carry.cont.noid.right.argon}{.30}
\setR{spo.carry.cont.noid.right.jury}{.26}
\setR{spo.carry.cont.right.argon}{.89}
\setR{spo.carry.cont.right.jury}{.72}
\setR{spo.carry.se.16to39.n}{2}
\setR{spo.carry.se.n}{40}
\setR{spo.carry.se.noid.right.argon}{.05}
\setR{spo.carry.se.noid.right.jury}{.05}
\setR{spo.carry.se.right.argon}{.93}
\setR{spo.carry.se.right.jury}{.80}
\setR{spo.carry.vimax.n}{0}
\setR{spo.carry.vimax.noid.right.argon}{.09}
\setR{spo.carry.vimax.noid.right.jury}{.09}
\setR{spo.carry.vmem.16to39.n}{2}
\setR{spo.carry.vmem.n}{26}
\setR{spo.carry.vmem.noid.right.argon}{.01}
\setR{spo.carry.vmem.noid.right.jury}{.01}
\setR{spo.carry.vmem.right.argon}{.00}
\setR{spo.carry.vmem.right.jury}{.00}
\setR{spo.cause.cc.argon}{.72}
\setR{spo.cause.cc.jury}{.50}
\setR{spo.cause.cc.noopus}{.67}
\setR{spo.cause.ceil.argon}{1.00}
\setR{spo.cause.ceil.jury}{.83}
\setR{spo.cause.cont.argon}{.89}
\setR{spo.cause.cont.jury}{.72}
\setR{spo.cause.cont.noopus}{.78}
\setR{spo.cause.se.argon}{.72}
\setR{spo.cause.se.jury}{.61}
\setR{spo.cause.se.noopus}{.67}
\setR{spo.cause.vimax.argon}{.72}
\setR{spo.cause.vimax.jury}{.50}
\setR{spo.cause.vimax.noopus}{.56}
\setR{spo.cause.vmem.argon}{.28}
\setR{spo.cause.vmem.jury}{.22}
\setR{spo.cause.vmem.noopus}{.11}
\setR{spo.cov.cc}{197}
\setR{spo.cov.se}{155}
\setR{spo.cov.vimax}{185}
\setR{spo.cov.vmem}{142}
\setR{spo.covscore.all.cc.argon}{.35}
\setR{spo.covscore.all.cc.jury}{.34}
\setR{spo.covscore.all.cont.vscc.argon}{.65}
\setR{spo.covscore.all.cont.vscc.jury}{.58}
\setR{spo.covscore.all.cont.vsse.argon}{.64}
\setR{spo.covscore.all.cont.vsse.jury}{.58}
\setR{spo.covscore.all.cont.vsvimax.argon}{.65}
\setR{spo.covscore.all.cont.vsvimax.jury}{.59}
\setR{spo.covscore.all.cont.vsvmem.argon}{.66}
\setR{spo.covscore.all.cont.vsvmem.jury}{.61}
\setR{spo.covscore.all.se.argon}{.56}
\setR{spo.covscore.all.se.jury}{.51}
\setR{spo.covscore.all.vimax.argon}{.19}
\setR{spo.covscore.all.vimax.jury}{.15}
\setR{spo.covscore.all.vmem.argon}{.08}
\setR{spo.covscore.all.vmem.jury}{.07}
\setR{spo.covscore.diff.cc.argon}{.20}
\setR{spo.covscore.diff.cc.jury}{.23}
\setR{spo.covscore.diff.cont.vscc.argon}{.57}
\setR{spo.covscore.diff.cont.vscc.jury}{.64}
\setR{spo.covscore.diff.cont.vsse.argon}{.56}
\setR{spo.covscore.diff.cont.vsse.jury}{.62}
\setR{spo.covscore.diff.cont.vsvimax.argon}{.57}
\setR{spo.covscore.diff.cont.vsvimax.jury}{.63}
\setR{spo.covscore.diff.cont.vsvmem.argon}{.61}
\setR{spo.covscore.diff.cont.vsvmem.jury}{.68}
\setR{spo.covscore.diff.se.argon}{.76}
\setR{spo.covscore.diff.se.jury}{.73}
\setR{spo.covscore.diff.vimax.argon}{.18}
\setR{spo.covscore.diff.vimax.jury}{.13}
\setR{spo.covscore.diff.vmem.argon}{.15}
\setR{spo.covscore.diff.vmem.jury}{.12}
\setR{spo.covscore.same.cc.argon}{.36}
\setR{spo.covscore.same.cc.jury}{.36}
\setR{spo.covscore.same.cont.vscc.argon}{.65}
\setR{spo.covscore.same.cont.vscc.jury}{.53}
\setR{spo.covscore.same.cont.vsse.argon}{.65}
\setR{spo.covscore.same.cont.vsse.jury}{.53}
\setR{spo.covscore.same.cont.vsvimax.argon}{.65}
\setR{spo.covscore.same.cont.vsvimax.jury}{.54}
\setR{spo.covscore.same.cont.vsvmem.argon}{.65}
\setR{spo.covscore.same.cont.vsvmem.jury}{.55}
\setR{spo.covscore.same.se.argon}{.43}
\setR{spo.covscore.same.se.jury}{.38}
\setR{spo.covscore.same.vimax.argon}{.10}
\setR{spo.covscore.same.vimax.jury}{.10}
\setR{spo.covscore.same.vmem.argon}{.01}
\setR{spo.covscore.same.vmem.jury}{.01}
\setR{spo.diff.cc.argon}{.19}
\setR{spo.diff.cc.jury}{.23}
\setR{spo.diff.cc.noopus}{.18}
\setR{spo.diff.ceil.argon}{1.00}
\setR{spo.diff.ceil.jury}{1.00}
\setR{spo.diff.cont.argon}{.58}
\setR{spo.diff.cont.jury}{.65}
\setR{spo.diff.cont.noopus}{.58}
\setR{spo.diff.se.argon}{.55}
\setR{spo.diff.se.jury}{.53}
\setR{spo.diff.se.noopus}{.55}
\setR{spo.diff.vimax.argon}{.18}
\setR{spo.diff.vimax.jury}{.13}
\setR{spo.diff.vimax.noopus}{.15}
\setR{spo.diff.vmem.argon}{.10}
\setR{spo.diff.vmem.jury}{.08}
\setR{spo.diff.vmem.noopus}{.08}
\setR{spo.film.bta.cc.argon}{.23}
\setR{spo.film.bta.cc.jury}{.23}
\setR{spo.film.bta.ceil.argon}{1.00}
\setR{spo.film.bta.ceil.jury}{.92}
\setR{spo.film.bta.cont.argon}{.69}
\setR{spo.film.bta.cont.jury}{.38}
\setR{spo.film.bta.n}{13}
\setR{spo.film.bta.pairs}{11}
\setR{spo.film.bta.se.argon}{.23}
\setR{spo.film.bta.se.jury}{.23}
\setR{spo.film.bta.vimax.argon}{.15}
\setR{spo.film.bta.vimax.jury}{.00}
\setR{spo.film.bta.vmem.argon}{.08}
\setR{spo.film.bta.vmem.jury}{.00}
\setR{spo.film.civildead.cc.argon}{.19}
\setR{spo.film.civildead.cc.jury}{.19}
\setR{spo.film.civildead.ceil.argon}{1.00}
\setR{spo.film.civildead.ceil.jury}{1.00}
\setR{spo.film.civildead.cont.argon}{.61}
\setR{spo.film.civildead.cont.jury}{.61}
\setR{spo.film.civildead.n}{31}
\setR{spo.film.civildead.pairs}{30}
\setR{spo.film.civildead.se.argon}{.52}
\setR{spo.film.civildead.se.jury}{.48}
\setR{spo.film.civildead.vimax.argon}{.16}
\setR{spo.film.civildead.vimax.jury}{.13}
\setR{spo.film.civildead.vmem.argon}{.06}
\setR{spo.film.civildead.vmem.jury}{.03}
\setR{spo.film.detail.cc.argon}{.78}
\setR{spo.film.detail.cc.jury}{.78}
\setR{spo.film.detail.ceil.argon}{1.00}
\setR{spo.film.detail.ceil.jury}{1.00}
\setR{spo.film.detail.cont.argon}{.56}
\setR{spo.film.detail.cont.jury}{.50}
\setR{spo.film.detail.n}{18}
\setR{spo.film.detail.pairs}{17}
\setR{spo.film.detail.se.argon}{.22}
\setR{spo.film.detail.se.jury}{.22}
\setR{spo.film.detail.vimax.argon}{.56}
\setR{spo.film.detail.vimax.jury}{.33}
\setR{spo.film.detail.vmem.argon}{.06}
\setR{spo.film.detail.vmem.jury}{.06}
\setR{spo.film.fastcolor.cc.argon}{.50}
\setR{spo.film.fastcolor.cc.jury}{.47}
\setR{spo.film.fastcolor.ceil.argon}{.86}
\setR{spo.film.fastcolor.ceil.jury}{1.00}
\setR{spo.film.fastcolor.cont.argon}{.75}
\setR{spo.film.fastcolor.cont.jury}{.64}
\setR{spo.film.fastcolor.n}{36}
\setR{spo.film.fastcolor.pairs}{35}
\setR{spo.film.fastcolor.se.argon}{.58}
\setR{spo.film.fastcolor.se.jury}{.56}
\setR{spo.film.fastcolor.vimax.argon}{.08}
\setR{spo.film.fastcolor.vimax.jury}{.11}
\setR{spo.film.fastcolor.vmem.argon}{.06}
\setR{spo.film.fastcolor.vmem.jury}{.03}
\setR{spo.film.jockey.cc.argon}{.25}
\setR{spo.film.jockey.cc.jury}{.19}
\setR{spo.film.jockey.ceil.argon}{1.00}
\setR{spo.film.jockey.ceil.jury}{.94}
\setR{spo.film.jockey.cont.argon}{.44}
\setR{spo.film.jockey.cont.jury}{.56}
\setR{spo.film.jockey.n}{16}
\setR{spo.film.jockey.pairs}{14}
\setR{spo.film.jockey.se.argon}{.44}
\setR{spo.film.jockey.se.jury}{.44}
\setR{spo.film.jockey.vimax.argon}{.19}
\setR{spo.film.jockey.vimax.jury}{.12}
\setR{spo.film.jockey.vmem.argon}{.00}
\setR{spo.film.jockey.vmem.jury}{.00}
\setR{spo.film.kin.cc.argon}{.22}
\setR{spo.film.kin.cc.jury}{.13}
\setR{spo.film.kin.ceil.argon}{1.00}
\setR{spo.film.kin.ceil.jury}{1.00}
\setR{spo.film.kin.cont.argon}{.57}
\setR{spo.film.kin.cont.jury}{.48}
\setR{spo.film.kin.n}{23}
\setR{spo.film.kin.pairs}{21}
\setR{spo.film.kin.se.argon}{.52}
\setR{spo.film.kin.se.jury}{.48}
\setR{spo.film.kin.vimax.argon}{.13}
\setR{spo.film.kin.vimax.jury}{.22}
\setR{spo.film.kin.vmem.argon}{.00}
\setR{spo.film.kin.vmem.jury}{.04}
\setR{spo.film.smalltime.cc.argon}{.33}
\setR{spo.film.smalltime.cc.jury}{.00}
\setR{spo.film.smalltime.ceil.argon}{1.00}
\setR{spo.film.smalltime.ceil.jury}{.67}
\setR{spo.film.smalltime.cont.argon}{1.00}
\setR{spo.film.smalltime.cont.jury}{.33}
\setR{spo.film.smalltime.n}{3}
\setR{spo.film.smalltime.pairs}{2}
\setR{spo.film.smalltime.se.argon}{1.00}
\setR{spo.film.smalltime.se.jury}{.67}
\setR{spo.film.smalltime.vimax.argon}{.67}
\setR{spo.film.smalltime.vimax.jury}{.33}
\setR{spo.film.smalltime.vmem.argon}{.00}
\setR{spo.film.smalltime.vmem.jury}{.00}
\setR{spo.film.stepback.cc.argon}{.48}
\setR{spo.film.stepback.cc.jury}{.52}
\setR{spo.film.stepback.ceil.argon}{1.00}
\setR{spo.film.stepback.ceil.jury}{1.00}
\setR{spo.film.stepback.cont.argon}{.81}
\setR{spo.film.stepback.cont.jury}{.71}
\setR{spo.film.stepback.n}{21}
\setR{spo.film.stepback.pairs}{20}
\setR{spo.film.stepback.se.argon}{.43}
\setR{spo.film.stepback.se.jury}{.29}
\setR{spo.film.stepback.vimax.argon}{.10}
\setR{spo.film.stepback.vimax.jury}{.10}
\setR{spo.film.stepback.vmem.argon}{.05}
\setR{spo.film.stepback.vmem.jury}{.10}
\setR{spo.film.wildlife.cc.argon}{.19}
\setR{spo.film.wildlife.cc.jury}{.24}
\setR{spo.film.wildlife.ceil.argon}{1.00}
\setR{spo.film.wildlife.ceil.jury}{1.00}
\setR{spo.film.wildlife.cont.argon}{.65}
\setR{spo.film.wildlife.cont.jury}{.65}
\setR{spo.film.wildlife.n}{37}
\setR{spo.film.wildlife.pairs}{35}
\setR{spo.film.wildlife.se.argon}{.35}
\setR{spo.film.wildlife.se.jury}{.32}
\setR{spo.film.wildlife.vimax.argon}{.14}
\setR{spo.film.wildlife.vimax.jury}{.11}
\setR{spo.film.wildlife.vmem.argon}{.14}
\setR{spo.film.wildlife.vmem.jury}{.11}
\setR{spo.first.argon}{6}
\setR{spo.first.jury}{6}
\setR{spo.gap.16to39.all.cc.argon}{.23}
\setR{spo.gap.16to39.all.cc.jury}{.17}
\setR{spo.gap.16to39.all.ceil.argon}{1.00}
\setR{spo.gap.16to39.all.ceil.jury}{1.00}
\setR{spo.gap.16to39.all.cont.argon}{.62}
\setR{spo.gap.16to39.all.cont.jury}{.57}
\setR{spo.gap.16to39.all.se.argon}{.36}
\setR{spo.gap.16to39.all.se.jury}{.28}
\setR{spo.gap.16to39.all.vimax.argon}{.15}
\setR{spo.gap.16to39.all.vimax.jury}{.15}
\setR{spo.gap.16to39.all.vmem.argon}{.09}
\setR{spo.gap.16to39.all.vmem.jury}{.13}
\setR{spo.gap.16to39.dec.cov.same.argon}{4}
\setR{spo.gap.16to39.dec.cov.same.jury}{5}
\setR{spo.gap.16to39.dec.id.same.argon}{51}
\setR{spo.gap.16to39.dec.id.same.jury}{30}
\setR{spo.gap.16to39.dec.id.same.jury.cl.hi}{82}
\setR{spo.gap.16to39.dec.id.same.jury.cl.lo}{-15}
\setR{spo.gap.16to39.lead.same.argon}{55}
\setR{spo.gap.16to39.lead.same.argon.cl.hi}{82}
\setR{spo.gap.16to39.lead.same.argon.cl.lo}{24}
\setR{spo.gap.16to39.lead.same.jury}{35}
\setR{spo.gap.16to39.lead.same.jury.cl.hi}{84}
\setR{spo.gap.16to39.lead.same.jury.cl.lo}{-8}
\setR{spo.gap.16to39.n.all}{47}
\setR{spo.gap.16to39.n.same}{20}
\setR{spo.gap.16to39.pairs}{43}
\setR{spo.gap.16to39.same.best.argon}{.10}
\setR{spo.gap.16to39.same.best.jury}{.10}
\setR{spo.gap.16to39.same.cc.argon}{.10}
\setR{spo.gap.16to39.same.cc.jury}{.10}
\setR{spo.gap.16to39.same.ceil.argon}{1.00}
\setR{spo.gap.16to39.same.ceil.jury}{1.00}
\setR{spo.gap.16to39.same.cont.argon}{.65}
\setR{spo.gap.16to39.same.cont.jury}{.45}
\setR{spo.gap.16to39.same.se.argon}{.10}
\setR{spo.gap.16to39.same.se.jury}{.05}
\setR{spo.gap.16to39.same.vimax.argon}{.00}
\setR{spo.gap.16to39.same.vimax.jury}{.00}
\setR{spo.gap.16to39.same.vmem.argon}{.00}
\setR{spo.gap.16to39.same.vmem.jury}{.00}
\setR{spo.gap.2to5.all.cc.argon}{.40}
\setR{spo.gap.2to5.all.cc.jury}{.40}
\setR{spo.gap.2to5.all.ceil.argon}{.98}
\setR{spo.gap.2to5.all.ceil.jury}{.97}
\setR{spo.gap.2to5.all.cont.argon}{.72}
\setR{spo.gap.2to5.all.cont.jury}{.60}
\setR{spo.gap.2to5.all.se.argon}{.47}
\setR{spo.gap.2to5.all.se.jury}{.45}
\setR{spo.gap.2to5.all.vimax.argon}{.19}
\setR{spo.gap.2to5.all.vimax.jury}{.14}
\setR{spo.gap.2to5.all.vmem.argon}{.04}
\setR{spo.gap.2to5.all.vmem.jury}{.02}
\setR{spo.gap.2to5.dec.cov.same.argon}{14}
\setR{spo.gap.2to5.dec.cov.same.jury}{12}
\setR{spo.gap.2to5.dec.cov.same.jury.cl.hi}{21}
\setR{spo.gap.2to5.dec.cov.same.jury.cl.lo}{5}
\setR{spo.gap.2to5.dec.id.same.argon}{12}
\setR{spo.gap.2to5.dec.id.same.jury}{4}
\setR{spo.gap.2to5.lead.same.argon}{26}
\setR{spo.gap.2to5.lead.same.argon.cl.hi}{50}
\setR{spo.gap.2to5.lead.same.argon.cl.lo}{4}
\setR{spo.gap.2to5.lead.same.jury}{17}
\setR{spo.gap.2to5.lead.same.jury.cl.hi}{39}
\setR{spo.gap.2to5.lead.same.jury.cl.lo}{-5}
\setR{spo.gap.2to5.n.all}{93}
\setR{spo.gap.2to5.n.same}{66}
\setR{spo.gap.2to5.pairs}{86}
\setR{spo.gap.2to5.same.cc.argon}{.39}
\setR{spo.gap.2to5.same.cc.jury}{.39}
\setR{spo.gap.2to5.same.ceil.argon}{.97}
\setR{spo.gap.2to5.same.ceil.jury}{1.00}
\setR{spo.gap.2to5.same.cont.argon}{.70}
\setR{spo.gap.2to5.same.cont.jury}{.59}
\setR{spo.gap.2to5.same.se.argon}{.44}
\setR{spo.gap.2to5.same.se.jury}{.42}
\setR{spo.gap.2to5.same.vimax.argon}{.12}
\setR{spo.gap.2to5.same.vimax.jury}{.12}
\setR{spo.gap.2to5.same.vmem.argon}{.02}
\setR{spo.gap.2to5.same.vmem.jury}{.02}
\setR{spo.gap.6to15.all.cc.argon}{.34}
\setR{spo.gap.6to15.all.cc.jury}{.36}
\setR{spo.gap.6to15.all.ceil.argon}{.95}
\setR{spo.gap.6to15.all.ceil.jury}{1.00}
\setR{spo.gap.6to15.all.cont.argon}{.57}
\setR{spo.gap.6to15.all.cont.jury}{.57}
\setR{spo.gap.6to15.all.se.argon}{.47}
\setR{spo.gap.6to15.all.se.jury}{.43}
\setR{spo.gap.6to15.all.vimax.argon}{.17}
\setR{spo.gap.6to15.all.vimax.jury}{.14}
\setR{spo.gap.6to15.all.vmem.argon}{.07}
\setR{spo.gap.6to15.all.vmem.jury}{.03}
\setR{spo.gap.6to15.lead.same.argon}{9}
\setR{spo.gap.6to15.lead.same.argon.cl.hi}{53}
\setR{spo.gap.6to15.lead.same.argon.cl.lo}{-22}
\setR{spo.gap.6to15.lead.same.jury}{0}
\setR{spo.gap.6to15.lead.same.jury.cl.hi}{45}
\setR{spo.gap.6to15.lead.same.jury.cl.lo}{-34}
\setR{spo.gap.6to15.n.all}{58}
\setR{spo.gap.6to15.n.same}{32}
\setR{spo.gap.6to15.pairs}{56}
\setR{spo.gap.6to15.same.cc.argon}{.47}
\setR{spo.gap.6to15.same.cc.jury}{.47}
\setR{spo.gap.6to15.same.ceil.argon}{.91}
\setR{spo.gap.6to15.same.ceil.jury}{1.00}
\setR{spo.gap.6to15.same.cont.argon}{.56}
\setR{spo.gap.6to15.same.cont.jury}{.47}
\setR{spo.gap.6to15.same.se.argon}{.31}
\setR{spo.gap.6to15.same.se.jury}{.22}
\setR{spo.gap.6to15.same.vimax.argon}{.09}
\setR{spo.gap.6to15.same.vimax.jury}{.09}
\setR{spo.gap.6to15.same.vmem.argon}{.00}
\setR{spo.gap.6to15.same.vmem.jury}{.00}
\setR{spo.gap.max}{39}
\setR{spo.gap.median}{6}
\setR{spo.gap.min}{2}
\setR{spo.gate.diff.cc.argon}{.79}
\setR{spo.gate.diff.cc.jury}{.84}
\setR{spo.gate.diff.ceil.argon}{1.00}
\setR{spo.gate.diff.ceil.jury}{1.00}
\setR{spo.gate.diff.cont.argon}{.94}
\setR{spo.gate.diff.cont.jury}{.94}
\setR{spo.gate.diff.se.argon}{.69}
\setR{spo.gate.diff.se.jury}{.68}
\setR{spo.gate.diff.vimax.argon}{.77}
\setR{spo.gate.diff.vimax.jury}{.76}
\setR{spo.gate.diff.vmem.argon}{.44}
\setR{spo.gate.diff.vmem.jury}{.44}
\setR{spo.gate.ident.cc.argon}{.79}
\setR{spo.gate.ident.cc.jury}{.79}
\setR{spo.gate.ident.ceil.argon}{.97}
\setR{spo.gate.ident.ceil.jury}{1.00}
\setR{spo.gate.ident.cont.argon}{.95}
\setR{spo.gate.ident.cont.jury}{.96}
\setR{spo.gate.ident.se.argon}{.75}
\setR{spo.gate.ident.se.jury}{.74}
\setR{spo.gate.ident.vimax.argon}{.78}
\setR{spo.gate.ident.vimax.jury}{.79}
\setR{spo.gate.ident.vmem.argon}{.46}
\setR{spo.gate.ident.vmem.jury}{.44}
\setR{spo.gate.same.cc.argon}{.79}
\setR{spo.gate.same.cc.jury}{.77}
\setR{spo.gate.same.ceil.argon}{.96}
\setR{spo.gate.same.ceil.jury}{1.00}
\setR{spo.gate.same.cont.argon}{.96}
\setR{spo.gate.same.cont.jury}{.97}
\setR{spo.gate.same.se.argon}{.78}
\setR{spo.gate.same.se.jury}{.78}
\setR{spo.gate.same.vimax.argon}{.79}
\setR{spo.gate.same.vimax.jury}{.81}
\setR{spo.gate.same.vmem.argon}{.47}
\setR{spo.gate.same.vmem.jury}{.45}
\setR{spo.gateacc.diff.cc.argon}{.24}
\setR{spo.gateacc.diff.cc.jury}{.27}
\setR{spo.gateacc.diff.ceil.argon}{1.00}
\setR{spo.gateacc.diff.ceil.jury}{1.00}
\setR{spo.gateacc.diff.cont.argon}{.62}
\setR{spo.gateacc.diff.cont.jury}{.69}
\setR{spo.gateacc.diff.se.argon}{.79}
\setR{spo.gateacc.diff.se.jury}{.79}
\setR{spo.gateacc.diff.vimax.argon}{.23}
\setR{spo.gateacc.diff.vimax.jury}{.17}
\setR{spo.gateacc.diff.vmem.argon}{.22}
\setR{spo.gateacc.diff.vmem.jury}{.19}
\setR{spo.gateacc.ident.cc.argon}{.39}
\setR{spo.gateacc.ident.cc.jury}{.40}
\setR{spo.gateacc.ident.ceil.argon}{1.00}
\setR{spo.gateacc.ident.ceil.jury}{1.00}
\setR{spo.gateacc.ident.cont.argon}{.66}
\setR{spo.gateacc.ident.cont.jury}{.60}
\setR{spo.gateacc.ident.se.argon}{.56}
\setR{spo.gateacc.ident.se.jury}{.51}
\setR{spo.gateacc.ident.vimax.argon}{.16}
\setR{spo.gateacc.ident.vimax.jury}{.13}
\setR{spo.gateacc.ident.vmem.argon}{.09}
\setR{spo.gateacc.ident.vmem.jury}{.07}
\setR{spo.gateacc.same.cc.argon}{.46}
\setR{spo.gateacc.same.cc.jury}{.47}
\setR{spo.gateacc.same.ceil.argon}{1.00}
\setR{spo.gateacc.same.ceil.jury}{1.00}
\setR{spo.gateacc.same.cont.argon}{.68}
\setR{spo.gateacc.same.cont.jury}{.55}
\setR{spo.gateacc.same.se.argon}{.45}
\setR{spo.gateacc.same.se.jury}{.39}
\setR{spo.gateacc.same.vimax.argon}{.12}
\setR{spo.gateacc.same.vimax.jury}{.12}
\setR{spo.gateacc.same.vmem.argon}{.02}
\setR{spo.gateacc.same.vmem.jury}{.02}
\setR{spo.ident.cc.noopus}{.31}
\setR{spo.ident.cont.noopus}{.57}
\setR{spo.ident.se.noopus}{.39}
\setR{spo.ident.vimax.noopus}{.10}
\setR{spo.ident.vmem.noopus}{.03}
\setR{spo.lead.all.argon}{21}
\setR{spo.lead.all.argon.cl.hi}{34}
\setR{spo.lead.all.argon.cl.lo}{8}
\setR{spo.lead.all.jury}{18}
\setR{spo.lead.all.jury.cl.hi}{32}
\setR{spo.lead.all.jury.cl.lo}{4}
\setR{spo.lead.all.noopus}{17}
\setR{spo.lead.all.noopus.cl.hi}{31}
\setR{spo.lead.all.noopus.cl.lo}{3}
\setR{spo.lead.bal.jury}{17}
\setR{spo.lead.bal.jury.cl.hi}{34}
\setR{spo.lead.bal.jury.cl.lo}{-2}
\setR{spo.lead.cause.argon}{17}
\setR{spo.lead.cause.argon.cl.hi}{57}
\setR{spo.lead.cause.argon.cl.lo}{-17}
\setR{spo.lead.cause.jury}{11}
\setR{spo.lead.cause.jury.cl.hi}{44}
\setR{spo.lead.cause.jury.cl.lo}{-28}
\setR{spo.lead.cause.jury.mde}{51}
\setR{spo.lead.cause.noopus}{11}
\setR{spo.lead.cause.noopus.cl.hi}{39}
\setR{spo.lead.cause.noopus.cl.lo}{-19}
\setR{spo.lead.diff.argon}{3}
\setR{spo.lead.diff.argon.cl.hi}{33}
\setR{spo.lead.diff.argon.cl.lo}{-36}
\setR{spo.lead.diff.jury}{11}
\setR{spo.lead.diff.jury.cl.hi}{38}
\setR{spo.lead.diff.jury.cl.lo}{-23}
\setR{spo.lead.diff.noopus}{3}
\setR{spo.lead.diff.noopus.cl.hi}{33}
\setR{spo.lead.diff.noopus.cl.lo}{-35}
\setR{spo.lead.ident.noopus}{18}
\setR{spo.lead.ident.noopus.cl.hi}{33}
\setR{spo.lead.ident.noopus.cl.lo}{2}
\setR{spo.lead.same.argon}{29}
\setR{spo.lead.same.argon.cl.hi}{51}
\setR{spo.lead.same.argon.cl.lo}{8}
\setR{spo.lead.same.jury}{17}
\setR{spo.lead.same.jury.cl.hi}{42}
\setR{spo.lead.same.jury.cl.lo}{-7}
\setR{spo.lead.same.jury.mde}{35}
\setR{spo.lead.same.noopus}{19}
\setR{spo.lead.same.noopus.cl.hi}{44}
\setR{spo.lead.same.noopus.cl.lo}{-5}
\setR{spo.n.all}{198}
\setR{spo.n.cause}{18}
\setR{spo.n.diff}{62}
\setR{spo.n.gates}{360}
\setR{spo.n.ident}{180}
\setR{spo.n.links}{9}
\setR{spo.n.same}{118}
\setR{spo.nodev.all.best.jury}{.40}
\setR{spo.nodev.all.cc.argon}{.35}
\setR{spo.nodev.all.cc.jury}{.35}
\setR{spo.nodev.all.cont.argon}{.66}
\setR{spo.nodev.all.cont.jury}{.59}
\setR{spo.nodev.all.se.argon}{.44}
\setR{spo.nodev.all.se.jury}{.40}
\setR{spo.nodev.all.vimax.argon}{.17}
\setR{spo.nodev.all.vimax.jury}{.14}
\setR{spo.nodev.all.vmem.argon}{.07}
\setR{spo.nodev.all.vmem.jury}{.06}
\setR{spo.nodev.cause.cc.argon}{.71}
\setR{spo.nodev.cause.cc.jury}{.57}
\setR{spo.nodev.cause.cont.argon}{.86}
\setR{spo.nodev.cause.cont.jury}{.79}
\setR{spo.nodev.cause.se.argon}{.64}
\setR{spo.nodev.cause.se.jury}{.57}
\setR{spo.nodev.cause.vimax.argon}{.71}
\setR{spo.nodev.cause.vimax.jury}{.57}
\setR{spo.nodev.cause.vmem.argon}{.36}
\setR{spo.nodev.cause.vmem.jury}{.29}
\setR{spo.nodev.diff.cc.argon}{.22}
\setR{spo.nodev.diff.cc.jury}{.26}
\setR{spo.nodev.diff.cont.argon}{.65}
\setR{spo.nodev.diff.cont.jury}{.67}
\setR{spo.nodev.diff.se.argon}{.56}
\setR{spo.nodev.diff.se.jury}{.52}
\setR{spo.nodev.diff.vimax.argon}{.19}
\setR{spo.nodev.diff.vimax.jury}{.13}
\setR{spo.nodev.diff.vmem.argon}{.11}
\setR{spo.nodev.diff.vmem.jury}{.09}
\setR{spo.nodev.ident.cc.argon}{.32}
\setR{spo.nodev.ident.cc.jury}{.33}
\setR{spo.nodev.ident.cont.argon}{.65}
\setR{spo.nodev.ident.cont.jury}{.58}
\setR{spo.nodev.ident.se.argon}{.42}
\setR{spo.nodev.ident.se.jury}{.38}
\setR{spo.nodev.ident.vimax.argon}{.12}
\setR{spo.nodev.ident.vimax.jury}{.10}
\setR{spo.nodev.ident.vmem.argon}{.04}
\setR{spo.nodev.ident.vmem.jury}{.04}
\setR{spo.nodev.lead.all.argon}{23}
\setR{spo.nodev.lead.all.argon.cl.hi}{37}
\setR{spo.nodev.lead.all.argon.cl.lo}{10}
\setR{spo.nodev.lead.all.jury}{20}
\setR{spo.nodev.lead.all.jury.cl.hi}{34}
\setR{spo.nodev.lead.all.jury.cl.lo}{5}
\setR{spo.nodev.lead.cause.argon}{14}
\setR{spo.nodev.lead.cause.argon.cl.hi}{67}
\setR{spo.nodev.lead.cause.argon.cl.lo}{-25}
\setR{spo.nodev.lead.cause.jury}{21}
\setR{spo.nodev.lead.cause.jury.cl.hi}{50}
\setR{spo.nodev.lead.cause.jury.cl.lo}{-22}
\setR{spo.nodev.lead.diff.argon}{9}
\setR{spo.nodev.lead.diff.argon.cl.hi}{39}
\setR{spo.nodev.lead.diff.argon.cl.lo}{-33}
\setR{spo.nodev.lead.diff.jury}{15}
\setR{spo.nodev.lead.diff.jury.cl.hi}{42}
\setR{spo.nodev.lead.diff.jury.cl.lo}{-23}
\setR{spo.nodev.lead.ident.argon}{23}
\setR{spo.nodev.lead.ident.argon.cl.hi}{37}
\setR{spo.nodev.lead.ident.argon.cl.lo}{9}
\setR{spo.nodev.lead.ident.jury}{19}
\setR{spo.nodev.lead.ident.jury.cl.hi}{35}
\setR{spo.nodev.lead.ident.jury.cl.lo}{3}
\setR{spo.nodev.lead.same.argon}{28}
\setR{spo.nodev.lead.same.argon.cl.hi}{51}
\setR{spo.nodev.lead.same.argon.cl.lo}{6}
\setR{spo.nodev.lead.same.jury}{16}
\setR{spo.nodev.lead.same.jury.cl.hi}{43}
\setR{spo.nodev.lead.same.jury.cl.lo}{-9}
\setR{spo.nodev.n.all}{179}
\setR{spo.nodev.n.cause}{14}
\setR{spo.nodev.n.diff}{54}
\setR{spo.nodev.n.ident}{165}
\setR{spo.nodev.n.same}{111}
\setR{spo.nodev.pairs}{169}
\setR{spo.nodev.pairs.mixed}{1}
\setR{spo.nodev.pairs.same}{108}
\setR{spo.nodev.same.cc.argon}{.37}
\setR{spo.nodev.same.cc.jury}{.37}
\setR{spo.nodev.same.cont.argon}{.65}
\setR{spo.nodev.same.cont.jury}{.53}
\setR{spo.nodev.same.se.argon}{.35}
\setR{spo.nodev.same.se.jury}{.32}
\setR{spo.nodev.same.vimax.argon}{.09}
\setR{spo.nodev.same.vimax.jury}{.09}
\setR{spo.nodev.same.vmem.argon}{.01}
\setR{spo.nodev.same.vmem.jury}{.01}
\setR{spo.pairs}{185}
\setR{spo.pairs.cause}{9}
\setR{spo.pairs.one}{172}
\setR{spo.pairs.twoobj}{4}
\setR{spo.same.cc.argon}{.36}
\setR{spo.same.cc.jury}{.36}
\setR{spo.same.cc.noopus}{.37}
\setR{spo.same.ceil.argon}{.96}
\setR{spo.same.ceil.jury}{1.00}
\setR{spo.same.cont.argon}{.65}
\setR{spo.same.cont.jury}{.53}
\setR{spo.same.cont.noopus}{.57}
\setR{spo.same.se.argon}{.35}
\setR{spo.same.se.jury}{.31}
\setR{spo.same.se.noopus}{.31}
\setR{spo.same.vimax.argon}{.09}
\setR{spo.same.vimax.jury}{.09}
\setR{spo.same.vimax.noopus}{.08}
\setR{spo.same.vmem.argon}{.01}
\setR{spo.same.vmem.jury}{.01}
\setR{spo.same.vmem.noopus}{.01}
\setR{spo.unit.cc.other.cc.jury}{.04}
\setR{spo.unit.cc.other.cont.jury}{.61}
\setR{spo.unit.cc.other.lead.argon}{61}
\setR{spo.unit.cc.other.lead.argon.cl.hi}{83}
\setR{spo.unit.cc.other.lead.argon.cl.lo}{34}
\setR{spo.unit.cc.other.lead.jury}{57}
\setR{spo.unit.cc.other.lead.jury.cl.hi}{81}
\setR{spo.unit.cc.other.lead.jury.cl.lo}{29}
\setR{spo.unit.cc.other.n}{28}
\setR{spo.unit.cc.same.cc.jury}{.47}
\setR{spo.unit.cc.same.cont.jury}{.51}
\setR{spo.unit.cc.same.n}{90}
\setR{spo.unit.se.other.cont.jury}{.55}
\setR{spo.unit.se.other.lead.argon}{57}
\setR{spo.unit.se.other.lead.argon.cl.hi}{73}
\setR{spo.unit.se.other.lead.argon.cl.lo}{40}
\setR{spo.unit.se.other.lead.jury}{49}
\setR{spo.unit.se.other.lead.jury.cl.hi}{68}
\setR{spo.unit.se.other.lead.jury.cl.lo}{30}
\setR{spo.unit.se.other.n}{69}
\setR{spo.unit.se.other.se.jury}{.06}
\setR{spo.unit.se.same.cont.jury}{.51}
\setR{spo.unit.se.same.lead.jury}{-14}
\setR{spo.unit.se.same.lead.jury.cl.hi}{10}
\setR{spo.unit.se.same.lead.jury.cl.lo}{-35}
\setR{spo.unit.se.same.n}{49}
\setR{spo.unit.se.same.se.jury}{.65}
\setR{spx.all.best.argon}{.68}
\setR{spx.all.best.jury}{.66}
\setR{spx.all.cc.argon}{.66}
\setR{spx.all.cc.jury}{.64}
\setR{spx.all.ceil.argon}{.98}
\setR{spx.all.ceil.jury}{.98}
\setR{spx.all.cont.argon}{.84}
\setR{spx.all.cont.jury}{.80}
\setR{spx.all.se.argon}{.68}
\setR{spx.all.se.jury}{.66}
\setR{spx.all.vimax.argon}{.61}
\setR{spx.all.vimax.jury}{.55}
\setR{spx.all.vmem.argon}{.45}
\setR{spx.all.vmem.jury}{.43}
\setR{spx.lead.all.argon}{15}
\setR{spx.lead.all.argon.cl.hi}{26}
\setR{spx.lead.all.argon.cl.lo}{6}
\setR{spx.lead.all.jury}{13}
\setR{spx.lead.all.jury.cl.hi}{23}
\setR{spx.lead.all.jury.cl.lo}{5}
\setR{spx.n.all}{823}
\setR{why.cc.absentempty}{.01}
\setR{why.cc.absentshots}{.33}
\setR{why.cc.contra}{.10}
\setR{why.cc.otherpart.n}{19}
\setR{why.cc.otherpart.right}{.05}
\setR{why.cc.samepart.n}{63}
\setR{why.cc.samepart.right}{.59}
\setR{why.cc.sidesnamed}{.82}
\setR{why.cc.undet}{.55}
\setR{why.cont.absentempty}{.00}
\setR{why.cont.absentshots}{.13}
\setR{why.cont.contra}{.30}
\setR{why.cont.sidesnamed}{.93}
\setR{why.cont.undet}{.48}
\setR{why.empty.cc}{14}
\setR{why.empty.cont}{0}
\setR{why.empty.se}{142}
\setR{why.empty.vimax}{40}
\setR{why.empty.vmem}{168}
\setR{why.se.absentempty}{.36}
\setR{why.se.absentshots}{.08}
\setR{why.se.contra}{.30}
\setR{why.se.otherwin.contra}{.61}
\setR{why.se.otherwin.n}{54}
\setR{why.se.otherwin.right}{.07}
\setR{why.se.otherwin4.n}{14}
\setR{why.se.otherwin4.right}{.07}
\setR{why.se.otherwin5.n}{40}
\setR{why.se.otherwin5.right}{.07}
\setR{why.se.samewin.n}{38}
\setR{why.se.samewin.right}{.82}
\setR{why.se.samewin4.n}{28}
\setR{why.se.samewin4.right}{.86}
\setR{why.se.samewin5.n}{10}
\setR{why.se.samewin5.right}{.70}
\setR{why.se.sidesnamed}{.83}
\setR{why.se.undet}{.22}
\setR{why.spo.all.cc.wrong.jury}{132}
\setR{why.spo.all.cont.wrong.jury}{82}
\setR{why.spo.all.se.wrong.jury}{118}
\setR{why.spo.all.vimax.wrong.jury}{170}
\setR{why.spo.all.vmem.wrong.jury}{188}
\setR{why.vimax.absentempty}{.08}
\setR{why.vimax.absentshots}{.19}
\setR{why.vimax.contra}{.01}
\setR{why.vimax.sidesnamed}{.89}
\setR{why.vimax.undet}{.71}
\setR{why.vmem.absentempty}{.29}
\setR{why.vmem.absentshots}{.31}
\setR{why.vmem.contra}{.02}
\setR{why.vmem.sidesnamed}{.60}
\setR{why.vmem.undet}{.37}

\usepackage[breaklinks,colorlinks,allcolors=cvprblue]{hyperref}

\def\confName{CVPR}
\def\confYear{2027}

\title{\method: Ledger-Based Planning and Request-Space Evaluation\\for Ultra-Long-Form Video Generation}

\author{Ritwik Reddy Munagala$^{1}$ \qquad Sai Krishna Mendu$^{2}$ \qquad Mahesh KMB$^{1}$\\[4pt]
$^{1}$Induce Media Labs ({\tt\small inducelabs.ai})\\
$^{2}$Independent ({\tt\small independent-researcher.org})}

\hypersetup{pdftitle={Continuo: Ledger-Based Planning and Request-Space Evaluation for Ultra-Long-Form Video Generation},
  pdfauthor={Ritwik Reddy Munagala, Sai Krishna Mendu, Mahesh KMB}}

\begin{document}

\twocolumn[{%
\renewcommand\twocolumn[1][]{#1}%
\maketitle
\ifdefined\FigStyleVThree\expandafter \fi
\def\FigStyleVThree{3}

\definecolor{cInk}{HTML}{1E1E1E}        %
\definecolor{cInk2}{HTML}{5C5C5C}       %
\definecolor{cInk3}{HTML}{8C8C8C}       %
\definecolor{cRule}{HTML}{D4D4D4}       %
\definecolor{cPanel}{HTML}{F5F5F5}      %
\definecolor{cOurs}{HTML}{1764AB}
\definecolor{cOursFill}{HTML}{E3EEF8}
\definecolor{cOursSoft}{HTML}{9DBFE3}   %
\colorlet{cBase}{cInk3}
\definecolor{cBaseFill}{HTML}{DADADA}
\definecolor{cLedger}{HTML}{E69F00}
\definecolor{cLedgerInk}{HTML}{8A5A00}
\definecolor{cLedgerFill}{HTML}{FDF2D5}
\definecolor{cLLM}{HTML}{8E44AD}
\definecolor{cLLMFill}{HTML}{F4EAF8}
\colorlet{cCode}{cInk2}
\definecolor{cCodeFill}{HTML}{ECECEC}
\definecolor{cRight}{HTML}{0B8A65}
\definecolor{cRightFill}{HTML}{E2F3EC}
\definecolor{cWrong}{HTML}{C8501E}
\definecolor{cWrongFill}{HTML}{FBE9E0}

\newcommand{\figtitle}{\sffamily\bfseries\fontsize{8}{9.4}\selectfont}   %
\newcommand{\figlabel}{\sffamily\mdseries\upshape\fontsize{7}{8.2}\selectfont}  %
\newcommand{\figlabelb}{\sffamily\bfseries\upshape\fontsize{7}{8.2}\selectfont} %
\newcommand{\fignote}{\sffamily\mdseries\upshape\fontsize{6.5}{7.5}\selectfont} %
\newcommand{\figquote}{\rmfamily\mdseries\itshape\fontsize{7}{8.2}\selectfont}  %
\newcommand{\figmono}{\fontfamily{lmtt}\fontseries{m}\fontshape{n}\fontsize{6.5}{7.5}\selectfont} %
\newcommand{\figtwo}[2]{{\figlabelb #1}\\[-0.6pt]{\fignote\color{cInk2}#2}}
\DeclareRobustCommand{\figcarry}[1]{{\bfseries\color{cLedgerInk}#1}}

\ifdefined\figboxyes\else
  \newsavebox\figboxyes\newsavebox\figboxno\newsavebox\figboxunk\newsavebox\figboxna\newsavebox\figboxdisc
  \newsavebox\figboxllm\newsavebox\figboxperson\newsavebox\figboxlock
\fi
\makeatletter
\def\fig@setboxes{%
\sbox\figboxyes{\tikz[x=1pt,y=1pt]{\fill[cRight] (0,0) circle[radius=3.4];%
  \draw[white,line width=.9pt,line cap=round,line join=round] (-1.8,-.1)--(-.5,-1.45)--(1.8,1.5);}}%
\sbox\figboxno{\tikz[x=1pt,y=1pt]{\fill[cWrong] (0,0) circle[radius=3.4];%
  \draw[white,line width=.9pt,line cap=round] (-1.35,-1.35)--(1.35,1.35) (-1.35,1.35)--(1.35,-1.35);}}%
\sbox\figboxunk{\tikz[x=1pt,y=1pt]{\path (0,0) circle[radius=3.4];\draw[cInk3,line width=.7pt] (0,0) circle[radius=3.05];%
  \node[inner sep=0pt,text=cInk3,font=\sffamily\bfseries\fontsize{5}{5}\selectfont] at (0,0) {?};}}%
\sbox\figboxna{\tikz[x=1pt,y=1pt]{\path (0,0) circle[radius=3.4];%
  \draw[cInk3,line width=.9pt,line cap=round] (-1.7,0)--(1.7,0);}}%
\sbox\figboxdisc{\tikz[x=1pt,y=1pt]{\fill[cInk] (0,0) circle[radius=4.2];}}%
\sbox\figboxllm{\tikz{\node[star,star points=4,star point ratio=2.7,fill=cLLM,inner sep=0pt,minimum size=6.8pt] {};}}%
\sbox\figboxperson{\tikz[x=1pt,y=1pt]{\fill[cInk2] (0,4.6) circle[radius=1.4];%
  \fill[cInk2] (-2.4,0)--(-2.4,1.2) arc[start angle=180,end angle=0,x radius=2.4,y radius=1.9]--(2.4,0)--cycle;}}%
\sbox\figboxlock{\tikz[x=1pt,y=1pt]{\fill[cInk,rounded corners=.5pt] (-2.3,0) rectangle (2.3,3.4);%
  \draw[cInk,line width=.8pt] (-1.35,3.4)--(-1.35,4.5) arc[start angle=180,end angle=0,radius=1.35]--(1.35,3.4);}}%
}
\ifx\@onlypreamble\@notprerr \fig@setboxes \else \AtBeginDocument{\fig@setboxes}\fi
\makeatother
\newcommand{\figmarkbox}[1]{\raisebox{-.42ex}{\resizebox{!}{1.9ex}{\usebox#1}}}
\DeclareRobustCommand{\figyes}{\figmarkbox\figboxyes}   %
\DeclareRobustCommand{\figno}{\figmarkbox\figboxno}     %
\DeclareRobustCommand{\figunk}{\figmarkbox\figboxunk}   %
\DeclareRobustCommand{\figna}{\figmarkbox\figboxna}     %
\DeclareRobustCommand{\fignum}[1]{\mbox{\rlap{\raisebox{-2.1pt}{\usebox\figboxdisc}}\makebox[8.4pt][c]{%
  \color{white}\sffamily\bfseries\upshape\fontsize{6}{6}\selectfont #1}}}
\DeclareRobustCommand{\figllmglyph}{\raisebox{-.25ex}{\resizebox{!}{1.75ex}{\usebox\figboxllm}}}
\DeclareRobustCommand{\figcodeglyph}{{\fontfamily{lmtt}\fontseries{b}\selectfont\color{cInk2}\{\,\}}}
\DeclareRobustCommand{\figpersonglyph}{\raisebox{-.1ex}{\resizebox{!}{1.75ex}{\usebox\figboxperson}}}
\DeclareRobustCommand{\figlock}{\raisebox{-.1ex}{\resizebox{!}{1.75ex}{\usebox\figboxlock}}}

\tikzset{
  fig/.style={font=\figlabel, text=cInk, line width=.5pt, line cap=round, line join=round,
    >={Stealth[length=3.8pt,width=3.2pt]}},
  box/.style={draw=cInk2, fill=white, line width=.5pt, rounded corners=2pt, align=center, inner xsep=3pt, inner ysep=2.6pt,
    text=cInk, font=\figlabel},
  llm/.style={box, draw=cLLM, fill=cLLMFill, rounded corners=3.6pt},              %
  code/.style={box, draw=cCode, fill=cCodeFill, rounded corners=.5pt},            %
  people/.style={box, draw=cInk, fill=white, rounded corners=3.6pt},              %
  ext/.style={box, draw=cInk3, fill=white, dash pattern=on 2pt off 1.3pt},        %
  panel/.style={draw=cRule, fill=cPanel, line width=.4pt, rounded corners=4pt},
  panel plain/.style={draw=cRule, fill=white, line width=.4pt, rounded corners=4pt},
  panel ours/.style={draw=cOurs, fill=cOursFill, line width=.8pt, rounded corners=4pt},
  ledger band/.style={draw=cLedger, fill=cLedgerFill, line width=1.2pt, rounded corners=3pt},
  tab/.style={fill=cInk, text=white, font=\fignote\bfseries, rounded corners=1.5pt, inner xsep=3.2pt, inner ysep=1.4pt},
  stage/.style={font=\figtitle, text=cInk, inner sep=0pt, anchor=base west},
  chip/.style={rounded rectangle, draw=cRule, fill=white, line width=.4pt, inner xsep=3pt, inner ysep=1.3pt, font=\fignote,
    text=cInk},
  fact/.style={draw=cLedger, fill=cLedgerFill, line width=.5pt, rounded corners=1.5pt, inner xsep=2.6pt, inner ysep=1.6pt,
    font=\figmono, text=cInk},                                                     %
  judge/.style={chip, draw=cLLM, fill=cLLMFill},
  flow/.style={->, draw=cInk, line width=.6pt},                                    %
  repair/.style={->, draw=cInk2, line width=.5pt, dash pattern=on 2.2pt off 1.4pt},%
  ledger io/.style={->, draw=cLedgerInk, line width=.7pt},                         %
  alabel/.style={font=\fignote, text=cInk2, fill=white, inner sep=1pt},            %
  time axis/.style={->, draw=cInk2, line width=.5pt},
  interval/.style={draw=cRule, line width=2.4pt, line cap=butt},
  interval on/.style={draw=cInk3, line width=2.4pt, line cap=butt},                %
  interval carry/.style={draw=cLedger, line width=3pt, line cap=butt},             %
  now/.style={draw=cInk, line width=.5pt, dash pattern=on 1.6pt off 1.2pt},        %
  rowbar ours/.style={draw=cOurs, line width=2pt, line cap=butt},
  rowbar base/.style={draw=cInk3, line width=2pt, line cap=butt},
  hl/.style={draw=cLedger, line width=1.1pt, rounded corners=.8pt},                %
  frame tag/.style={font=\sffamily\bfseries\fontsize{6}{6.6}\selectfont, fill=white, line width=.5pt, rounded corners=1pt,
    inner xsep=1.8pt, inner ysep=1.1pt},
  tag wrong/.style={frame tag, draw=cWrong, text=cWrong},
  tag right/.style={frame tag, draw=cRight, text=cRight},
  tag note/.style={frame tag, draw=cInk2, text=cInk2},
  missing/.style={draw=cRule, fill=cPanel, line width=.3pt, rounded corners=1.5pt, font=\fignote, text=cInk3,
    align=center, inner sep=0pt},                                                  %
}

\newcommand{\figimg}[4][]{%
  \node[inner sep=0pt,outer sep=0pt,#1] (#2) {\phantom{\includegraphics[width=#3]{#4}}};%
  \begin{scope}\clip[rounded corners=1.5pt] (#2.south west) rectangle (#2.north east);%
    \node[inner sep=0pt,outer sep=0pt] at (#2.center) {\includegraphics[width=#3]{#4}};\end{scope}%
  \draw[cRule,line width=.3pt,rounded corners=1.5pt] (#2.south west) rectangle (#2.north east);}
\newcommand{\figimgcrop}[6][]{%
  \node[inner sep=0pt,outer sep=0pt,minimum width=#3,minimum height=#4,#1] (#2) {};%
  \begin{scope}\clip[rounded corners=1.5pt] (#2.south west) rectangle (#2.north east);%
    \node[inner sep=0pt,outer sep=0pt] at (#2.center) {\includegraphics[#5]{#6}};\end{scope}%
  \draw[cRule,line width=.3pt,rounded corners=1.5pt] (#2.south west) rectangle (#2.north east);}
\newcommand{\figmissing}[5][]{\node[missing,minimum width=#3,minimum height=#4,#1] (#2) {#5};}
\newcommand{\fighl}[6][]{\draw[hl,#1] let \p1=(#2.north west), \p2=(#2.south east) in
  ({\x1+#3*(\x2-\x1)},{\y1+#4*(\y2-\y1)}) rectangle ({\x1+#5*(\x2-\x1)},{\y1+#6*(\y2-\y1)});}
\newcommand{\figtag}[3][tag wrong]{\node[#1,anchor=north east] at ([xshift=-1.6pt,yshift=-1.6pt]#2.north east) {#3};}

\begin{center}
\captionsetup{type=figure}
\newcommand{\tfstrip}[5]{%
  \xdef\tfx{#2}%
  \foreach \kind/\asp/\cont [count=\k] in {#5} {
    \pgfmathsetmacro{\w}{#4*\asp}
    \if\kind m
      \figmissing[anchor=south west, at={(\tfx,#3)}]{#1\k}{\w cm}{#4 cm}{\cont}
    \else
      \figimgcrop[anchor=south west, at={(\tfx,#3)}]{#1\k}{\w cm}{#4 cm}{height=#4cm}{fig/frames/teaser/\cont.jpg}
    \fi
    \if\kind c \fighl{#1\k}{.012}{.02}{.988}{.98}\fi
    \pgfmathsetmacro{\nx}{\tfx+\w+0.05}\xdef\tfx{\nx}
  }}
\newcommand{\tfq}[1]{{\figquote #1}}                 %
\newcommand{\tfw}[1]{{\bfseries #1}}                 %
\begin{tikzpicture}[fig, x=1cm, y=1cm]
  \path[use as bounding box] (0,0) rectangle (17.4,5.42);
  \def\ha{1.47}\def\hb{0.75}     %
  \def\ql{0.30}                  %
  \def\xa{0.05}\def\xr{15.70}    %
  \def\xe{1.90}                  %
  \pgfmathsetmacro{\xer}{\xe+4.7851*\ha+.15}   %
  \pgfmathsetmacro{\xl}{\xr-4.0384*\ha-.10}    %
  \def\xv{16.66}                 %
  \def\ry{4.63}                  %
  \def\yt{4.12}                  %
  \pgfmathsetmacro{\ya}{\yt-\ha}             %
  \pgfmathsetmacro{\yb}{\ya-\ql-.16-\hb}
  \pgfmathsetmacro{\yc}{\yb-\ql-.09-\hb}
  \newcommand{\sx}[1]{\xa+(#1-1)/(\exN-1)*(\xr-\xa)}   %
  \pgfmathsetmacro{\sI}{\sx{\exIntro}}
  \pgfmathsetmacro{\sE}{\sx{\exE}}
  \pgfmathsetmacro{\sL}{\sx{\exL}}
  \fill[cOursFill, rounded corners=3pt] (0,\ya-\ql-.04) rectangle (17.4,\yt+.06);
  \fill[cRule!40] (\sE,\ry) -- (\xe,\yt+.06) -- (\xer,\yt+.06) -- cycle;
  \fill[cRule!40] (\sL,\ry) -- (\xl,\yt+.06) -- (\xr,\yt+.06) -- cycle;
  \draw[cRule, line width=3pt, line cap=butt] (\xa,\ry) -- (\xr,\ry);
  \draw[interval carry] (\sI,\ry) -- (\sL,\ry);
  \foreach \s in \exNotches {\pgfmathsetmacro{\x}{\sx{\s}}\draw[white, line width=.5pt, line cap=butt] (\x,\ry-.07) -- (\x,\ry+.07);}
  \foreach \x in {\sI,\sE,\sL} {\fill[white] (\x,\ry) circle[radius=.09]; \fill[cInk] (\x,\ry) circle[radius=.058];}
  \node[anchor=base west, inner sep=0pt] at (\xa,5.14) {\figtitle\itshape \exFilm};
  \node[anchor=north west, inner sep=0pt] at (\xa,\ry-.12) {\fignote\color{cInk2} scene 1};
  \node[anchor=north east, inner sep=0pt] at (\xr,\ry-.12) {\fignote\color{cInk2} scene \exN};
  \node[anchor=base west, inner sep=0pt] at (\sI-.06,\ry+.19) {\figlabelb scene \exIntro\ \,{\fignote\color{cInk2}screenplay:}\ \tfq{``a 1956 Chevrolet Wagon''}};
  \node[anchor=base west, inner sep=0pt] at (\sE+.10,\ry+.19) {\figlabelb scene \exE\ \ \tfq{``the car''}};
  \node[anchor=base west, inner sep=0pt] at (\sL+.10,\ry+.19) {\figlabelb scene \exL\ \ \tfq{``her car''}};
  \node[fact, anchor=east] (fe) at (\sE+.14,5.215) {\exFactE};
  \node[fact, anchor=east] (fl) at (\sL+.14,5.215) {\exFactL};
  \draw[cLedgerInk, line width=.5pt] (\sE,\ry+.09) -- (\sE,5.215-.165);
  \draw[cLedgerInk, line width=.5pt] (\sL,\ry+.09) -- (\sL,5.215-.165);
  \node[anchor=east, inner sep=0pt] at ([xshift=-4pt]fe.west) {\fignote\bfseries\color{cLedgerInk} story ledger};
  \node[anchor=west, inner sep=0pt] at ([xshift=4pt]fl.east) {\fignote\color{cLedgerInk} written in scene \exW};
  \node[font=\fignote, text=cInk2, inner xsep=2pt, inner ysep=0pt] (gp) at ({.5*(\sE+\sL)},\ry-.215) {\exGap{} scenes, \exShots{} shots};
  \draw[cInk2, line width=.4pt, ->, >={Stealth[length=2.6pt,width=2.4pt]}] (gp.west) -- (\sE+.14,\ry-.215);
  \draw[cInk2, line width=.4pt, ->, >={Stealth[length=2.6pt,width=2.4pt]}] (gp.east) -- (\sL-.14,\ry-.215);
  \tfstrip{ae}{\xe}{\ya}{\ha}{c/1.5/w24_continuo_s77_car, i/.7467/w24_continuo_s77_jeanette, i/.7467/w24_continuo_s77_joe, i/1.7917/w24_continuo_s77_place}
  \tfstrip{al}{\xl}{\ya}{\ha}{c/1.5/w24_continuo_s112_car, i/.7467/w24_continuo_s112_joe, i/1.7917/w24_continuo_s112_place}
  \tfstrip{be}{\xe}{\yb}{\hb}{i/1.5/w24_storyengine_s77_car, i/1.7917/w24_storyengine_s77_place, i/.7467/w24_storyengine_s77_joe, i/.7467/w24_storyengine_s77_jeanette}
  \tfstrip{bl}{\xl}{\yb}{\hb}{m/1.5/{no still\\[-1pt]of the car}, i/.7467/w24_storyengine_s112_joe, i/1.7917/w24_storyengine_s112_place}
  \tfstrip{ce}{\xe}{\yc}{\hb}{m/1.5/{no still\\[-1pt]of the car}, i/.7467/w24_cinecrew_s77_joe, i/.7467/w24_cinecrew_s77_jeanette}
  \tfstrip{cl}{\xl}{\yc}{\hb}{m/1.5/{no still\\[-1pt]of the car}, i/.7467/w24_cinecrew_s112_joe, i/.7467/w24_cinecrew_s112_jeanette, i/1.7917/w24_cinecrew_s112_place}
  \foreach \y/\h/\sys/\rb in {\ya/\ha/Continuo/rowbar ours, \yb/\hb/StoryEngine/rowbar base, \yc/\hb/CineCrew/rowbar base} {
    \draw[\rb] (0.05,\y-\ql+.04) -- (0.05,\y+\h);
    \node[anchor=west, inner sep=0pt] at (0.2,\y+.5*\h) {\figlabelb \sys};
  }
  \foreach \y/\qe/\qlate in {
    \ya/{``\dots eases \figcarry{the Chevrolet wagon} to the curb beneath bare trees \dots''}/{``\dots patchy yard, \figcarry{the 1956 Chevrolet Wagon} stirs into gear.''},
    \yb/{``\dots \tfw{The sedan} slows to a stop alongside the dark suburban curb.''}/{``\dots get into \tfw{her automobile} at the curb and speed away \dots''},
    \yc/{``\dots idles \tfw{the car} outside \dots''}/{``\dots gets into \tfw{her car} and drives away \dots''}} {
    \node[anchor=north west, inner sep=0pt] at (\xe,\y-.055) {\tfq{\qe}};
    \node[anchor=north west, inner sep=0pt] at (\xl,\y-.055) {\tfq{\qlate}};
  }
  \draw[cRule, line width=.3pt] (\xr+.22,0.05) -- (\xr+.22,5.37);
  \node[anchor=base, inner sep=0pt, align=center] at (\xv,5.13) {\figlabelb Same car in\\[-1pt]\figlabelb both scenes?};
  \node[anchor=base, inner sep=0pt] at (\xv-.3,4.30) {\fontsize{6}{6}\selectfont\sffamily\color{cInk2} jury};
  \node[anchor=base, inner sep=0pt] at (\xv+.3,4.30) {\fontsize{6}{6}\selectfont\sffamily\color{cInk2} Argon};
  \foreach \y/\h/\a/\b in {\ya/\ha/\figyes/\figyes, \yb/\hb/\figunk/\figunk, \yc/\hb/\figunk/\figunk} {
    \node[inner sep=0pt] at (\xv-.3,\y+.5*\h) {\large \a};
    \node[inner sep=0pt] at (\xv+.3,\y+.5*\h) {\large \b};
  }
\end{tikzpicture}%

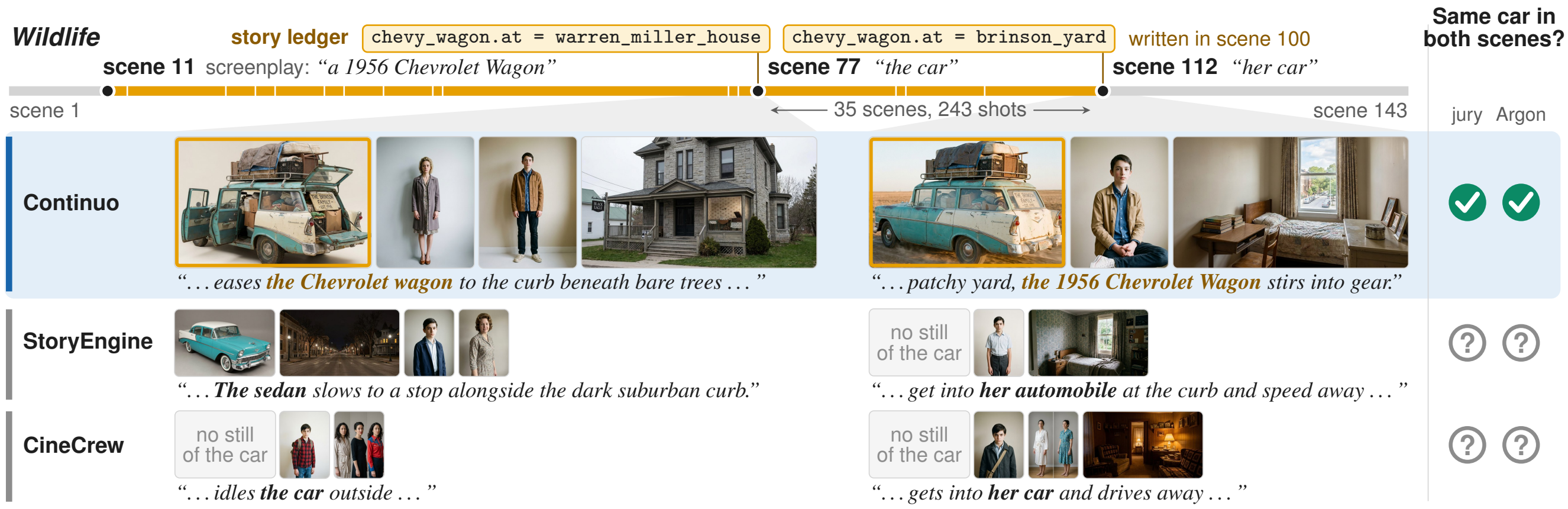
\captionof{figure}{\textbf{The same car, \exGap{} scenes later.} Top: \emph{\exFilm} with the screenplay's words for the car; amber: \ledger entries for where the wagon is. Rows: each plan's reference stills for scenes \exE{} and \exL, with one request line. Right: the pair's frozen \speobj question, answered from requests and stills (\figyes{} right, \figunk{} can't tell; ViMax, VideoMemory, not shown: \figunk). A best case by a fixed rule (\cref{sec:supp_qual}).}
\label{fig:teaser}
\end{center}

}]

\begin{abstract}
Current video generation methods mostly extend one continuous video to a few minutes or plan stories of a few dozen shots.
This leaves a gap in ultra-long-form video generation, where a feature film spans about a thousand shots over two hours and what one scene establishes must still hold many scenes later.
We introduce \method, an engine that plans and checks a shot plan for an entire film.
Its core is the planner, which takes the right information into context for each scene, including the id and name of every person and prop registered earlier, and consults the story ledger system: typed facts over story time, each naming the operation, rule or checked proposal that wrote it, with the checks and repairs that keep them consistent.
\method plans each scene from only the part of the ledger it depends on, so its ledger context is selected by dependency, not recency (the whole draft request still grows \texttimes\R{run.ctx.ratio} over a film), and it checks every planned shot against the ledger before any frame is rendered.
To measure coherence at this scale, we introduce request-space evaluation, which judges what a planner would send to the video model; it measures the plan, not the rendered film.
On \spelong (\speall), \R{spx.n.all} frozen questions on scene pairs from nine feature screenplays planned into \R{run.shots.cont} shots (\R{spo.n.all} person-designed and verified questions on objects that return in a later scene, \R{spe.n.all} on scene pairs 20 or more scenes apart), \method's shot requests are answered correctly \R{spx.all.cont.jury} of the time by a jury of three LLMs (\R{spx.all.cont.argon} by a second judge), against at most \R{spx.all.best.jury} (\R{spx.all.best.argon}) for StoryEngine, ViMax, VideoMemory and CineCrew, run plan-only through our adapters with some stages disabled, on the same planning LLM; on the object questions alone, \R{spo.all.cont.jury} (\R{spo.all.cont.argon}) against at most \R{spo.all.best.jury} (\R{spo.all.best.argon}).
If request-space fidelity carries to the screen, much of what is lost between scenes is lost in the plan.
\end{abstract}

\section{Introduction}
\label{sec:intro}

We plan a whole film for a video model.
The input is a feature screenplay; the output is a shot plan.
A \emph{scene} is one screenplay scene, the text under one heading; scenes are numbered in screenplay order, and a \emph{gap} between two scenes is counted in scenes.
Story time is when an event happens in the story; screen time is the order and length of the shots.
A \emph{shot} is one planned clip: what the video model renders in one call, one cut, as long as its action and speech take.
Its \emph{shot request} is the text a planner would send to the video model for it, and a \emph{plan} is a film's shots in screen order with their requests.
Following CineCrew~\cite{cinecrew}, we call this setting \emph{ultra-long-form} video generation and fix it by numbers: one work of an hour or more, made of hundreds to about a thousand shots; long-video methods report four to ten minutes~\cite{longlive,a2rd}.

\exStory{}
\Cref{fig:teaser} shows what three planners request for these two scenes, \exGap{} scenes apart.
At feature scale each clip can be good on its own, but the film needs all of them to agree on the facts of the story.
Four kinds of fact are easy to lose over such a distance: the indirect effects of an action (what is in a bag travels with the bag), who knows what, a setup whose payoff comes many scenes later, and errors that add up because nobody records them.

Prior work stops well short of this length.
LLM agents plan in levels, some with free-text nodes~\cite{vimax,movieagent,filmworld}, or take a segment-level storyline as given~\cite{a2rd}; systems with typed or per-entity state report 10 to 50 shots~\cite{storyengine,videomemory,entitybench} or state no plan length~\cite{cinecrew}; memory inside the video model reaches a few minutes~\cite{storymem,longlive}.
The longest plans we found, in FilmWorld's scalability test, reach 244 shots~\cite{filmworld}.
None keeps one checked record of story facts over a feature, so none can say which fact a shot relied on or where that fact came from.

\method keeps such a record, the \ledger: typed story facts over story time, each naming what wrote it and the reasons it rests on.
The ledger context a scene is planned with is its \subgraph: the facts in force as the scene opens about its people, place and props, and what they hold, contain or carry, so it is selected by dependency, not by recency.
Code checks every planned shot against the \ledger; a fault goes back to the draft, the scene's outline or an earlier scene, and what cannot be repaired is listed with the plan.

Evaluation has the matching gap.
Benchmarks score single clips~\cite{vbench,evalcrafter} or rendered multi-shot and long-form videos~\cite{msvbench,msavbench,directorbench,filmbench}, with episodes of up to 50 shots~\cite{entitybench}.
A film also changes on purpose, so similarity cannot tell a planned change from drift; questions about facts fixed in advance can, as they already do for single images~\cite{tifa,dsg}.
We introduce \reqeval: it judges a plan's shot requests, before any video is rendered, with frozen questions whose truth comes from the screenplay.
Its benchmark, \speall, asks all its questions on pairs of scenes, in two parts. \speobj asks whether an object that returns \R{spo.gap.min} to \R{spo.gap.max} scenes later (median \R{spo.gap.median}) is the very same one, and whether an earlier cause shows its effect.
Code and two LLM readings propose its answers; a person designed it and verified every answer against the screenplay (\cref{sec:speobj_proto}). \spe asks whether two scenes 20 or more scenes apart each carry the character, emotion, colour and story-state facts their text states (\cref{sec:spe}).
Four LLM judges from three providers answer both tests from the shot requests.

We run \method and four publicly released planners, StoryEngine, ViMax, VideoMemory and CineCrew~\cite{storyengine,vimax,videomemory,cinecrew}, on nine feature screenplays (\R{run.scenes} scenes; \method's plans hold \R{run.shots.cont} shots) with the same planning LLM; the baselines run plan-only from their released code through our adapters, with some of their stages disabled (\cref{sec:setup}).
On all \R{spx.n.all} \speall questions, the judges answer \R{spx.all.cont.jury} correctly from \method's shot requests under the jury, the majority of three judges, and \R{spx.all.cont.argon} under the second judge, against \R{spx.all.best.jury} and \R{spx.all.best.argon} for the best baseline (difference \R{spx.lead.all.jury} points from unrounded rates, 95\% interval over films \R{spx.lead.all.jury.cl.lo} to \R{spx.lead.all.jury.cl.hi}). On the object questions the rates are \R{spo.all.cont.jury} and \R{spo.all.cont.argon}, against \R{spo.all.best.jury} and \R{spo.all.best.argon}.
On same-object items the difference is \R{spo.lead.same.jury} points [\R{spo.lead.same.jury.cl.lo}, \R{spo.lead.same.jury.cl.hi}]; when the object returns 16 to 39 scenes later, the best baseline is right on \R{spo.gap.16to39.same.best.jury} and \method on \R{spo.gap.16to39.same.cont.jury}, though with \R{spo.gap.16to39.n.same} items the jury interval includes zero.
\Cref{sec:carry} relates the differences to three parts of the planner: every draft gets the ids of the people and props registered earlier, every scene is planned, and every shot names what it shows.
On the \R{spe.n.all} scene questions, \method scores \R{spe.all.cont.jury} (\R{spe.all.cont.argon}), against \R{spe.all.best.jury} (\R{spe.all.best.argon}) for the best baseline.

Our contributions are:
\begin{itemize}[leftmargin=1.2em,itemsep=1pt,topsep=2pt,parsep=0pt]
  \item \textbf{Planner.} \method: an outline, a film-wide registry of people and props, the \ledger with \subgraph{} context, and checked shots; every window is committed, and a check failure that repair does not clear is listed with the plan.
  \item \textbf{Evaluation.} \Reqeval: \speall (object and scene questions), four judges, a \ceiling and a video-judge noise control.
  \item \textbf{Evidence.} Plans of \R{run.shots.cont} shots from nine screenplays, four baselines on one LLM through our adapters.
\end{itemize}

\section{Related Work}
\label{sec:related}

\paragraph{Video generation.}
Film agents render with video models~\cite{veo3,wan,seedance}.
Multi-shot models keep shots coherent~\cite{lct,cinetrans,onestory}; others reach about one minute~\cite{holocine,storymem} or 240 seconds~\cite{longlive}.
Their memory is pixels or latents; \method keeps facts a person can read and check, and its requests do not depend on a particular renderer.

\paragraph{Agentic film generation.}
StoryEngine~\cite{storyengine} is closest: on ten-shot stories, code applies the effects the LLM names.
MovieAgent~\cite{movieagent}, FilmWorld~\cite{filmworld} and ViMax~\cite{vimax} plan in levels; CineCrew~\cite{cinecrew} plans in a typed film language with locked props and edits it after a critic reviews clips; VideoMemory~\cite{videomemory}, A$^2$RD~\cite{a2rd} and EntityMem~\cite{entitybench} condition shots on a memory.
As far as their papers state (\cref{tab:related}), they leave indirect effects and knowledge to the LLM.
\method's \ledger entries carry a story-time interval, a writer and reasons, rules derive indirect effects, and code checks the outline before drafting; its plans have \R{run.shots.min} to \R{run.shots.max} shots.

\paragraph{State tracking and long story planning.}
Long stories are written with chained LLM calls~\cite{dramatron}, detailed outlines~\cite{doc}, subtasks~\cite{agentwrite} or tracked state~\cite{nstagent,conwriter}, and narrative planning plans stories~\cite{ipocl,sabre,mediation}.
LLMs often miss indirect effects of actions~\cite{actionreasoningbench} and others' beliefs~\cite{fantom}.
Truth maintenance~\cite{jtms} records reasons for derived facts; incremental view maintenance~\cite{dred} withdraws facts whose reasons are withdrawn.
\method applies both to film state (story time, looks, props, knowledge) and measures a repair by the shots it touches; in our runs \R{run.noticescleared.cont} of \R{run.notices.cont} notices cleared their fault.

\paragraph{Evaluation.}
Beyond the rendered-video benchmarks of \cref{sec:intro}, FilmEval~\cite{filmworld} scores whole films.
Question answering scores image faithfulness~\cite{tifa,dsg}; LLM judges sit as juries~\cite{poll} and can favour their own output~\cite{selfpref}.
Nearest to \reqeval are StoryEngine's plan metric~\cite{storyengine} and ViMax's storyboard rating~\cite{vimax}; our questions instead ask about facts that must return in a later scene, answered from the shot requests before any video is rendered.

\section{Method: \method}
\label{sec:method}

\begin{figure*}[t]
  \centering
  \input{fig/style_v3}
\tikzset{
  interval new/.style={draw=cLedger, line width=3pt, line cap=butt},
  interval new core/.style={draw=cLedgerFill, line width=1.4pt, line cap=butt},
  win/.style={draw=cInk3, fill=cBaseFill, line width=.3pt, minimum size=.17cm, inner sep=0pt, outer sep=0pt},
  win later/.style={win, draw=cRule, fill=white, line width=.4pt},
  win this/.style={win, draw=cInk, fill=white, line width=.8pt, minimum size=.22cm},
  rowbox/.style={text width=1.49cm, minimum height=1.3cm},
  outbox/.style={text width=2.2cm, minimum height=.66cm, inner ysep=2pt},
  wlabel/.style={alabel, fill=cLedgerFill, text=cInk2, inner xsep=2pt, inner ysep=0pt},
  goes on/.style={interval on, -{Stealth[length=3.2pt,width=3.8pt]}},
  lnode/.style={draw=cLedgerInk, fill=white, line width=.5pt, rounded corners=1.5pt, inner xsep=2.2pt, inner ysep=0pt,
    minimum height=.235cm, outer sep=0pt, font=\figmono, text=cInk},
  lane/.style={draw=cInk3, line width=.3pt},
  edge on/.style={draw=cInk2, line width=.7pt, -{Stealth[length=3.2pt,width=3pt]}},
  edge off/.style={edge on, draw=cInk3, line width=.5pt},
  edge carry/.style={edge on, draw=cLedgerInk, line width=.9pt},
  edge new/.style={edge on, draw=cLedgerInk, line width=.9pt},
}
\def\ovstill#1#2#3{\IfFileExists{#3}{\figimg[anchor=west, at={#2}]{#1}{.82cm}{#3}}{\figmissing[anchor=west, at={#2}]{#1}{.82cm}{.82cm}{}}}
\begin{tikzpicture}[fig, x=1cm, y=1cm]
  \path[use as bounding box] (0,0) rectangle (17.4,5.7);
  \node[panel, anchor=south west, minimum width=3.1cm, minimum height=3.23cm] at (0,2.47) {};
  \node[panel, anchor=south west, minimum width=9.9cm, minimum height=3.23cm] at (3.3,2.47) {};
  \node[panel, anchor=south west, minimum width=4.0cm, minimum height=3.23cm] at (13.4,2.47) {};
  \node[anchor=base west, inner sep=0pt] at (0.15,5.36) {\fignum{1}\ \figtitle Outline};
  \node[anchor=base west, inner sep=0pt] at (3.45,5.36) {\fignum{2}\ \figtitle Plan and verify\ \ \fignote\color{cInk2} one window at a time, in screenplay order};
  \node[anchor=base west, inner sep=0pt] at (13.55,5.36) {\fignum{3}\ \figtitle Export\ \ \fignote\color{cInk2} for any video model};
  \node[ext, minimum height=.32cm, inner ysep=1.4pt, inner xsep=2.2pt, anchor=east] (sp) at (2.915,5.43) {\fignote Screenplay};
  \node[llm,  outbox] (film) at (1.71,4.733) {\figtwo{Film planner}{bible, colour script}};
  \node[llm,  outbox] (act)  at (1.71,3.869) {\figtwo{Act planner}{one brief per scene}};
  \node[code, outbox] (chk)  at (1.71,2.935) {\figtwo{Outline check}{briefs, in story order}};
  \foreach \n in {film,act} {\node[anchor=north east, inner sep=1.5pt] at (\n.north east) {\figllmglyph};}
  \draw[flow] (sp.south) -- (sp.south |- film.north);
  \draw[flow] (film) -- (act); \draw[flow] (act) -- (chk);
  \draw[repair] (chk.west) -- ++(-.27,0) |- (act.west);
  \path ($(act.south west)!.5!(chk.north west)$) coordinate (fy);
  \node[alabel, fill=cPanel, inner ysep=0pt, anchor=west] at (0.29,0 |- fy) {fault};
  \foreach \i in {0,...,26} {
    \pgfmathsetmacro{\x}{4.75+\i*.29}
    \ifnum\i<13 \node[win] at (\x,4.86) {};\else\ifnum\i>13 \node[win later] at (\x,4.86) {};\fi\fi
  }
  \fill[cRule!45] (8.41,4.75) -- (3.42,3.97) -- (13.08,3.97) -- (8.63,4.75) -- cycle;
  \node[win this] (wt) at (8.52,4.86) {};
  \node[anchor=south, inner sep=1.2pt] at (wt.north) {\fignote this window};
  \node[anchor=east, inner sep=0pt] at (4.56,4.86) {\fignote\color{cInk2} committed};
  \node[anchor=west, inner sep=0pt] at (12.48,4.86) {\fignote\color{cInk2} later};
  \node[code, rowbox] (sel) at (4.27,3.26)  {\figtwo{Dependency sub-graph}{facts in force}};
  \node[llm,  rowbox] (dr)  at (6.26,3.26)  {\figtwo{Drafter}{typed operations}};
  \node[code, rowbox] (kp)  at (8.25,3.26)  {\figtwo{Keeper}{eight rules}};
  \node[llm,  rowbox] (gt)  at (10.24,3.26) {\figtwo{AI gates}{extractor, proposer}};
  \node[code, rowbox] (cm)  at (12.23,3.26) {\figtwo{Commit}{facts, reasons}};
  \node[code, rowbox] (rq)  at (14.37,3.26) {\figtwo{Shot requests}{one per shot}};
  \foreach \n in {dr,gt} {\node[anchor=north east, inner sep=1.5pt] at (\n.north east) {\figllmglyph};}
  \draw[flow] (chk.east) -- (chk.east -| sel.west);
  \draw[flow] (sel) -- (dr); \draw[flow] (dr) -- (kp); \draw[flow] (gt) -- (cm); \draw[flow] (cm) -- (rq);
  \draw[flow] ([yshift=.17cm]kp.east) -- ([yshift=.17cm]gt.west);
  \draw[flow] ([yshift=-.17cm]gt.west) -- ([yshift=-.17cm]kp.east);
  \draw[repair] ([xshift=-.25cm]kp.north) -- ++(0,.18) -| node[alabel, fill=cRule!45, pos=.25] {redraft} (dr.north);
  \draw[repair] ([xshift=.25cm]kp.north) -- ++(0,.43) -- (3.2,4.34) |- (act.east);
  \node[alabel, fill=cPanel] at (4.35,4.34) {brief rewrite};
  \node[draw=cRule, fill=white, line width=.3pt, rounded corners=1.5pt, text width=3.5cm, align=left, inner sep=2.5pt, anchor=north west] (qx) at (13.55,5.12)
    {\figquote\hyphenpenalty=10000 ``\dots into the open \figcarry{black duffle bag} \dots\ closing tight around the grip of the \figcarry{captor revolver} \dots''};
  \draw[cInk2, line width=.4pt] (rq.north) -- (rq.north |- qx.south);
  \ovstill{sa}{(15.55,3.26)}{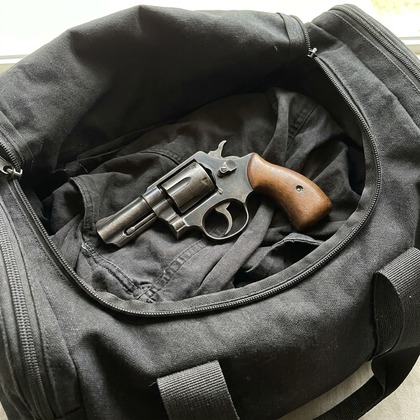}
  \ovstill{sb}{(16.43,3.26)}{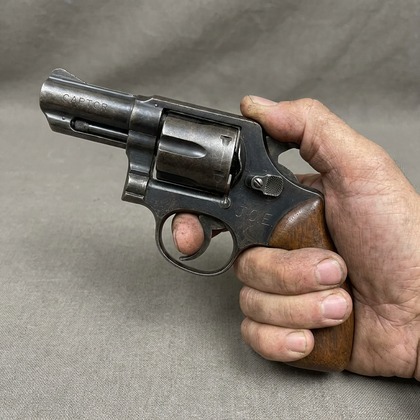}
  \draw[cInk3, line width=.5pt, dash pattern=on 2pt off 1.3pt, rounded corners=2pt] ([shift={(-.05,-.05)}]sa.south west) rectangle ([shift={(.05,.05)}]sb.north east);
  \draw[flow] (rq.east) -- (15.5,3.26);
  \node[anchor=south west, inner sep=0pt] at (15.5,3.80) {\fignote Reference stills};
  \node[anchor=north, inner sep=0pt] at ([yshift=-.1cm]$(sa.south)!.5!(sb.south)$) {\fignote\color{cInk2} scenes 2 and 28};
  \node[ledger band, anchor=south west, minimum width=17.4cm, minimum height=1.83cm] (L) at (0,0.28) {};
  \node[anchor=base west, inner sep=0pt] at (0.2,1.85) {\figtitle Story ledger\ \ \fignote\color{cLedgerInk} nodes: people, props, places; edges: typed facts};
  \foreach \n/\k/\y/\id in {a/person/1.56/ruth, b/place/1.27/station\_wagon\_interior, c/prop/0.98/black\_duffle\_bag, d/prop/0.69/captor\_revolver} {
    \node[anchor=east, inner sep=0pt] at (0.98,\y) {\fignote\color{cInk2} \k};
    \node[lnode, anchor=west] (n\n) at (1.06,\y) {\id};
    \draw[lane] (n\n.east) -- (17.14,\y);
  }
  \draw[time axis] (3.72,0.41) -- (17.2,0.41);
  \node[anchor=west, inner sep=0pt] at (0.2,0.41) {\fignote\color{cInk2} story time};
  \draw[interval] (4.25,0.69) -- (6.8,0.69);
  \draw[edge off] (4.25,0.69) -- (4.25,1.56);
  \fill[cInk3] (4.25,0.69) circle[radius=.045];
  \node[wlabel, text=cInk3] at (5.52,0.69) {{\figmono heldBy} \textperiodcentered\ scene 1};
  \draw[interval carry] (6.8,0.69) -- (12.4,0.69);
  \draw[edge carry] (6.8,0.69) -- (6.8,0.98);
  \fill[cLedgerInk] (6.8,0.69) circle[radius=.055];
  \node[wlabel, anchor=west, text=cInk] at (7.02,0.69) {{\figmono\bfseries in} \textperiodcentered\ scene 2 \textperiodcentered\ {\figmono put}};
  \draw[interval on] (8.6,0.98) -- (17.1,0.98); \draw[goes on] (17.05,0.98) -- (17.27,0.98);
  \draw[edge on] (8.6,0.98) -- (8.6,1.27);
  \fill[cInk2] (8.6,0.98) circle[radius=.05];
  \node[wlabel, anchor=west] at (8.82,0.98) {{\figmono at} \textperiodcentered\ scene 26 \textperiodcentered\ {\figmono put}};
  \draw[interval on] (11.2,1.56) -- (15.5,1.56);
  \draw[cInk2, line width=.6pt] (15.5,1.50) -- (15.5,1.62);
  \draw[edge on] (11.2,1.56) -- (11.2,1.27);
  \fill[cInk2] (11.2,1.56) circle[radius=.05];
  \node[wlabel, anchor=west] at (12.62,1.56) {{\figmono at} \textperiodcentered\ scene 28 \textperiodcentered\ {\figmono presence}};
  \draw[interval new] (12.4,0.69) -- (16.6,0.69); \draw[interval new core] (12.428,0.69) -- (16.572,0.69);
  \draw[cLedgerFill, line width=2.4pt, line cap=butt] (12.4,0.80) -- (12.4,1.44);
  \draw[edge new] (12.4,0.69) -- (12.4,1.56);
  \fill[cLedgerInk] (12.4,0.69) circle[radius=.055];
  \node[wlabel, anchor=west, text=cInk] at (12.62,0.69) {{\figmono\bfseries heldBy} \textperiodcentered\ this window \textperiodcentered\ {\figmono take}};
  \draw[now] (12.05,0.41) -- (12.05,2.0);
  \foreach \y in {0.69,0.98,1.56} {\node[circle, draw=cInk, fill=white, line width=.5pt, minimum size=.13cm, inner sep=0pt] at (12.05,\y) {};}
  \node[anchor=base east, inner sep=0pt] at (11.96,1.85) {\fignote dependency sub-graph: the edges in force};
  \node[anchor=base west, inner sep=0pt] at (12.14,1.85) {\fignote as this window opens};
  \draw[ledger io, <->] (0.82,0 |- L.north) -- (0.82,0 |- chk.south);
  \node[alabel, anchor=west] at (0.92,2.29) {briefs};
  \draw[ledger io] (sel.south |- L.north) -- (sel.south);
  \node[alabel, anchor=west] at (4.37,2.29) {read};
  \draw[repair] ([xshift=-.3cm]kp.south) -- ([xshift=-.3cm]kp.south |- L.north);
  \node[alabel, anchor=west] at (8.05,2.29) {notice to the fact's writer};
  \draw[ledger io] (cm.south) -- (cm.south |- L.north);
  \node[alabel, anchor=west] at (12.33,2.29) {write};
  \draw[ledger io] (rq.south |- L.north) -- (rq.south);
  \node[alabel, anchor=west] at (14.47,2.29) {looks and states};
  \node[llm,  minimum width=.42cm, minimum height=.19cm, inner sep=0pt] (k1) at (0.26,0.11) {};
  \node[anchor=west, inner sep=0pt] (k1t) at ([xshift=.1cm]k1.east) {\fignote LLM call};
  \node[code, minimum width=.42cm, minimum height=.19cm, inner sep=0pt, anchor=west] (k2) at ([xshift=.4cm]k1t.east) {};
  \node[anchor=west, inner sep=0pt] (k2t) at ([xshift=.1cm]k2.east) {\fignote code};
  \node[ext,  minimum width=.42cm, minimum height=.19cm, inner sep=0pt, anchor=west] (k3) at ([xshift=.4cm]k2t.east) {};
  \node[anchor=west, inner sep=0pt] (k3t) at ([xshift=.1cm]k3.east) {\fignote input};
  \draw[ledger io] ([xshift=.4cm]k3t.east) -- ++(.45,0) coordinate (k4);
  \node[anchor=west, inner sep=0pt] (k4t) at ([xshift=.1cm]k4) {\fignote ledger read or write};
  \draw[repair] ([xshift=.4cm]k4t.east) -- ++(.45,0) coordinate (k5);
  \node[anchor=west, inner sep=0pt] at ([xshift=.1cm]k5) {\fignote repair path};
\end{tikzpicture}%
  \caption{\textbf{\method{} in one picture.}
  (1)~Outline: a bible and one brief per scene, checked in story order.
  (2)~Each window is drafted from its \subgraph, checked by the keeper and two AI gates, and committed.
  (3)~One request per shot.
  All stages share the \ledger (amber), drawn as a graph: people, props and places are nodes, typed facts are edges over story time, and the edges in force as the window opens form its \subgraph.
  Example (\emph{Fast Color}): a revolver stowed in scene~2 is still in the bag when scene~28 takes it out.}
  \label{fig:overview}
\end{figure*}

\subsection{Overview and outline}
\label{sec:outline}
The input is a screenplay with scenes $s_1,\dots,s_N$; the output is a plan and the \ledger that justifies it.
\method plans at five levels: film, three acts, one brief per scene, windows and shots (\cref{fig:overview}).
The method has two parts.
The \emph{planner} reads the screenplay into an outline, cuts it into windows, keeps a film-wide registry that gives every draft the id and name of each person and prop introduced earlier, drafts natural-length shots, commits every window, listing what it could not fix, and exports one request per shot.
The \emph{\ledger system} records what holds over story time, with the writer of each fact, plans each window from the part of it the window depends on, and checks and repairs every shot against it (\cref{sec:ledger,sec:draft,sec:verify}).

\emph{Outline.}
One film call reads the whole screenplay and writes the \emph{bible}: people with their aliases and looks, places, props, information with its first knowers, starting facts and three acts.
It also writes a \emph{colour script}: palettes and turns in story order.
Act calls, twelve scenes at a time, then write one typed \emph{brief} per scene: place, story time, any declared time jump, cast and props; the facts that hold as it opens and closes; what it needs from earlier scenes and delivers to later ones; look changes; and the people it introduces.

\emph{Scene parts and windows.}
A scene that changes place or time under one heading (a montage, a phone call cut between rooms) declares \emph{scene parts}, and a \emph{window}, what the drafter plans in one call, is one scene part or, for a part longer than 1,746 characters of screenplay, a run of whole paragraphs of it, never cutting a speech from its cue.

\emph{Outline check.}
Before drafting, code checks the briefs in story order: each opening fact follows from the bible, an earlier brief or a declared jump; each need has a producer; props reach the scenes that use them; nobody is in two places at once; and story time runs forward unless a jump is declared.
A fault goes back to its brief, which the act planner rewrites (at most three rounds; drafting starts from the round with the fewest faults).
Rewrites are corrective: the screenplay wins, and code drops any added person, prop or off-screen change whose quote is not in the screenplay.

\emph{Outline graph.}
Each brief is written into the \ledger at its story time, with the brief as writer; act calls, rewrites and the outline check read earlier briefs from these entries, so a flashback written later does not overwrite the present.

\subsection{The \ledger}
\label{sec:ledger}
\emph{Entries.}
An entry is a fact $f=(e,a,v,[t_0,t_1))$: entity $e$ has value $v$ for attribute $a$ over an interval of story time (\cref{fig:ledger}).
There are ten attributes: \textsf{at}, \textsf{heldBy}, \textsf{in}, \textsf{aboard}, \textsf{look}, \textsf{state}, \textsf{condition}, \textsf{knows}, \textsf{emotion} and \textsf{tension}.
\textsf{knows} and \textsf{condition} are sets; the others hold one value at a time, and \textsf{at}, \textsf{heldBy} and \textsf{in} share one slot.
Each entry names its writer (window, shot and kind of writer) and, except for the bible's starting facts, its reasons: the operation, rule or gate answer that produced it and the earlier entries it rests on, as in a justification-based truth maintenance system~\cite{jtms}.
The \ledger stores state over story time, not causal links between events; what one scene needs from another travels in the briefs' needs and delivers.

\emph{Who writes.}
The drafter attaches typed operations to each shot (16 in five kinds: movement, objects, appearance, perception and feeling, speech and knowledge; \cref{tab:ops}), whose preconditions code checks against the \ledger and whose effects it records, as for the extractor's operations (\cref{sec:verify}); the outline's declarations (starting facts, what each brief sets as its scene opens, the backstory of anyone introduced) and the proposer's answers are recorded with their writers.
Rules add what an operation implies but does not state, such as contents travelling with their container and a line said aloud reaching everyone present, each with its rule as writer (\cref{sec:supp_method}).

\emph{Story time.}
Intervals are in story time, not screen time, so a flashback writes into the past (its fact holds until the next fact in the same slot), and a declared time jump lets a brief reset placement, look, state, condition and feelings as its scene opens, but never knowledge.
An operation can close an entry's interval but never deletes it, a repair to an earlier window replays the windows after it, and every fact names its writer: the nine plans' ledgers hold \R{run.facts} facts, \R{run.factswriter.pct}\% of them written by a window and the rest by the bible, and \R{run.factswithreason.pct}\% with a recorded reason.

\subsection{Planning from a \subgraph}
\label{sec:draft}
\emph{The \subgraph.}
For each window, code selects from the \ledger the entries in force as the window opens about its people and place, the props of its brief, anyone the scene's earlier windows introduced and any changed prop lying at the place, together with what these hold, contain or carry.
The selection follows one link and is not cut to a length; each entry carries the window that wrote it, and the drafter gets it with the brief, the bible rows, the ids and names of every person and prop registered earlier in the film, the colour script, where earlier shots left props, what later scenes rely on, the established details (below), the scene so far and the surrounding screenplay text (\cref{sec:supp_method}).

\emph{Natural-length shots.}
The drafter returns as many shots as the window's text needs, each with its camera setup, action, blocking, cast on screen, dialogue, operations, colour, light, length and the source events it shows, quoted from the window.
Code sets a shot's length to the larger of the length the drafter states and its spoken words at 2.8 words a second; there is no fixed shot count, shot length or word limit, and every speech is spoken whole and in order.

\emph{Continuity rows.}
Operations record where things are and who knows what, not how a wound or a dressing looks.
So the extractor (\cref{sec:verify}) also lists each lasting physical state the accepted shots show (an injury and its side, a dressing, clothes, a broken prop, a damaged place); code writes these as continuity rows of the \ledger at their shots' story times, and later windows see them as established details.
A shot that contradicts one with nothing in the window to change it gets one redraft; if the difference remains, it is listed as an open issue.

\subsection{Verification and repair}
\label{sec:verify}
\emph{Keeper.}
The keeper (code) replays a draft's operations on a copy of the \ledger and checks eight rules (\cref{sec:supp_method}): operation preconditions; the brief's opening and closing facts; the forward constraint (nothing a later brief needs is removed unless a scene in between restores it); source quotes; speech; cast and registry; camera and light; and the feelings the brief requires.

\emph{AI gates.}
The extractor (an LLM) reads a passing draft against the window text and lists changes no operation records; code adds them as operations and reruns the keeper, with one redraft if they break a rule.
The proposer (an LLM) answers what rules cannot decide, such as the cause of a condition no scene shows; an answer below confidence 0.3 is not recorded.

\emph{Repair.}
The keeper's reasons go back to the drafter for one redraft; if the fault rests on facts an earlier window wrote, code follows the reasons to the windows that wrote those facts and sends a notice to the one whose change touches the fewest shots, which may approve, adjust or reject it, the change being replayed through the windows after it and reverted if any of them breaks (one notice per ten windows).
If the fault remains, the act planner rewrites the scene's brief once to agree with its scene text, checked against the committed \ledger before redrafting.

A window that still fails is committed with its open issues listed, so the plan completes.
On the nine plans, \R{run.clean.cont} of \R{run.windows.cont} windows were committed clean, \R{run.openissues.cont} with open issues listed, and \R{run.failed.cont} could not be drafted; \R{run.notices.cont} notices were sent and \R{run.noticescleared.cont} cleared their fault.

\subsection{From plan to shot requests}
\label{sec:export}
Each shot becomes one plan entry, with the looks and states the \ledger gives its people, place and props.
Code joins its action, blocking and dialogue into one request text per shot; its camera, light and colour lines stay in the plan.
Each registered person and prop has one description and one still, so shots that reuse an id attach the same still.
The request does not depend on the renderer; it is what \reqeval judges (\cref{sec:eval}).
\begin{figure*}[t]
  \centering
  \input{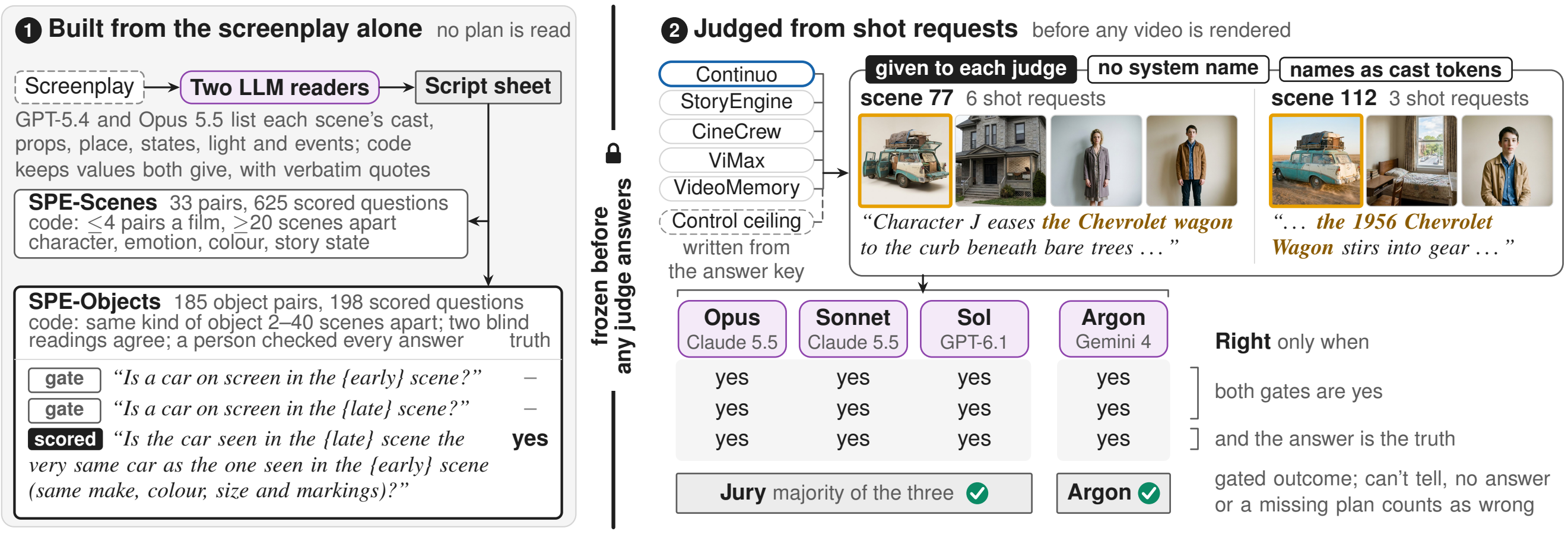}
\providecommand{\pxObj}{car}                           %
\providecommand{\pxE}{77}\providecommand{\pxL}{112}   %
\providecommand{\pxNE}{6}\providecommand{\pxNL}{3}     %
\providecommand{\pxdir}{fig/frames/protocol}           %
\begin{tikzpicture}[fig, x=1cm, y=1cm,
    pstep/.style={minimum height=.36cm, inner ysep=0pt, font=\figlabelb},
    ans/.style={font=\figlabel, text=cInk, inner sep=0pt, anchor=base},
    qgate/.style={tag note, anchor=base west, minimum width=.8cm},
    qscored/.style={frame tag, draw=cInk, fill=cInk, text=white, anchor=base west, minimum width=.8cm},
    plan/.style={chip, minimum width=1.86cm, minimum height=.27cm, inner xsep=2pt, inner ysep=0pt},
    jhead/.style={llm, inner xsep=1.5pt, minimum width=1.2cm, minimum height=.54cm},
    tab open/.style={tab, fill=white, text=cInk, draw=cInk, line width=.5pt},
  ]
  \path[use as bounding box] (0,0) rectangle (17.4,5.8);
  \def\yT{5.46}                                  %
  \def\yHd{2.21}                                 %
  \def\yGa{1.59}\def\yGb{1.25}\def\ySc{0.91}     %

  \node[panel, anchor=south west, minimum width=6.4cm, minimum height=5.8cm] at (0,0) {};
  \node[anchor=base west, inner sep=0pt] at (0.15,\yT)
    {\fignum{1}\ \figtitle Built from the screenplay alone\ \ \fignote\color{cInk2} no plan is read};
  \node[ext,  pstep, font=\figlabel, anchor=west] (sp) at (0.15,4.90) {Screenplay};
  \node[code, pstep, anchor=east] (sh) at (6.25,4.90) {Script sheet};
  \node[llm,  pstep] (rd) at ($(sp.east)!.5!(sh.west)$) {Two LLM readers};
  \draw[flow] (sp) -- (rd);
  \draw[flow] (rd) -- (sh);
  \draw[cInk2, fill=white, line width=.5pt, rounded corners=2pt] (0.15,3.00) rectangle (5.20,3.76);
  \draw[cInk, fill=white, line width=.8pt, rounded corners=2pt] (0.15,0.10) rectangle (6.25,2.68);
  \draw[flow] (sh.south) -- (sh.south |- 0,2.68);
  \draw[flow] (sh.south |- 0,3.41) -- (5.20,3.41);
  \node[anchor=north west, inner sep=0pt, align=left, text width=5.1cm] at (0.15,4.64) {\fignote\color{cInk2} GPT-5.4 and Opus~5.5 list each scene's cast,\\ props, place, states, light and events; code\\ keeps values both give, with verbatim quotes};
  \node[anchor=base west, inner sep=0pt] at (0.30,3.54)
    {{\figlabelb \spe}\ \ {\fignote\color{cInk2} \R{spe.pairs} pairs, \R{spe.n.all} scored questions}};
  \node[anchor=base west, inner sep=0pt] at (0.30,3.32) {\fignote\color{cInk2} code: $\le$4 pairs a film, $\ge$20 scenes apart};
  \node[anchor=base west, inner sep=0pt] at (0.30,3.10) {\fignote\color{cInk2} character, emotion, colour, story state};
  \node[anchor=base west, inner sep=0pt] (so) at (0.30,2.43)
    {{\figlabelb \speobj}\ \ {\fignote\color{cInk2} \R{spo.pairs} object pairs, \R{spo.n.all} scored questions}};
  \node[anchor=base west, inner sep=0pt] at (0.30,2.21) {\fignote\color{cInk2} code: same kind of object 2--40 scenes apart; two blind}; 
  \node[anchor=base west, inner sep=0pt] at (0.30,2.01) {\fignote\color{cInk2} readings agree; a person checked every answer};
  \node[anchor=base, inner sep=0pt] at (5.90,2.01) {\fignote\color{cInk2} truth};
  \draw[cRule, line width=.3pt] (0.15,1.87) -- (6.25,1.87);
  \node[qgate] at (0.30,\yGa) {gate};
  \node[anchor=base west, inner sep=0pt] at (1.22,\yGa) {\figquote ``Is a \pxObj{} on screen in the \{early\} scene?''};
  \node[ans, text=cInk3] at (5.90,\yGa) {--};   %
  \node[qgate] at (0.30,\yGb) {gate};
  \node[anchor=base west, inner sep=0pt] at (1.22,\yGb) {\figquote ``Is a \pxObj{} on screen in the \{late\} scene?''};
  \node[ans, text=cInk3] at (5.90,\yGb) {--};
  \node[qscored] at (0.30,\ySc) {scored};
  \node[anchor=base west, inner sep=0pt, align=left, text width=5.25cm] at (0.30,\ySc)
    {\hspace*{0.92cm}\figquote ``Is the \pxObj{} seen in the \{late\} scene the very same \pxObj{} as the one seen in the \{early\} scene (same make, colour, size and markings)?''};
  \node[ans, font=\figlabelb] at (5.90,\ySc) {yes};   %

  \draw[cInk, line width=1.2pt] (6.825,0) -- (6.825,5.8);
  \fill[white] (6.52,1.52) rectangle (7.13,4.42);
  \node[inner sep=0pt, rotate=90, align=center] (fz) at (6.825,2.80)
    {\figlabelb frozen before\\[-0.8pt]\figlabelb any judge answers};
  \node[inner sep=0pt, anchor=south] at ([yshift=4pt]fz.east) {\figlabelb\figlock};

  \node[anchor=base west, inner sep=0pt] at (7.35,\yT)
    {\fignum{2}\ \figtitle Judged from shot requests\ \ \fignote\color{cInk2} before any video is rendered};
  \node[plan, draw=cOurs, line width=.8pt] (p1) at (8.20,5.05) {\method};
  \node[plan] (p2) at (8.20,4.73) {StoryEngine};
  \node[plan] (p3) at (8.20,4.41) {CineCrew};
  \node[plan] (p4) at (8.20,4.09) {ViMax};
  \node[plan] (p5) at (8.20,3.77) {VideoMemory};
  \node[plan, draw=cInk3, dash pattern=on 2pt off 1.3pt] (p6) at (8.20,3.40) {\Ceiling};
  \node[anchor=north, inner sep=0pt, align=center] at (8.20,3.22)
    {\fignote\color{cInk2} written from\\[-0.6pt]\fignote\color{cInk2} the answer key};
  \draw[cInk2, line width=.5pt] (p1.east) -- ++(.09,0) |- (p5.east);
  \foreach \p in {p2,p3,p4} \draw[cInk2, line width=.5pt] (\p.east) -- ++(.09,0);
  \draw[repair, -] (p6.east) -- ++(.09,0) -- ([xshift=.09cm]p5.east);   %
  \node[panel plain, draw=cInk2, line width=.5pt, anchor=north west, minimum width=7.95cm, minimum height=2.29cm] (U) at (9.45,5.10) {};
  \draw[flow] ([xshift=.09cm]p1.east |- U.west) -- (U.west);
  \node[tab, anchor=west] (tg) at ([xshift=5pt]U.north west) {given to each judge};
  \node[tab open, anchor=west] (tn) at ([xshift=3pt]tg.east) {no system name};
  \node[tab open, anchor=west] at ([xshift=3pt]tn.east) {names as cast tokens};
  \node[anchor=base west, inner sep=0pt] at (9.57,4.70) {{\figlabelb scene \pxE}\ \ {\fignote\color{cInk2} \pxNE{} shot requests}};
  \foreach \i/\f in {0/prop-chevy_wagon-from74, 1/place-warren_miller_house-base, 2/character-jeanette_brinson-from74, 3/character-joe_brinson-from72}
    {\figimgcrop[anchor=north west, at={(9.57+\i*1.065,4.59)}]{e\i}{1.015cm}{1.015cm}{width=1.015cm}{\pxdir/\f.jpg}}
  \node[anchor=north west, inner sep=0pt, align=left, text width=4.2cm] at (9.57,3.505)
    {\figquote ``Character J eases \figcarry{the \mbox{Chevrolet} wagon} to the curb beneath bare trees~\dots''};
  \draw[cRule, line width=.3pt] (13.97,2.94) -- (13.97,4.88);
  \fighl{e0}{0}{0}{1}{1}
  \node[anchor=base west, inner sep=0pt] at (14.15,4.70) {{\figlabelb scene \pxL}\ \ {\fignote\color{cInk2} \pxNL{} shot requests}};
  \foreach \i/\f in {0/prop-chevy_wagon-from99, 1/place-brinson_bedroom_joe-base, 2/character-joe_brinson-from103}
    {\figimgcrop[anchor=north west, at={(14.15+\i*1.065,4.59)}]{l\i}{1.015cm}{1.015cm}{width=1.015cm}{\pxdir/\f.jpg}}
  \node[anchor=north west, inner sep=0pt, align=left, text width=3.15cm] at (14.15,3.505)
    {\figquote ``\dots\ \figcarry{the 1956 \mbox{Chevrolet} Wagon} stirs into gear~\dots''};
  \fighl{l0}{0}{0}{1}{1}

  \foreach \i/\x/\n/\m in {1/8.15/Opus/Claude 5.5, 2/9.50/Sonnet/Claude 5.5, 3/10.85/Sol/GPT-6.1, 4/12.40/Argon/Gemini 4}
    {\node[jhead] (j\i) at (\x,\yHd) {{\figlabelb \n}\\[-0.8pt]{\sffamily\fontsize{6}{6.6}\selectfont\color{cInk2}\m}};}
  \draw[cInk2, line width=.5pt] ([yshift=.04cm]j1.north west) -- ++(0,.07) -| ([yshift=.04cm]j4.north east);
  \draw[flow] (10.275,2.81) -- (10.275,2.60);
  \fill[cPanel, rounded corners=2pt] (7.52,0.74) rectangle (11.48,1.87);
  \fill[cPanel, rounded corners=2pt] (11.77,0.74) rectangle (13.03,1.87);
  \foreach \x/\ga/\gb/\sc in {8.15/yes/yes/yes, 9.50/yes/yes/yes, 10.85/yes/yes/yes, 12.40/yes/yes/yes} {
    \node[ans] at (\x,\yGa) {\ga}; \node[ans] at (\x,\yGb) {\gb}; \node[ans] at (\x,\ySc) {\sc};
  }
  \node[code, minimum height=.44cm, anchor=west, minimum width=3.96cm] at (7.52,0.38) {{\figlabelb Jury}\ {\fignote\color{cInk2} majority of the three}\ \ \figyes};
  \node[code, minimum height=.44cm, minimum width=1.26cm] at (12.40,0.38) {{\figlabelb Argon}\ \figyes};
  \node[anchor=base west, inner sep=0pt] at (13.52,1.99) {{\figlabelb Right}\ {\fignote\color{cInk2} only when}};
  \draw[cInk2, line width=.5pt] (13.26,1.78) -- ++(.08,0) -- (13.34,1.21) -- ++(-.08,0);
  \node[anchor=west, inner sep=0pt] at (13.52,1.495) {\fignote\color{cInk2} both gates are yes};
  \draw[cInk2, line width=.5pt] (13.26,1.10) -- ++(.08,0) -- (13.34,0.87) -- ++(-.08,0);
  \node[anchor=base west, inner sep=0pt] at (13.52,\ySc) {\fignote\color{cInk2} and the answer is the truth};
  \node[anchor=west, inner sep=0pt, align=left, text width=3.8cm] at (13.52,0.38) {\fignote\color{cInk2} gated outcome; can't tell, no answer\\ or a missing plan counts as wrong};
\end{tikzpicture}%
  \caption{\textbf{\Reqeval, on one \speobj item.} Left of the line: truth, pairs and questions come from the screenplay; their builders read no plan, and each set is frozen before any judge answers it. Right: each plan's shot requests and reference stills for the two scenes go, with no system name, to four judges; the scored answer counts only if both presence gates are yes. Shown: \method's plan for the pair of \cref{fig:teaser}.}
  \label{fig:protocol}
\end{figure*}
\section{Request-Space Evaluation}
\label{sec:eval}

\Reqeval judges a plan's shot requests before anything is rendered; its benchmark, \speall (\spelong), asks every question on a pair of scenes, in the two parts of \cref{sec:speobj_proto,sec:spe}, answered by LLM judges (\cref{sec:judges}, \cref{fig:protocol}).

\subsection{What is judged, and why before rendering}
\label{sec:reqeval}
Every plan is mapped to one schema: per shot, the request a video model would receive (the system's own prompt, or a shared assembly of its fields), plus reference stills drawn by one image model from that system's own descriptions.
The same code turns each plan's character names into neutral cast tokens.
We judge requests because they separate planning from rendering failures and need no video.
A \emph{probe scene} is a scene that some question asks about.

\emph{Script sheets.}
Both question sets start from one \emph{script sheet} per scene.
Two LLM readers from two model families, GPT-5.4 and Claude Opus~5.5, list each scene's cast, props, place, states, light and events.
Code keeps only values the two agree on, with quotes found verbatim in the scene (\cref{sec:supp_eval}).

\subsection{\speobj: object continuity}
\label{sec:speobj_proto}
\speobj asks whether a thing seen early is the same thing when it returns, and whether it is in the state an earlier event left it in.

\emph{Object pairs.}
An \emph{object pair} is two scenes, 2--40 scenes apart, in which one kind of object appears with a character on screen.
Code proposes candidates from agreed sheet values: consecutive appearances of one kind of object (head noun, synonyms folded), set dressing excluded.
Two blind readings by Claude Opus~5.5 of the two scenes' screenplay text propose whether it is the same object or a different one; a pair is kept only when both readings agree and every quote they give is found verbatim, and a person checked every answer against the screenplay.
Every agreed object is kept, with no cap; with the cause-effect links below, the set has \R{spo.pairs} scene pairs, \R{spo.gap.min}--\R{spo.gap.max} scenes apart.
Most object pairs are short (median \R{spo.gap.median} scenes); \spe covers gaps of 20 scenes or more.

\emph{Questions.}
Each kept object asks one scored identity question: is the [object] in the later scene the very same one as in the earlier scene (same make, colour, size and markings)?
Its wording names no owner, description or quote.
For the \R{spo.n.same} \emph{same-object} items the truth is yes.
The \R{spo.n.diff} \emph{different-object} items are controls with truth no, so a plan that gives every phone one identity is not rewarded.
Two unscored \emph{presence gates} per object ask whether an [object] is on screen in each scene.
The set adds \R{spo.n.cause} \emph{cause-effect} items: \R{spo.n.links} links in which the earlier scene shows a cause and the later scene its effect, each named by an object.
In all the \R{spo.pairs} pairs carry \R{spo.n.all} scored questions, since a pair can carry more than one, and \R{spo.n.gates} gates (\cref{sec:supp_objects}).

\emph{Provenance.}
A person designed the set and verified every scored answer against the screenplay; with one person, no agreement between people is measured.
Its pairs and truths come from the script sheets and the screenplay text; no plan, ledger or answer was read in building it, and it was frozen before any final plan was judged.
It is a targeted test of carried object state, not a general measure and not a test of causal reasoning.

\emph{Scoring.}
An identity answer counts only when the judge also answers yes to both gates.
Gate pass rates per system are in \cref{tab:supp_gates}.
Same-object and different-object items are reported apart; \emph{All} is the mean over all \R{spo.n.all}.

\subsection{\spe: scene pairs}
\label{sec:spe}
\spe tests per-scene fidelity at a distance.
A \emph{pair} is two probe scenes at least 20 scenes apart that share a principal character, prop or place.
Code picks up to four scene-disjoint pairs per screenplay, preferring pairs whose questions need both scenes.
Pairs and questions were fixed and hashed before \method's final plans, and no planner is told which scenes are probed.
Code writes yes/no and closed-choice questions from the sheets in four \emph{dynamics} (\cref{tab:factors}): character (how two people act toward each other), emotion (its strength, and which scene is stronger), colour (warmth, and whether the light differs) and story state (identity, wardrobe, place state, props).
Every question offers ``can't tell''.
Distractors ask about an event of another probe scene and are not scored.
A fifth, causal dynamic of 44 questions is not scored: cause and effect are each stated in their own scene's text, so it repeats per-scene fidelity; the pairs stay as frozen.
The set has \R{spe.pairs} pairs from nine screenplays, \R{spe.gap.min}--\R{spe.gap.max} scenes apart (median \R{spe.gap.median}), with \R{spe.n.all} scored questions (character \R{spe.n.char}, emotion \R{spe.n.emo}, colour \R{spe.n.col}, story state \R{spe.n.state}).

\subsection{\Planaudit}
\label{sec:audit_proto}
The \planaudit reads every system's final plan against its screenplay: in the \emph{whole-plan reading} Gemini~4 Argon scores nine criteria from 1 to 10 with code-verified quotes (three readings per film for \method, one per film for each baseline), and the \emph{plan checks} compare the plan with the script sheets (\cref{sec:supp_audit}).

\subsection{Controls}
\label{sec:controls}
The \emph{\ceiling} is a plan of the probe scenes that code writes from the answer key, with no LLM, stating every scored answer in its shot text.
It shares the naming rule, stills, request wrapper, judges, question order, seed and scoring of every system.
A free audit confirms that its text states every scored answer of both sets.
Its score is what the instruments recover when every answer is stated in the requests, so one minus it bounds the judges' misses when every answer is stated explicitly; it does not estimate false positives on realistic plans.

\subsection{Judges and scoring}
\label{sec:judges}
A \emph{judge} is one LLM that answers the frozen questions from the requests and stills of two scenes, with a cast key from the plan's character tokens to names and no system name (\cref{sec:supp_tokens}).
\speobj and \spe use the same prompt and the same four judges.
The \emph{jury} is the majority of Claude Opus~5.5, Claude Sonnet~5.5 and GPT-6.1 Sol, a three-judge panel as in~\cite{poll}, though two of the three share a family.
The \emph{second judge}, Gemini~4 Argon, comes from the planners' provider and is reported apart, since a judge can favour its own model's output~\cite{selfpref}.
Each judge's own scores are in \cref{sec:supp_judges}; since Opus~5.5 also helps build the truth of both sets and sits on the jury, it also reports the majority of Sonnet~5.5, Sol and Argon.
A wrong answer, ``can't tell'', no answer and a missing plan count as wrong.

\section{Experiments}
\label{sec:exp}

\subsection{Setup}
\label{sec:setup}

We run \method once on the \R{run.scenes} scenes of nine lesser-known feature screenplays, chosen to limit recall of the plot, which we did not test (titles in \cref{sec:supp_baselines}).
It plans \R{run.shots.cont} shots in \R{run.windows.cont} windows, \R{run.minutes.cont} minutes on screen (\R{run.shotsperscene} shots a scene, so \spe pairs lie at least about \R{run.shots20} shots apart); each shot lasts as long as its action and speech take (\cref{sec:draft}).
\label{sec:scale}Over a film the draft request grows from \R{run.ctx.first} to \R{run.ctx.last} tokens (\texttimes\R{run.ctx.ratio}; \cref{fig:plots}b) and the plan to \R{run.plan.max} tokens.

We compare with four recent planners with released code, StoryEngine~\cite{storyengine}, ViMax~\cite{vimax}, VideoMemory~\cite{videomemory} and CineCrew~\cite{cinecrew}, each run from that code through our adapter on the whole screenplay with one neutral brief (\cref{sec:supp_baselines}).
Every planner uses Gemini~3.8~Flash (\method's drafter at low thinking, baselines at medium), one attempt per screenplay each; missing output counts as wrong; no number is compared with their papers. Model API ids and access dates are in the LLM Usage Statement.
All run plan-only; adaptations (\cref{tab:supp_adapt}) follow from that, released-code limits, parity or our choice: StoryEngine's windows share only our outline (carrying state between them did not improve its scores on two screenplays, \cref{sec:supp_secarried}); VideoMemory runs only its storyboard stage, without appearance text or people and prop stills; we turned off CineCrew's four LLM text stages after staging, so our prompt assembly writes its requests (turning them on raised its scores on two screenplays, still below \method's, \cref{sec:supp_ccfull}).
Every baseline number comes from a rerun, decided after first-run results, that re-sent only calls lost to our adapters' and machines' faults (\cref{sec:supp_baselines}).

\subsection{\speall results}
\label{sec:spe_results}
\label{sec:speobj}

\Cref{tab:main} gives the scores on all \R{spx.n.all} \speall questions, by part and kind. Over all of them, \method scores \R{spx.all.cont.jury} / \R{spx.all.cont.argon}, against \R{spx.all.best.jury} / \R{spx.all.best.argon} for the best baseline: a difference of \R{spx.lead.all.jury} [\R{spx.lead.all.jury.cl.lo}, \R{spx.lead.all.jury.cl.hi}] points under the jury and \R{spx.lead.all.argon} [\R{spx.lead.all.argon.cl.lo}, \R{spx.lead.all.argon.cl.hi}] under Argon.
Brackets give 95\% intervals, bootstrap over the nine films (pairs resampled within each drawn film); a difference whose interval spans zero is not distinguishable at this $n$.
Differences are computed from unrounded rates.
On \speobj, \method scores \R{spo.all.cont.jury} / \R{spo.all.cont.argon}, against \R{spo.all.best.jury} / \R{spo.all.best.argon} for the best baseline: a difference of \R{spo.lead.all.jury} [\R{spo.lead.all.jury.cl.lo}, \R{spo.lead.all.jury.cl.hi}] points under the jury and \R{spo.lead.all.argon} [\R{spo.lead.all.argon.cl.lo}, \R{spo.lead.all.argon.cl.hi}] under Argon.
It is first on \R{spo.first.jury} of the nine screenplays under the jury and \R{spo.first.argon} under Argon (\cref{tab:supp_perfilm_obj}).
On same-object items the difference is \R{spo.lead.same.jury} [\R{spo.lead.same.jury.cl.lo}, \R{spo.lead.same.jury.cl.hi}] / \R{spo.lead.same.argon} [\R{spo.lead.same.argon.cl.lo}, \R{spo.lead.same.argon.cl.hi}] points, and on the different-object controls \R{spo.lead.diff.jury} [\R{spo.lead.diff.jury.cl.lo}, \R{spo.lead.diff.jury.cl.hi}] / \R{spo.lead.diff.argon} [\R{spo.lead.diff.argon.cl.lo}, \R{spo.lead.diff.argon.cl.hi}].
Balanced accuracy (mean of the two) is \R{spo.bal.cont.jury} / \R{spo.bal.cont.argon} against \R{spo.bal.best.jury} / \R{spo.bal.best.argon} (jury difference \R{spo.lead.bal.jury} [\R{spo.lead.bal.jury.cl.lo}, \R{spo.lead.bal.jury.cl.hi}]); both gates pass on \R{spo.gate.ident.cont.jury} of \method's identity items against at most \R{spo.gate.ident.cc.jury} for a baseline, and on the \R{spo.cov.se} questions StoryEngine covers it scores \R{spo.covscore.all.se.jury} against \method's \R{spo.covscore.all.cont.vsse.jury}: much of the lead is presence and coverage.
On the \R{spo.n.cause} cause-effect items it is \R{spo.lead.cause.jury} [\R{spo.lead.cause.jury.cl.lo}, \R{spo.lead.cause.jury.cl.hi}] / \R{spo.lead.cause.argon} [\R{spo.lead.cause.argon.cl.lo}, \R{spo.lead.cause.argon.cl.hi}]; one answer moves a rate on \R{spo.n.cause} items by about six points.

On \spe, \method scores \R{spe.all.cont.jury} / \R{spe.all.cont.argon} on the \R{spe.n.all} scored questions, against \R{spe.all.best.jury} / \R{spe.all.best.argon} for the best baseline.
The difference is \R{spe.lead.all.jury} [\R{spe.lead.all.jury.cl.lo}, \R{spe.lead.all.jury.cl.hi}] points under the jury and \R{spe.lead.all.argon} [\R{spe.lead.all.argon.cl.lo}, \R{spe.lead.all.argon.cl.hi}] under Argon, and \method is first on \R{spe.first.jury} / \R{spe.first.argon} of the nine screenplays (\cref{tab:supp_perscript}).
By dynamic, the jury difference from that dynamic's best baseline is \R{spe.lead.state.jury} [\R{spe.lead.state.jury.cl.lo}, \R{spe.lead.state.jury.cl.hi}] on story state, \R{spe.lead.emo.jury} [\R{spe.lead.emo.jury.cl.lo}, \R{spe.lead.emo.jury.cl.hi}] on emotion, \R{spe.lead.char.jury} [\R{spe.lead.char.jury.cl.lo}, \R{spe.lead.char.jury.cl.hi}] on character and \R{spe.lead.col.jury} [\R{spe.lead.col.jury.cl.lo}, \R{spe.lead.col.jury.cl.hi}] on colour (Argon in \cref{tab:main}).
On colour \method is behind CineCrew (\R{spe.col.cont.jury} against \R{spe.col.cc.jury}, jury); its requests carry no light or colour line, most of CineCrew's a light line.
On the \R{spe.cov.se} questions whose scenes StoryEngine's shots cover, it scores \R{spe.covscore.se.jury} / \R{spe.covscore.se.argon} and \method{} \R{spe.covscore.cont.vsse.jury} / \R{spe.covscore.cont.vsse.argon}, so most of the \spe difference from StoryEngine is coverage; the other baselines' covered-question scores are in \cref{sec:supp_why}.

\begin{table*}[t]
  \centering
  \caption{\speall scores: share of scored questions answered correctly from the shot requests and stills of the two scenes of a pair. Cell: jury / Argon. \spe: \R{spe.pairs} scene pairs at least 20 scenes apart. \speobj: \R{spo.pairs} object pairs \R{spo.gap.min}--\R{spo.gap.max} scenes apart; an identity answer counts only if the judge finds the object in both scenes. No shots or no answer counts as wrong. Baselines run plan-only through our adapters (VideoMemory: storyboard stage only; \cref{sec:setup}). \Ceiling: the plan that code writes from the answer key.}
  \label{tab:main}\label{tab:speobj}\label{tab:spe}
  \footnotesize
  \setlength{\tabcolsep}{2.6pt}
  \newcommand{\jc}[1]{\R{#1.jury}\,/\,\R{#1.argon}}
  \begin{tabular}{@{}lccccc|cccc|c@{}}
    \toprule
    & \multicolumn{5}{c|}{\spe: scene pairs} & \multicolumn{4}{c|}{\speobj: object pairs} & \speall \\
    System & Character & Emotion & Colour & Story state & All & Same & Different & Cause & All & All \\
           & (\R{spe.n.char}) & (\R{spe.n.emo}) & (\R{spe.n.col}) & (\R{spe.n.state}) & (\R{spe.n.all}) & (\R{spo.n.same}) & (\R{spo.n.diff}) & (\R{spo.n.cause}) & (\R{spo.n.all}) & (\R{spx.n.all}) \\
    \midrule
    VideoMemory          & \jc{spe.char.vmem} & \jc{spe.emo.vmem} & \jc{spe.col.vmem} & \jc{spe.state.vmem} & \jc{spe.all.vmem} & \jc{spo.same.vmem} & \jc{spo.diff.vmem} & \jc{spo.cause.vmem} & \jc{spo.all.vmem} & \jc{spx.all.vmem} \\
    ViMax                & \jc{spe.char.vimax} & \jc{spe.emo.vimax} & \jc{spe.col.vimax} & \jc{spe.state.vimax} & \jc{spe.all.vimax} & \jc{spo.same.vimax} & \jc{spo.diff.vimax} & \jc{spo.cause.vimax} & \jc{spo.all.vimax} & \jc{spx.all.vimax} \\
    CineCrew             & \jc{spe.char.cc} & \jc{spe.emo.cc} & \jc{spe.col.cc} & \jc{spe.state.cc} & \jc{spe.all.cc} & \jc{spo.same.cc} & \jc{spo.diff.cc} & \jc{spo.cause.cc} & \jc{spo.all.cc} & \jc{spx.all.cc} \\
    StoryEngine          & \jc{spe.char.se} & \jc{spe.emo.se} & \jc{spe.col.se} & \jc{spe.state.se} & \jc{spe.all.se} & \jc{spo.same.se} & \jc{spo.diff.se} & \jc{spo.cause.se} & \jc{spo.all.se} & \jc{spx.all.se} \\
    \midrule
    \method              & \jc{spe.char.cont} & \jc{spe.emo.cont} & \jc{spe.col.cont} & \jc{spe.state.cont} & \jc{spe.all.cont} & \jc{spo.same.cont} & \jc{spo.diff.cont} & \jc{spo.cause.cont} & \jc{spo.all.cont} & \jc{spx.all.cont} \\
    \midrule
    \textit{\Ceiling}    & \jc{spe.char.ceil} & \jc{spe.emo.ceil} & \jc{spe.col.ceil} & \jc{spe.state.ceil} & \jc{spe.all.ceil} & \jc{spo.same.ceil} & \jc{spo.diff.ceil} & \jc{spo.cause.ceil} & \jc{spo.all.ceil} & \jc{spx.all.ceil} \\
    \bottomrule
  \end{tabular}
\end{table*}

\subsection{Differences from the baselines as run}
\label{sec:carry}

\Cref{tab:carry} breaks the same-object items down by scene gap and by how an item is lost; \cref{fig:qual} (\cref{sec:supp_qual}) shows two examples.
Against the best baseline of each gap bin, \method's same-object rate differs by \R{spo.gap.2to5.lead.same.jury} [\R{spo.gap.2to5.lead.same.jury.cl.lo}, \R{spo.gap.2to5.lead.same.jury.cl.hi}] / \R{spo.gap.2to5.lead.same.argon} points at 2--5 scenes, \R{spo.gap.6to15.lead.same.jury} [\R{spo.gap.6to15.lead.same.jury.cl.lo}, \R{spo.gap.6to15.lead.same.jury.cl.hi}] / \R{spo.gap.6to15.lead.same.argon} at 6--15 and \R{spo.gap.16to39.lead.same.jury} [\R{spo.gap.16to39.lead.same.jury.cl.lo}, \R{spo.gap.16to39.lead.same.jury.cl.hi}] / \R{spo.gap.16to39.lead.same.argon} at 16--39, where the best baseline falls to \R{spo.gap.16to39.same.best.jury} / \R{spo.gap.16to39.same.best.argon}.
Brackets are jury intervals; with \R{spo.gap.16to39.n.same} items in the far bin they are wide, and none of the three excludes zero.

\emph{Shown, then the same.}
With the gates, a rate is the share shown in both scenes times the accuracy among those.
\method shows the object in both scenes on \R{spo.gate.same.cont.jury} / \R{spo.gate.same.cont.argon} of the \R{spo.n.same} items and the baselines on \R{spo.gate.same.vmem.jury} to \R{spo.gate.same.vimax.jury} (jury); among items shown in both, it is right on \R{spo.gateacc.same.cont.jury} / \R{spo.gateacc.same.cont.argon}, CineCrew on \R{spo.gateacc.same.cc.jury} / \R{spo.gateacc.same.cc.argon} and StoryEngine on \R{spo.gateacc.same.se.jury} / \R{spo.gateacc.same.se.argon}.
Split into presence and identity parts (each difference times the other's mean), the lead over the best baseline at 2--5 scenes is mostly presence, \R{spo.gap.2to5.dec.cov.same.jury} [\R{spo.gap.2to5.dec.cov.same.jury.cl.lo}, \R{spo.gap.2to5.dec.cov.same.jury.cl.hi}] / \R{spo.gap.2to5.dec.cov.same.argon} points against \R{spo.gap.2to5.dec.id.same.jury} / \R{spo.gap.2to5.dec.id.same.argon} for identity, and at 16--39 scenes mostly identity, \R{spo.gap.16to39.dec.id.same.jury} [\R{spo.gap.16to39.dec.id.same.jury.cl.lo}, \R{spo.gap.16to39.dec.id.same.jury.cl.hi}] / \R{spo.gap.16to39.dec.id.same.argon} against \R{spo.gap.16to39.dec.cov.same.jury} / \R{spo.gap.16to39.dec.cov.same.argon} for presence.
In these point estimates, near the earlier scene the baselines lose the object; far from it, they show it as another object.

\begin{table}[t]
  \centering
  \caption{\speobj same-object items (\R{spo.n.same}), jury / Argon. \emph{Gap}: rate by scene gap ($n$ items). \emph{Shown in both}: both presence gates yes; \emph{right if shown}: rate among those. \emph{One prop id}: items whose two scenes' shots carry one registered prop id, as a count (of them at 16--39 scenes), and the rate with and without one; ViMax registers no props.}
  \label{tab:carry}
  \scriptsize
  \setlength{\tabcolsep}{1.6pt}
  \begin{tabular}{@{}lccccc@{}}
    \toprule
    & VideoMemory & ViMax & CineCrew & StoryEngine & \method \\
    \midrule
    Gap 2--5 (\R{spo.gap.2to5.n.same}) & \R{spo.gap.2to5.same.vmem.jury}\,/\,\R{spo.gap.2to5.same.vmem.argon} & \R{spo.gap.2to5.same.vimax.jury}\,/\,\R{spo.gap.2to5.same.vimax.argon} & \R{spo.gap.2to5.same.cc.jury}\,/\,\R{spo.gap.2to5.same.cc.argon} & \R{spo.gap.2to5.same.se.jury}\,/\,\R{spo.gap.2to5.same.se.argon} & \R{spo.gap.2to5.same.cont.jury}\,/\,\R{spo.gap.2to5.same.cont.argon} \\
    Gap 6--15 (\R{spo.gap.6to15.n.same}) & \R{spo.gap.6to15.same.vmem.jury}\,/\,\R{spo.gap.6to15.same.vmem.argon} & \R{spo.gap.6to15.same.vimax.jury}\,/\,\R{spo.gap.6to15.same.vimax.argon} & \R{spo.gap.6to15.same.cc.jury}\,/\,\R{spo.gap.6to15.same.cc.argon} & \R{spo.gap.6to15.same.se.jury}\,/\,\R{spo.gap.6to15.same.se.argon} & \R{spo.gap.6to15.same.cont.jury}\,/\,\R{spo.gap.6to15.same.cont.argon} \\
    Gap 16--39 (\R{spo.gap.16to39.n.same}) & \R{spo.gap.16to39.same.vmem.jury}\,/\,\R{spo.gap.16to39.same.vmem.argon} & \R{spo.gap.16to39.same.vimax.jury}\,/\,\R{spo.gap.16to39.same.vimax.argon} & \R{spo.gap.16to39.same.cc.jury}\,/\,\R{spo.gap.16to39.same.cc.argon} & \R{spo.gap.16to39.same.se.jury}\,/\,\R{spo.gap.16to39.same.se.argon} & \R{spo.gap.16to39.same.cont.jury}\,/\,\R{spo.gap.16to39.same.cont.argon} \\
    \midrule
    Shown in both & \R{spo.gate.same.vmem.jury}\,/\,\R{spo.gate.same.vmem.argon} & \R{spo.gate.same.vimax.jury}\,/\,\R{spo.gate.same.vimax.argon} & \R{spo.gate.same.cc.jury}\,/\,\R{spo.gate.same.cc.argon} & \R{spo.gate.same.se.jury}\,/\,\R{spo.gate.same.se.argon} & \R{spo.gate.same.cont.jury}\,/\,\R{spo.gate.same.cont.argon} \\
    Right if shown & \R{spo.gateacc.same.vmem.jury}\,/\,\R{spo.gateacc.same.vmem.argon} & \R{spo.gateacc.same.vimax.jury}\,/\,\R{spo.gateacc.same.vimax.argon} & \R{spo.gateacc.same.cc.jury}\,/\,\R{spo.gateacc.same.cc.argon} & \R{spo.gateacc.same.se.jury}\,/\,\R{spo.gateacc.same.se.argon} & \R{spo.gateacc.same.cont.jury}\,/\,\R{spo.gateacc.same.cont.argon} \\
    \midrule
    One prop id & \R{spo.carry.vmem.n} (\R{spo.carry.vmem.16to39.n}) & \R{spo.carry.vimax.n} & \R{spo.carry.cc.n} (\R{spo.carry.cc.16to39.n}) & \R{spo.carry.se.n} (\R{spo.carry.se.16to39.n}) & \R{spo.carry.cont.n} (\R{spo.carry.cont.16to39.n}) \\
    \quad right with it & \R{spo.carry.vmem.right.jury}\,/\,\R{spo.carry.vmem.right.argon} & -- & \R{spo.carry.cc.right.jury}\,/\,\R{spo.carry.cc.right.argon} & \R{spo.carry.se.right.jury}\,/\,\R{spo.carry.se.right.argon} & \R{spo.carry.cont.right.jury}\,/\,\R{spo.carry.cont.right.argon} \\
    \quad right without & \R{spo.carry.vmem.noid.right.jury}\,/\,\R{spo.carry.vmem.noid.right.argon} & \R{spo.carry.vimax.noid.right.jury}\,/\,\R{spo.carry.vimax.noid.right.argon} & \R{spo.carry.cc.noid.right.jury}\,/\,\R{spo.carry.cc.noid.right.argon} & \R{spo.carry.se.noid.right.jury}\,/\,\R{spo.carry.se.noid.right.argon} & \R{spo.carry.cont.noid.right.jury}\,/\,\R{spo.carry.cont.noid.right.argon} \\
    \bottomrule
  \end{tabular}
\end{table}

\emph{What the planner does.}
The differences are consistent with three parts of the planner (\cref{sec:draft,sec:verify,sec:export}).
First, every draft request lists the id and name of every person and prop registered earlier, to be reused when one returns: \method's two scenes carry one prop id on \R{spo.carry.cont.n} of the \R{spo.n.same} items (StoryEngine \R{spo.carry.se.n}, CineCrew \R{spo.carry.cc.n}), and at 16--39 scenes on \R{spo.carry.cont.16to39.n} items against \R{spo.carry.se.16to39.n} and \R{spo.carry.cc.16to39.n}.
A shared id brings one description and one still to both scenes, so identity answers largely follow id reuse: with a shared id, StoryEngine and CineCrew are right more often than \method (\R{spo.carry.se.right.jury} and \R{spo.carry.cc.right.jury} against \R{spo.carry.cont.right.jury}, jury), and every planner is seldom right without one.
Second, every window is committed, so none of \method's \R{run.scenes} scenes is left without shots, against \R{why.empty.se} for StoryEngine and \R{why.empty.vmem} for VideoMemory.
Third, each shot names what it shows: \method's requests name the object in \R{why.cont.sidesnamed} of the scenes of an object pair, against \R{why.se.sidesnamed}, \R{why.cc.sidesnamed}, \R{why.vimax.sidesnamed} and \R{why.vmem.sidesnamed} for StoryEngine, CineCrew, ViMax and VideoMemory.

\emph{Where each released planner breaks} (\cref{sec:supp_why}).
Through our adapters, StoryEngine keeps identity only inside one draft window: within one (\R{spo.unit.se.same.n} same-object items) it is right on \R{spo.unit.se.same.se.jury} against \method's \R{spo.unit.se.same.cont.jury}, across windows (\R{spo.unit.se.other.n}) on \R{spo.unit.se.other.se.jury} against \R{spo.unit.se.other.cont.jury} (difference \R{spo.unit.se.other.lead.jury} [\R{spo.unit.se.other.lead.jury.cl.lo}, \R{spo.unit.se.other.lead.jury.cl.hi}]).
CineCrew falls the same way across its prop-library parts (\R{spo.unit.cc.other.cc.jury} against \R{spo.unit.cc.other.cont.jury}); ViMax writes no prop fields and is right on \R{spo.gateacc.same.vimax.jury} of the items it shows in both scenes; VideoMemory's storyboard requests leave scenes empty or the object out.
On the different-object controls shown in both scenes the order differs: StoryEngine is right on \R{spo.gateacc.diff.se.jury} / \R{spo.gateacc.diff.se.argon} and \method on \R{spo.gateacc.diff.cont.jury} / \R{spo.gateacc.diff.cont.argon}.

\subsection{\Planaudit and what the checks caught}
\label{sec:audit}

\speobj and \spe probe chosen scenes; the \planaudit reads every scene of every plan (\cref{sec:audit_proto}).
In the whole-plan reading \method scores \R{aud.read.cont} and the best baseline \R{aud.read.best}, but the reading is asymmetric (three readings for \method, one per baseline), comes from the planners' provider and ranks VideoMemory, last on every \speall total, first among baselines, so we give it little weight.
On the code checks \method is not clearly ahead of the best baseline: \R{aud.pres.cont} against \R{aud.pres.best} on presence, \R{aud.caus.cont} against \R{aud.caus.best} on causal links and \R{aud.rel.cont} against \R{aud.rel.best} on relations, where ViMax is ahead (\cref{tab:supp_checks,tab:supp_reading}).
Before rendering, the keeper raised \R{run.violations} reasons in \R{run.violwindows} windows (per rule in \cref{sec:supp_method}); \R{run.openissues.cont} windows were committed with open issues, which the plan lists.

\subsection{Are the instruments sound?}
\label{sec:valid}

\emph{Judges.}
Per answer over all systems, the jury and Argon agree on \R{agree.spo.pct}\% of \speobj answers ($\kappa=\R{agree.spo.kappa}$) and \R{agree.spe.pct}\% of \spe answers ($\kappa=\R{agree.spe.kappa}$); \R{agree.spo.rank} of the four judges rank \method first among the five compared systems on \speobj and \R{agree.spe.rank} on \spe (\cref{sec:supp_judges}).
Without Opus, which also fixes the truth of \speobj, the majority of Sonnet 5.5, Sol and Argon gives \method{} \R{spo.all.cont.noopus} against \R{spo.all.best.noopus} for the best baseline on \speobj, and \R{spe.all.cont.noopus} against \R{spe.all.best.noopus} on \spe (\cref{tab:supp_noopus}).

\emph{The \ceiling} scores \R{spo.all.ceil.jury} / \R{spo.all.ceil.argon} on \speobj and \R{spe.all.ceil.jury} / \R{spe.all.ceil.argon} on \spe; on its cause-effect items, all stated, the jury scores \R{spo.cause.ceil.jury}, the rest being the jury's miss rate on that kind (\cref{sec:supp_ceiling}).

\emph{Video.} In a small control (\cref{sec:supp_noise}), Argon's answers from video and from requests changed at similar rates (\R{noise.flip.video}, \R{noise.flip.text}); it does not test whether request-space scores predict video-space scores.

\subsection{Limitations}
\label{sec:limits}

Budget limited every planner to one planning LLM, and each screenplay was planned once; results with other or stronger planning models may differ, and a video model can ignore the requests we judge.
The checks cover facts, not taste; scores say what a plan asks for, not what a video model renders, and \speall does not test knowledge state or derived effects.
LLMs judge the plans and write the script sheets; a person checked only the \speobj answers.
The judges' cast key lacked a token in about one judged pair in ten (\cref{sec:supp_tokens}); this likely moves no score by more than about a point; \method stays first.
Baselines ran plan-only on a matched LLM through our adapters, some stages off, perhaps not at their best, in a rerun not fixed in advance.
We do not compare with a single long-context planner given the whole screenplay and the plan so far.
The screenplays were not held out, and we did not test the planning LLM's recall of them: an earlier version was built on \emph{Jockey}, and the final one was tuned on \emph{Smalltime} and run once on \emph{Fast Color} before the full run, only as an error check.
Without \emph{Jockey} and \emph{Smalltime}, \method scores \R{spo.nodev.all.cont.jury} on \speobj and \R{spe.nodev.all.cont.jury} on \spe, against \R{spo.nodev.all.best.jury} and \R{spe.nodev.all.best.jury} for the best baseline (jury; \cref{tab:supp_nodev}).
With nine films, the smallest differences from the best baseline detectable with 80\% power are about \R{spo.lead.same.jury.mde} points on same-object items, \R{spo.lead.cause.jury.mde} on cause-effect items and \R{spe.lead.all.jury.mde} on \spe (jury, film-clustered).

\section{Conclusion}
\label{sec:conclusion}

\method turns a feature screenplay into a complete shot plan with a film-wide registry of people and props and a \ledger that records each fact's writer.
On all \R{spx.n.all} \speall questions it scores \R{spx.all.cont.jury} against \R{spx.all.best.jury} for the best of four publicly released planners (jury), \R{spo.all.cont.jury} against \R{spo.all.best.jury} on the object questions.
The difference is consistent with planning every scene and giving every draft the ids registered before it (\cref{sec:carry}).
If request-space fidelity carries to the screen, much of what is lost between scenes is lost in the plan; we have not tested this.
Next steps are whole-film rendering, other planning LLMs and more screenplays.

\section*{LLM Usage Statement}
\label{sec:llm}

LLMs are part of the method. Gemini 3.8 Flash is the LLM of every planner in the study, ours included, and of every LLM role in \method.

LLMs are part of the evaluation. Claude Opus 5.5, Claude Sonnet 5.5 and GPT-6.1 Sol form the jury of \speobj and \spe. Gemini 4 Argon is the second judge, the whole-plan reader, the judge of one plan check (script events shown) and the judge of the video-judge noise control. GPT-5.4 and Claude Opus 5.5 read the screenplays into the script sheets, and two blind Claude Opus 5.5 readings fix the truth of \speobj; its results without Opus on the jury are in \cref{tab:supp_noopus}.

Generative models made the media. An image model (Nano Banana Pro) drew the reference stills, and a video model (Gemini Omni 1.2 Flash) rendered the clips of the video-judge noise control.

Beyond these roles, LLM coding assistants wrote and reviewed research code, ran experiments and computed statistics, and LLMs drafted and edited text, tables and figures. The authors checked everything and take responsibility for it.

Models and API ids, all called through their providers' APIs between 7 and 16 November 2026: Gemini 3.8 Flash (\texttt{gemini-3.8-flash}), Gemini 4 Argon (\texttt{gemini-4-argon}), Claude Opus 5.5 (\texttt{claude-opus-5-5}), Claude Sonnet 5.5 (\texttt{claude-sonnet-5-5}), GPT-6.1 Sol (\texttt{gpt-6.1-sol}), GPT-5.4 (\texttt{gpt-5.4-2026-03-05}), Nano Banana Pro (\texttt{gemini-3-pro-image}) and Gemini Omni 1.2 Flash (\texttt{gemini-omni-1.2-flash}).

{
    \small
    \bibliographystyle{ieeenat_fullname}
    \bibliography{main}
}

% ---- Supplementary material (numbered S1, S2, ...) ----
\clearpage
\setcounter{section}{0}\setcounter{table}{0}\setcounter{figure}{0}\setcounter{equation}{0}
\renewcommand{\thesection}{S\arabic{section}}\renewcommand{\thetable}{S\arabic{table}}
\renewcommand{\thefigure}{S\arabic{figure}}\renewcommand{\theequation}{S\arabic{equation}}
\renewcommand{\theHsection}{S\arabic{section}}\renewcommand{\theHtable}{S\arabic{table}}
\renewcommand{\theHfigure}{S\arabic{figure}}\renewcommand{\theHequation}{S\arabic{equation}}
\maketitlesupplementary

\section{Related systems in full}
\label{sec:supp_related}

\Cref{tab:related} compares \method with the closest systems, cell by cell, as their papers report them; where the released code we ran differs from the paper, the cell says so. A dash means the paper does not let us establish the cell.

\begin{table*}[h]
\caption{Comparison with the closest systems, as reported in their papers (``code'': as in the released code we ran). \emph{Longest} = largest evaluated output. \emph{Who knows} = tracks which characters know which facts. \emph{Plan check} and \emph{Plan repair} concern the plan before rendering. A dash = could not establish from the paper. $^\dagger$Rendered length; plans rated on the first 100 storyboards per novel. $^\ddagger$Segments, in videos of up to ten minutes. $^\S$A critic reviews rendered clips and proposes plan edits. $^a$After a failed render check the verifier rewrites the shot's motion text, up to three rounds; the scene and shot plan are not edited. $^b$Its repair loop re-renders a shot with the plan fixed; the released code also resends a story draft that fails its schema or compile check, with the error, up to three attempts. $^c$The released code has a validator of entity ids and a reviewer of render prompts, both on by default; both were off in our plan-only run (\cref{sec:supp_baselines}). $^d$The storyboard prompt printed in its supplement excludes camera terms; the released prompt asks for camera movement per shot. $^e$Implemented and recorded in the plan; no \speobj or \spe question tests it. $^\P$Over the nine plans \method{} sent \R{run.notices.cont} notices to earlier windows, and \R{run.noticescleared.cont} cleared their fault (\cref{sec:supp_method}).}
\label{tab:related}
\centering
\footnotesize
\setlength{\tabcolsep}{2pt}
\begin{tabular}{@{}l>{\raggedright\arraybackslash}p{1.5cm}l>{\raggedright\arraybackslash}p{1.9cm}ll>{\raggedright\arraybackslash}p{2.45cm}>{\raggedright\arraybackslash}p{2.1cm}>{\raggedright\arraybackslash}p{1.5cm}@{}}
\toprule
System & Plans story & Longest & Typed state & Derived & Who & Plan check & Plan repair & Camera \\
 & & (shots) & & effects & knows & & & plan \\
\midrule
FilmWorld~\cite{filmworld} & yes, in levels & 244 & hashed states & no & no & no (renders only) & no$^a$ & per shot \\
MovieAgent~\cite{movieagent} & yes, in levels & $\le$100 & no & no & no & no & no & per shot \\
ViMax~\cite{vimax} & yes, in levels & 16$^\dagger$ & no (untyped graph) & no & no & no (renders only) & no & camera graph \\
StoryEngine~\cite{storyengine} & yes & 10 & placements,\newline attributes & named only & no & schema & no$^b$ & per shot \\
VideoMemory~\cite{videomemory} & yes & 12 & entity memory bank & -- & -- & -- & -- & per shot (code)$^d$ \\
CineCrew~\cite{cinecrew} & yes & -- & assets,\newline locked props & -- & -- & no in paper; id check in code$^c$ & local edits$^\S$ & per clip \\
A$^2$RD~\cite{a2rd} & given or planned & 80$^\ddagger$ & multimodal\newline memory & no & no & no (checks keyframes) & no (segment re-render, prompt rewrite) & tracked, not planned \\
EntityMem~\cite{entitybench} & given & 50 & entity memory & no & no & -- & no & per shot \\
\midrule
\method & yes, top-down & \R{run.shots.max} & \ledger,\newline with reasons & yes$^e$ & yes$^e$ &outline, operations,\newline quotes, speech, camera & redraft, notice to an earlier window$^\P$, brief rewrite & yes \\
\bottomrule
\end{tabular}
\end{table*}

\section{Method details}
\label{sec:supp_method}

\begin{figure}[t]
  \centering
  \input{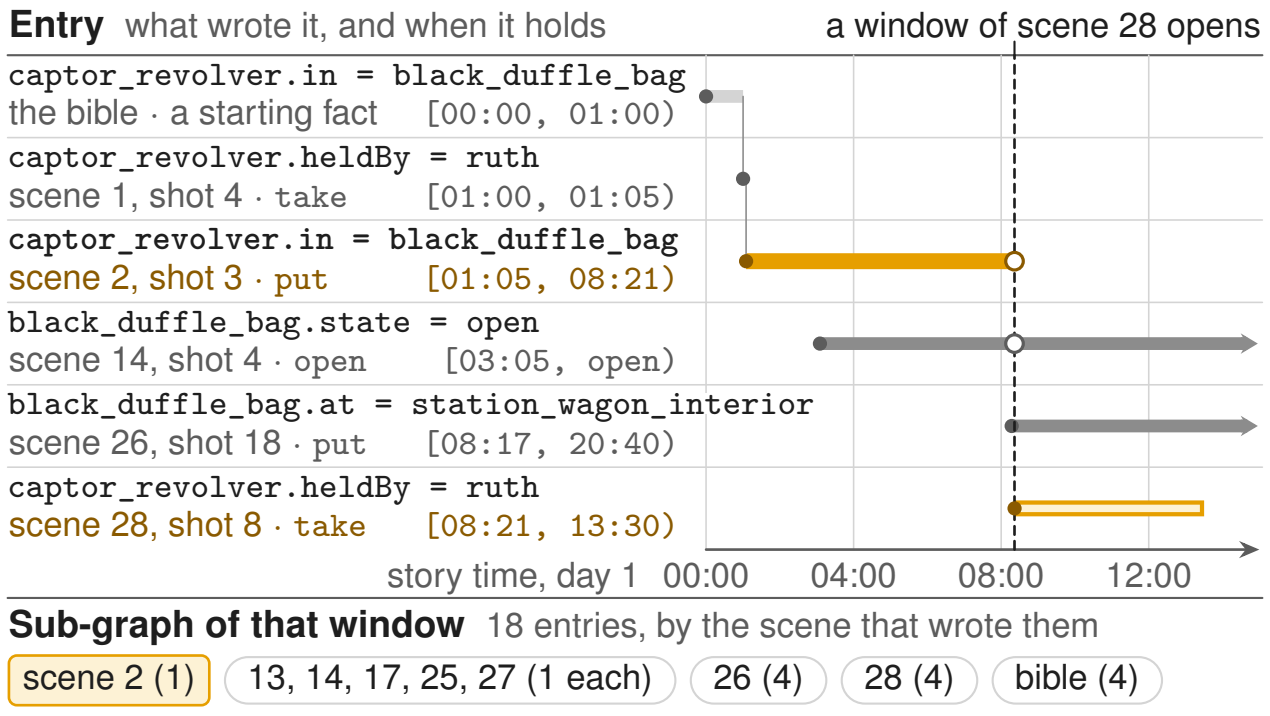}
\begin{tikzpicture}[fig, x=1cm, y=1cm,
  interval new/.style={draw=cLedger, line width=3pt, line cap=butt},
  interval new core/.style={draw=cLedgerFill, line width=1.4pt, line cap=butt}]
  \path[use as bounding box] (0,0) rectangle (8.3,4.62);
  \def\xo{4.62}\def\hw{0.244}            %
  \def\lx#1{\xo+(#1)*\hw}                 %
  \def\rp{0.545}                           %
  \def\ytop{4.29}\def\yax{1.02}\def\tnow{8.3614}
  \def\ry#1{\ytop-.5*\rp-(#1)*\rp}        %
  \node[anchor=base west, inner sep=0pt] at (0,4.41) {\figlabelb Entry\ \ \fignote\color{cInk2}what wrote it, and when it holds};
  \node[anchor=base east, inner sep=0pt] at (8.3,4.41) {\fignote a window of scene 28 opens};
  \draw[cInk2, line width=.4pt] (0,\ytop) -- (8.3,\ytop);
  \foreach \h/\t in {0/{00:00}, 4/{04:00}, 8/{08:00}, 12/{12:00}} {
    \draw[cRule, line width=.3pt] ({\lx{\h}},\yax) -- ({\lx{\h}},\ytop);
    \node[anchor=north, inner sep=2.2pt] at ({\lx{\h}},\yax) {\fignote\color{cInk2}\t};}
  \node[anchor=north east, inner sep=2.2pt] at (4.27,\yax) {\fignote\color{cInk2}story time, day 1};
  \foreach \k in {1,...,5} {\draw[cRule, line width=.3pt] (0,{\ytop-\k*\rp}) -- (8.3,{\ytop-\k*\rp});}
  \draw[time axis] (\xo,\yax) -- (8.28,\yax);
  \foreach \k/\fact/\wr/\iv/\ink in {
     0/{captor\_revolver.in = black\_duffle\_bag}/{the bible $\cdot$ a starting fact}/{[00:00, 01:00)}/cInk2,
     1/{captor\_revolver.heldBy = ruth}/{scene 1, shot 4 $\cdot$ {\figmono take}}/{[01:00, 01:05)}/cInk2,
     2/{\fontseries{b}\selectfont captor\_revolver.in = black\_duffle\_bag}/{scene 2, shot 3 $\cdot$ {\figmono put}}/{[01:05, 08:21)}/cLedgerInk,
     3/{black\_duffle\_bag.state = open}/{scene 14, shot 4 $\cdot$ {\figmono open}}/{[03:05, open)}/cInk2,
     4/{black\_duffle\_bag.at = station\_wagon\_interior}/{scene 26, shot 18 $\cdot$ {\figmono put}}/{[08:17, 20:40)}/cInk2,
     5/{captor\_revolver.heldBy = ruth}/{scene 28, shot 8 $\cdot$ {\figmono take}}/{[08:21, 13:30)}/cLedgerInk} {
    \pgfmathsetmacro{\y}{\ry{\k}}
    \node[anchor=base west, inner sep=0pt, outer sep=0pt, fill=white, text depth=0pt] at (0,\y+.078) {\figmono \fact\,};
    \node[anchor=base west, inner sep=0pt] at (0,\y-.184) {\fignote\color{\ink}\wr};
    \node[anchor=base east, inner sep=0pt] at (4.45,\y-.184) {\figmono\color{\ink}\iv};
  }
  \draw[interval] ({\lx{0}},{\ry{0}}) -- ({\lx{1.0039}},{\ry{0}});
  \draw[interval] ({\lx{1.0039}},{\ry{1}}) -- ({\lx{1.0853}},{\ry{1}});
  \draw[interval carry] ({\lx{1.0853}},{\ry{2}}) -- ({\lx{8.3628}},{\ry{2}});
  \draw[interval on] ({\lx{3.0872}},{\ry{3}}) -- ({\lx{14.55}},{\ry{3}});
  \draw[interval on, -{Stealth[length=3.2pt,width=3.8pt]}] ({\lx{14.5}},{\ry{3}}) -- (8.27,{\ry{3}});
  \draw[interval on] ({\lx{8.2844}},{\ry{4}}) -- ({\lx{14.55}},{\ry{4}});
  \draw[interval on, -{Stealth[length=3.2pt,width=3.8pt]}] ({\lx{14.5}},{\ry{4}}) -- (8.27,{\ry{4}});
  \draw[interval new] ({\lx{8.3628}},{\ry{5}}) -- ({\lx{13.5017}},{\ry{5}});
  \draw[interval new core] ({\lx{8.3628}+.028},{\ry{5}}) -- ({\lx{13.5017}-.028},{\ry{5}});
  \foreach \t/\ka/\kb in {1.0039/0/1, 1.0853/1/2} {
    \draw[cInk3, line width=.3pt] ({\lx{\t}},{\ry{\ka}}) -- ({\lx{\t}},{\ry{\kb}});}
  \foreach \t/\k/\c in {0/0/cInk2, 1.0039/1/cInk2, 1.0853/2/cLedgerInk, 3.0872/3/cInk2, 8.2844/4/cInk2} {
    \fill[\c] ({\lx{\t}},{\ry{\k}}) circle[radius=.047];}
  \draw[now] ({\lx{\tnow}},\yax) -- ({\lx{\tnow}},\ytop);
  \draw[cInk, line width=.4pt] ({\lx{\tnow}},\ytop) -- ({\lx{\tnow}},\ytop+.09);
  \foreach \k/\c in {2/cLedgerInk, 3/cInk2} {\draw[\c, fill=white, line width=.6pt] ({\lx{\tnow}},{\ry{\k}}) circle[radius=.058];}
  \fill[cLedgerInk] ({\lx{8.3628}},{\ry{5}}) circle[radius=.047];
  \draw[cInk2, line width=.4pt] (0,.70) -- (8.3,.70);
  \node[anchor=base west, inner sep=0pt] at (0,.44) {\figlabelb Sub-graph of that window\ \ \fignote\color{cInk2}18 entries, by the scene that wrote them};
  \node[fact, anchor=west, font=\fignote] (g0) at (0,.155) {scene 2 (1)};
  \node[chip, anchor=west] (g1) at ([xshift=2.5pt]g0.east) {13, 14, 17, 25, 27 (1 each)};
  \node[chip, anchor=west] (g2) at ([xshift=2.5pt]g1.east) {26 (4)};
  \node[chip, anchor=west] (g3) at ([xshift=2.5pt]g2.east) {28 (4)};
  \node[chip, anchor=west] (g4) at ([xshift=2.5pt]g3.east) {bible (4)};
\end{tikzpicture}%
  \caption{\textbf{What the \ledger records, and what one window is sent} (\emph{Fast Color}, the example of \cref{fig:overview}). Each row is one entry: the fact, what wrote it and the story time over which it holds. A thin line joins an entry to the earlier entry in its slot, which it ends and rests on. Dashed line: the second window of scene~28 opens and is sent the entries in force: the amber one, written 26 scenes earlier (the revolver is still in the bag), and the dark ones (the bag is open, in the station wagon). Hollow bar: the entry this window's own shot then writes. Below: the window's \subgraph, by the scene that wrote each entry.}
  \label{fig:ledger}
\end{figure}

This section gives the final engine's details that \cref{sec:method} leaves out. Every role is one LLM, Gemini 3.8 Flash; draft calls use its low thinking level, and the other calls set no thinking level, so the model's default applies.

\paragraph{Schemas.} \emph{Bible} (one film call): logline; colour script (palettes; turns in story order with palette, hope 1--10, warmth $-5$ to 5, intensity 1--10 and the causing event); characters (aliases, age, face, first wardrobe); places as a tree, with interior and shelter flags; props with container and vehicle flags and the information they carry; information items with their first knowers; looks (a temporary one names the event that ends it); starting facts; three acts. Code checks that the acts cover every scene once and that every information item has a first knower or a prop that carries it, and sends the reasons back once. \emph{Brief} (act calls of twelve scenes): place, time of day, story time, declared time jump, weather, cast, props; the facts true at opening and at close; what the scene needs from earlier scenes and delivers to later ones; look changes; its point on the colour script; scene parts; new minor figures with backstory. \emph{Draft} (one window): source events (exact quote, certainty, assigned shot); coverage (axis, setups, establishing setup, key, fill and rim light); the window's shots, each with setup, scale, angle and movement, action, blocking, on-screen ids, dialogue, operations, intensity, warmth, colour, its own light, its length and its source events; closing state; named exceptions.

\paragraph{Operations.} \Cref{tab:ops} lists the 16 operations in five kinds, plus \texttt{nochange} for a shot that changes nothing. No operation, no change: code replays each operation, checks its preconditions against the \ledger and writes its effects with their reasons. The extractor may add 13 of them (all but \texttt{see}, \texttt{feel} and \texttt{say}).

\begin{table}[h]
  \centering
  \caption{Typed operations a shot may carry. Code checks and records them; a state change without an operation is a violation.}
  \label{tab:ops}
  \scriptsize
  \setlength{\tabcolsep}{2pt}
  \begin{tabular}{@{}l>{\raggedright\arraybackslash}p{6.4cm}@{}}
    \toprule
    Operation & Precondition $\to$ effect \\
    \midrule
    \multicolumn{2}{@{}l}{\emph{Movement}} \\
    \texttt{move} & $\to$ actor at place; a vehicle as actor carries its riders and what they hold \\
    \texttt{board} & vehicle within reach, not destroyed $\to$ actor aboard \\
    \texttt{alight} & actor aboard something $\to$ no longer aboard \\
    \multicolumn{2}{@{}l}{\emph{Objects}} \\
    \texttt{take} & prop within reach, held by no one, not destroyed $\to$ held by actor \\
    \texttt{put} & actor holds it (or it lies unheld within reach); target a container within reach $\to$ in container or at place \\
    \texttt{give} & as \texttt{put}; receiver present $\to$ held by receiver \\
    \texttt{open}, \texttt{close} & prop within reach, not destroyed $\to$ new state \\
    \texttt{damage} & prop: damaged, destroyed or spent; person: wounded, dead, healed (only a condition they have) or none \\
    \multicolumn{2}{@{}l}{\emph{Appearance}} \\
    \texttt{wear} & a bible look of the actor, not already worn $\to$ new look \\
    \multicolumn{2}{@{}l}{\emph{Perception, feeling}} \\
    \texttt{see} & thing present $\to$ noticed \\
    \texttt{feel} & $\to$ emotion word and tension 0--10; the only way emotion changes \\
    \multicolumn{2}{@{}l}{\emph{Speech, knowledge}} \\
    \texttt{say} & talk that carries no fact \\
    \texttt{tell}, \texttt{show} & actor knows the item; hearers present $\to$ hearers know it (aloud: all present; whisper: named hearers) \\
    \texttt{read} & prop within reach and carries the item $\to$ actor knows it \\
    \bottomrule
  \end{tabular}
\end{table}

\paragraph{The \ledger.} A fact stores entity, attribute, value, the story-time interval $[t_0,t_1)$ in seconds from the first day, its writer (window and shot), its source, the rule that derived it, and the ids of the facts it rests on. Placement (\textsf{at}, \textsf{heldBy}, \textsf{in}) is one shared slot, so a thing is in one place at a time; \textsf{aboard}, \textsf{look}, \textsf{state}, \textsf{emotion} and \textsf{tension} hold one value each; \textsf{knows} and \textsf{condition} are sets. As a scene opens, a prop a cast member holds that the scene's brief does not use is recorded as still with them; it returns to hand in the next scene whose brief names it. The nine plans hold \R{run.facts} facts, \R{run.factswithreason.pct}\% of them with a recorded reason. Every fact names its writer: a window (and shot) for all but \R{run.facts.bible}, which are the bible's starting facts, with the bible as writer.

\paragraph{Derived effects.} Rules add what an operation implies but does not state, and are recorded as the writers. Contents travel with their container or holder, and a vehicle carries its riders and what they hold. A line said aloud reaches everyone present, a whisper only its named hearers; otherwise knowledge moves only by \texttt{tell}, \texttt{show} and \texttt{read}, so a character absent when a secret was told cannot reveal it later. The dead do not act or speak, and a destroyed prop stays destroyed. A condition carries a lifetime (the scene, the day, or until an operation ends it); weather writes none. For example, a character carries a bag holding a letter into another room, and the \ledger records the letter there with the hold chain as its reason.

\paragraph{Outline check, outline graph and time jumps.} Before any shot is drafted, code checks the briefs in story order: every opening fact must follow from the bible, an earlier brief or a declared time jump; every need has a producer; delivered knowledge has a source in the scene; a prop can reach the scene that uses it; a temporary look ends; nobody is in two places at once; story time runs forward unless a jump is declared. A fault goes back to the brief that caused it, which the act planner rewrites, up to three rounds; drafting starts from the round with the fewest faults. Rewrites are corrective: the screenplay wins over the brief, and code drops any added person, prop or off-screen change whose quote is not in the scene text. A scene with no declared jump opens when the scene before it lets its cast go. Each checked brief is written into the \ledger at its story time, part by part, with the brief as writer: as the scene opens, its cast at its place, its look changes and opening facts; each later part's arrivals; at its planned end, what it closes with and delivers. Act calls and rewrites see these rows (entity, attribute, value, since when, which brief set it), and the check reads each brief's opening from them at its story time, so a flashback written later does not overwrite the present. While drafting, each scene opening is checked against the committed \ledger before its first draft. A declared time jump may reset placement, look, state, condition and feelings as the scene opens; knowledge is never reset. Faults found in the first outline round: \R{run.outlinefaults}; left after the rewrites: \R{run.outlineleft}.

\paragraph{Scene parts and windows.} When a scene changes place or time under one heading, its brief declares parts; each part quotes the paragraph it begins with and names its place, time and people. Code checks only structure (quotes found, in order, places registered); otherwise the scene keeps one part. A part longer than 1{,}746 characters is cut at paragraph starts, never after a speaker cue. The limit comes from the draft output cap: 32{,}768 tokens, less a fixed base of 8{,}657 and a margin of 18{,}014 for thinking, at 3.49 output tokens per screenplay character (a least-squares fit on earlier drafts).

\paragraph{The \subgraph.} The seeds are the people and place of the window's scene part, the props of its brief, anyone the scene's earlier windows introduced, and any changed prop lying at the place. Code takes the facts in force about the seeds as the window opens, plus every \textsf{heldBy}, \textsf{in} or \textsf{aboard} fact whose value is a seed: one link, with no length cap. Rows are ordered from placement to knowledge, each with the window that wrote it. Continuity rows go to the established details instead.

\paragraph{The draft request.} In order: the colour script's reading and each palette's name and reason; the turns in force; the brief; the facts the scene must open and close with; the bible rows of every entity named, and the id and name of every other person and prop registered so far, to be reused when one returns (in both arms); then five \ledger sections, each only when it has rows:
(a)~\textsc{World as this window opens}, the \subgraph;
(b)~\textsc{Earlier windows these facts come from};
(c)~\textsc{Where earlier shots left these}: for a prop an earlier shot put down or put away, that shot's action, verbatim, so a returning object's spot is described in the same words;
(d)~\textsc{Later scenes rely on}: facts true now that a later brief requires and no scene in between restores, each with its nearest consumer;
(e)~\textsc{Established details}: the continuity rows in force for the window's people, props and places.
Then come the scene so far (or the last two shots drafted, and the last shots of a scene this one picks up), the lights last seen at this place, and the scene's text before, in and after the window. The colour words of the palettes and of the brief's arc are not sent: they are written for the whole scene before any shot has a place or a light, so the drafter writes each shot's colour and light line for its own frame; the plan keeps those lines, but the judged request omits them.

\paragraph{Continuity rows.} The extractor lists every lasting physical state the accepted shots show of a registered person, prop or place (an injury and its side, a dressing, clothes, eyewear, hair, how a prop is wrapped or broken, a vehicle's identity, a place's damage), and marks each row new, same, changed (with the words that change it) or differing from what is established. Code checks structure only (a registered entity, a key, a shot in range) and writes the rows at their shots' story times. A row ends with its lifetime or at a later change; one set in a flashback ends where the film has already shown that time. Later windows see all rows for people and the twelve newest for each prop or place. A shot that shows an established state otherwise, with nothing in the window to change it, gets one redraft; if that does not settle it, the difference is listed as an open issue. Continuity rows live only in the drafting \ledger, never in the outline rows.

\begin{figure*}[t]
  \centering
  \includegraphics[width=\textwidth]{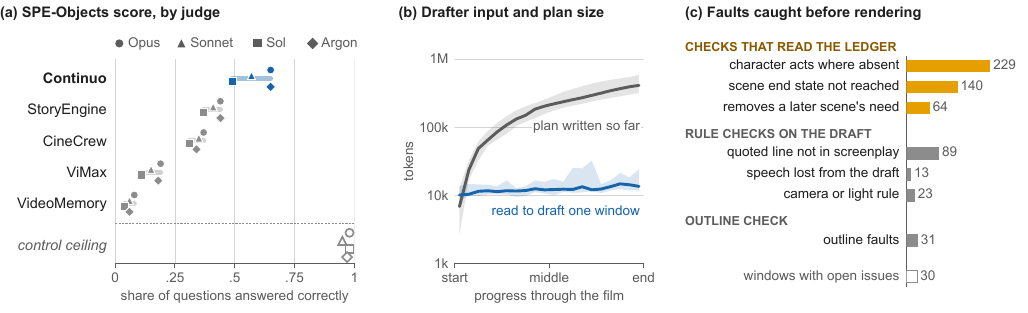}
  \caption{\textbf{Judges, drafter input and checks.} (a)~\speobj score of each system under each of the four judges (marker shape; line: their range); hollow markers: the \ceiling. (b)~Tokens the drafter reads for one window and tokens of the plan written so far, as a film progresses; line: median, band: range over the nine films. (c)~Faults caught before rendering, summed over the nine films: six named groups of keeper reasons, checks that read the \ledger (amber) and other checks (grey), the outline faults, and windows committed with open issues (hollow). The six groups are part of the \R{run.violations} keeper reasons; every rule's count is in \cref{sec:supp_method} (Keeper rules). The bar ``removes a later scene's need'' also counts the brief's closing and delivered facts not reached; the forward check alone is \R{run.chk.forward.cont}.}
  \label{fig:plots}
\end{figure*}

\paragraph{Keeper rules.} The keeper checks eight rules: it replays a draft's operations on a copy of the \ledger and rejects the draft when
(i)~an operation's precondition fails (its actor is not in the scene, its prop is out of reach or held by someone else, or someone reveals what they have not learned);
(ii)~the facts the brief requires as the scene opens or closes do not hold, or the draft's claimed end state does not follow from its operations;
(iii)~the scene destroys or removes something a later brief needs and no scene in between restores (the forward constraint; destroying a radio a later scene needs intact fails);
(iv)~a quoted source event is not exact text of the window, or a shot shows no source event;
(v)~a line is not the window's speech, or a speech loses words;
(vi)~a person on screen or speaking is not in the scene, or an id is not registered or is registered as another kind;
(vii)~a camera setup is unknown, or a key light does not fit the time of day (a night scene keyed by the sun fails); or
(viii)~a feeling the brief requires is not reached by a \texttt{feel} operation.
Feeling faults (viii) become warnings from the second draft; tension is not checked. A fault message states the conflict and where it comes from, never a fix. Checks (i)--(iii) read the \ledger; (iv)--(vii) are rule checks that need none.
Over the nine plans the keeper gave \R{run.keeper.total} reasons in draft rounds: (i)~\R{run.keeper.i}, (ii)~\R{run.keeper.ii}, (iii)~\R{run.keeper.iii}, (iv)~\R{run.keeper.iv}, (v)~\R{run.keeper.v}, (vi)~\R{run.keeper.vi}, (vii)~\R{run.keeper.vii}, (viii)~\R{run.keeper.viii}, and \R{run.keeper.other} for a named exception that cites no earlier cause. Outline faults are counted apart: \R{run.outlinefaults.cont} in the first round, \R{run.outlineleft.cont} left.
\Cref{fig:plots}c shows six groups of these: characters acting where absent (\R{run.chk.absent}: the presence and placement preconditions of (i) and the on-screen presence of (vi)), claimed end states not reached (\R{run.chk.endstate}, in (ii) and (viii)), the brief's closing or delivered facts not reached plus the forward check (\R{run.chk.laterneed}, of which (iii) is \R{run.chk.forward.cont}), quotes not in the screenplay (\R{run.chk.quote}, in (iv)), speech that loses words (\R{run.chk.speech}, in (v)) and camera or light (\R{run.chk.camera}, (vii)). The rest are wrong or unregistered ids, events cited by the wrong shot, a look already worn, lines not under a speaker cue, and the other preconditions. The longest film took \R{run.hours.max}~h to plan.

\paragraph{AI gates.} The extractor reads a passing draft's action and dialogue against the window text, with what each person wears as the window opens and the established details, and lists changes that no operation records; code adds them as operations (and any new look they need) and runs the keeper again. If the added operations break a rule, the drafter gets one redraft with the extractor's reasons; if that fails, the draft that passed is kept with a warning. The proposer answers what rules cannot decide (the cause of a condition no scene shows; whether something inside a destroyed container survived). An answer below confidence 0.3 is not recorded, and the question stays open.

\paragraph{Repair.} A window gets a draft and one redraft with the keeper's reasons. If both fail and the fault rests on facts an earlier window wrote, code follows the stored reasons to the windows that wrote them (roots) and sends a notice to the root whose change touches the fewest shots, counted as its own shots plus the shots elsewhere that read the facts it would change. The root's drafter gets the failing facts, a proposed edit and its committed continuity rows, and approves, adjusts (another edit meeting the same need) or rejects with a prior cause. A change is replayed through the windows after the root; only windows whose reads changed are re-checked, and the change is reverted if one breaks. A failed fix is recorded per root and fault, and two rule a root out. The film has one notice per ten windows (plus one), a window at most two repairs. If the fault remains, the act planner rewrites the scene's brief once, to agree with the scene text, and code checks the rewrite against the committed \ledger before it is drafted. A window that still fails is committed with its open issues listed.

\paragraph{The \ledger in the full plans.}
Over the nine full plans the \ledger holds \R{run.facts} facts, \R{run.factswriter.pct}\% written by a named window (the other \R{run.facts.bible} are the bible's starting facts, with the bible as writer) and \R{run.factswithreason.pct}\% with a recorded reason.
The extractor made \R{run.extract.findings.cont} findings and added \R{run.extract.ops.cont} operations the drafts left out, in \R{run.extract.windows.cont} windows; code stored \R{run.controws.cont} continuity rows and flagged \R{run.contdiffers.cont} differences from what was established.
The \R{run.notices.cont} notices to earlier windows ended as follows: fault cleared, \R{run.noticescleared.cont}; fault left standing, \R{run.notices.persist.cont}; revised window failed the keeper, \R{run.notices.failedkeeper.cont}; change reverted, \R{run.notices.reverted.cont}; notice rejected, \R{run.notices.rejected.cont}.
The forward check over later briefs, rule (iii), raised \R{run.chk.forward.cont} reasons.

\paragraph{Drafting ahead.} For speed, code drafts up to twelve later scenes at once (twelve calls in flight) against the committed \ledger plus what the checked briefs in between promise. Windows still commit in screenplay order: a guess stands only if its brief is unchanged and the keeper, run again on the committed \ledger, passes and reads every earlier fact with the value the guess assumed; otherwise the window is drafted again in order. A scene that picks up the one before (same people, within two minutes of story time) is never drafted ahead. Of the \R{run.windows.cont} windows, \R{run.spec.hit} committed a draft made ahead, \R{run.spec.miss} were drafted again in order after their guess missed, and \R{run.spec.none} had no guess that passed its own keeper ahead.

\paragraph{Constants.} Twelve scenes per act call; at most three outline rounds; one repair call for the film call and for each act call; windows cut above 1{,}746 characters; speech at 2.8 words a second; two keeper rounds (a draft and one redraft), plus one round for an extractor redraft or a continuity difference; one brief rewrite per brief; two repairs per window; two failed fixes rule out a root; one notice per ten windows, plus one; proposer confidence 0.3; twelve continuity rows per prop or place, people uncapped; twelve scenes drafted ahead; input cap 150k tokens per call.

\paragraph{LLM roles.} \Cref{tab:roles} gives each role's inputs and outputs. Every call returns JSON validated against the role's schema; no sampling setting was changed.

\begin{table*}[h]
\centering
\caption{LLM roles in \method. Out cap = output token cap. Calls: every call sent over the nine films, including drafts made ahead and not kept; Request = mean input tokens over those calls. Drafter: \R{run.calls.draft.first} first drafts (mean request \R{run.req.draft.first}) and \R{run.calls.draft.redraft} redrafts, which also carry the previous draft and the keeper's reasons (\R{run.req.draft.redraft}); \R{run.rounds.cont} of the drafts are the committed windows' drafts (\R{run.drafts.cont} per window, \cref{tab:internals}), and the other \R{run.calls.draft.ahead} were drafted ahead and not kept: a guess that failed its own keeper ahead, missed at commit, or was cancelled and asked again when an earlier window changed what it reads. The drafter request's growth over a film, from \R{run.ctx.first} to \R{run.ctx.last} tokens (mean of the first and of the last tenth of a film's windows), is measured on each window's first draft call, so it sits below the mean over all draft calls.}
\label{tab:roles}
\footnotesize
\setlength{\tabcolsep}{3pt}
\begin{tabular}{@{}l>{\raggedright\arraybackslash}p{5.6cm}>{\raggedright\arraybackslash}p{3.8cm}rrr@{}}
\toprule
Role & Inputs & Output & Out cap & Calls & Request \\
\midrule
Film planner & whole screenplay & bible, colour script, acts & 32k & \R{run.calls.film} & \R{run.req.film} \\
Act planner & screenplay, bible, outline rows so far, twelve scenes & one brief per scene & 65k & \R{run.calls.act} & \R{run.req.act} \\
Brief rewrite & brief, faults, what holds as the scene opens, scene text & whole revised brief & 65k & \R{run.calls.rewrite} & \R{run.req.rewrite} \\
Drafter & colour script, brief, bible rows, sections (a)--(e), scene so far, scene text; keeper reasons & coverage, shots, operations, closing state & 32k & \R{run.calls.draft} & \R{run.req.draft} \\
Extractor & registry, what each person wears, established details, drafted shots, window text & added operations, continuity rows & 16k & \R{run.calls.extract} & \R{run.req.extract} \\
Proposer & questions rules cannot settle, with the facts & entity, value, cause, confidence & 8k & \R{run.calls.propose} & \R{run.req.propose} \\
Notice reply & notice, committed draft, its continuity rows & approve / adjust / reject & 32k & \R{run.calls.notice} & \R{run.req.notice} \\
\bottomrule
\end{tabular}
\end{table*}

\paragraph{Per-film internals.} \Cref{tab:internals} gives the planner's internals per film.

\begin{table}[h]
\centering
\caption{Planner internals per film. Scenes are fixed by the screenplay. Minutes = planned screen time, rounded per film; the total is rounded from the summed screen time, so the rounded rows need not add up to it. First pass = share of windows committed on the first draft. Drafts = mean drafts per window. Open = windows committed with open issues.}
\label{tab:internals}
\scriptsize
\setlength{\tabcolsep}{2pt}
\begin{tabular}{@{}lrrrrrrr@{}}
\toprule
Film & Scenes & Windows & Shots & Minutes & 1st pass & Drafts & Open \\
\midrule
Smalltime & 53 & \R{run.film.smalltime.windows.cont} & \R{run.film.smalltime.shots.cont} & \R{run.film.smalltime.minutes.cont} & \R{run.film.smalltime.firstpass.cont} & \R{run.film.smalltime.drafts.cont} & \R{run.film.smalltime.open.cont} \\
Jockey & 70 & \R{run.film.jockey.windows.cont} & \R{run.film.jockey.shots.cont} & \R{run.film.jockey.minutes.cont} & \R{run.film.jockey.firstpass.cont} & \R{run.film.jockey.drafts.cont} & \R{run.film.jockey.open.cont} \\
Step Back, D.\,C. & 64 & \R{run.film.stepback.windows.cont} & \R{run.film.stepback.shots.cont} & \R{run.film.stepback.minutes.cont} & \R{run.film.stepback.firstpass.cont} & \R{run.film.stepback.drafts.cont} & \R{run.film.stepback.open.cont} \\
Detail of a Life & 79 & \R{run.film.detail.windows.cont} & \R{run.film.detail.shots.cont} & \R{run.film.detail.minutes.cont} & \R{run.film.detail.firstpass.cont} & \R{run.film.detail.drafts.cont} & \R{run.film.detail.open.cont} \\
Big Time Adol. & 103 & \R{run.film.bta.windows.cont} & \R{run.film.bta.shots.cont} & \R{run.film.bta.minutes.cont} & \R{run.film.bta.firstpass.cont} & \R{run.film.bta.drafts.cont} & \R{run.film.bta.open.cont} \\
The Civil Dead & 128 & \R{run.film.civildead.windows.cont} & \R{run.film.civildead.shots.cont} & \R{run.film.civildead.minutes.cont} & \R{run.film.civildead.firstpass.cont} & \R{run.film.civildead.drafts.cont} & \R{run.film.civildead.open.cont} \\
Wildlife & 143 & \R{run.film.wildlife.windows.cont} & \R{run.film.wildlife.shots.cont} & \R{run.film.wildlife.minutes.cont} & \R{run.film.wildlife.firstpass.cont} & \R{run.film.wildlife.drafts.cont} & \R{run.film.wildlife.open.cont} \\
Kin & 180 & \R{run.film.kin.windows.cont} & \R{run.film.kin.shots.cont} & \R{run.film.kin.minutes.cont} & \R{run.film.kin.firstpass.cont} & \R{run.film.kin.drafts.cont} & \R{run.film.kin.open.cont} \\
\midrule
Fast Color & 196 & \R{run.film.fastcolor.windows.cont} & \R{run.film.fastcolor.shots.cont} & \R{run.film.fastcolor.minutes.cont} & \R{run.film.fastcolor.firstpass.cont} & \R{run.film.fastcolor.drafts.cont} & \R{run.film.fastcolor.open.cont} \\
All nine & \R{run.scenes} & \R{run.windows.cont} & \R{run.shots.cont} & \R{run.minutes.cont} & \R{run.firstpass.cont} & \R{run.drafts.cont} & \R{run.openissues.cont} \\
\bottomrule
\end{tabular}
\end{table}

\section{Evaluation details}
\label{sec:supp_eval}

Two question sets are scored from the shot requests: \speobj, the main test (\cref{sec:supp_objects}), and \spe, the second (\cref{sec:supp_spe}). Both draw their truth from the same script sheets and are scored by the same judges.

\paragraph{Script sheets.} Two readers, GPT-5.4 and Claude Opus 5.5, annotate each screenplay independently from its text alone, seeing no answer of the other and no plan, pair or render. The \emph{inventory} layer covers every scene (cast with wardrobe, hair and age quotes; props, state and holder; place; time and light; intensity; causes; events; relationship changes). The \emph{probe} layer covers 21--40 scenes per screenplay in finer closed vocabularies (emotion with intensity, prop state, size and placement, place features, light, palette and warmth, relationship stance, atomic statements). A fixed rule merges the readers: an agreed field keeps its value; a quote, list, event or statement field both filled keeps both readings, a quote only if found verbatim in the scene; every other field reads as ``the script does not say''. No person adjudicates. On inventory fields both readers filled, they agree in \R{sheet.agree.pct}\% of \R{sheet.agree.n} cases.

\paragraph{Judges and scoring.} Each judge sees only the shot requests and attached stills (at most 512\,px) of the two scenes, with a cast key from the neutral tokens of the plan's own characters to their names and no system name. The two scenes are labelled Scene 1 and Scene 2 in a seeded random order, and question order is drawn per judge from a fixed seed; a scene with no planned shots is shown as such. Answers are yes, no or ``can't tell'', with a closed list of options where the question has one. The jury answer is the majority of at least two of Claude Opus 5.5, Claude Sonnet 5.5 and GPT-6.1 Sol; ``can't tell'', no majority, fewer than two answering jurors and a missing plan count as wrong. Gemini 4 Argon answers alone and is reported apart; a cell with no answer gets one resend. Per-judge results are in \cref{sec:supp_judges}.

\paragraph{Cast tokens.}\label{sec:supp_tokens} One naming rule, the same code for every system, replaces character names in each plan's requests, so the tokens depend on each plan's names. Every speaker cue of the screenplay gets a token in order of first appearance, and each spelling gets its own (\emph{Big Time Adolescence} cues Mo as MO and as MONROE, two tokens). A cue is matched to a plan character when it is a whole-word run inside that character's name; a plan character whose name contains no cue gets a spare token (\method names Zeke ``Isaac Presanti''). The judges' cast key lists only the tokens that the plan's own characters hold, so a cue token that no plan character holds can appear in a request without being in the key, and one person can carry two tokens. Role cues that never appear in lower case also replace ordinary words in descriptions (``an Character M''). Questions use the screenplay's names, not tokens. A free check after judging found a token missing from the key in 101 of 1{,}106 judged scene pairs (one system's requests for one pair; 64 of 908 on \speobj, 37 of 198 on \spe), in every system and in the \ceiling: \method 19 and 9 (\speobj and \spe), StoryEngine 9 and 4, CineCrew 4 and 2, ViMax 27 and 13, VideoMemory 5 and 2, the \ceiling 7 on \spe. In 31 of them one person appears under two tokens, mostly Mo and, for \method, Zeke. Scoring the questions that name an affected person at each system's own rate on its other questions changes no system's overall score by more than about 1 point; scoring every question of the affected pairs that way changes them by at most 2.2 points, except ViMax on \spe ($-5.8$; it scores higher on these pairs than on the rest). \method stays first under both, and its leads hold even if every question naming an affected person goes against it and for the baselines (on \speobj, even if every question of the affected pairs does); only StoryEngine and CineCrew, under a point apart on \spe, swap places. Of the 31 two-token pairs, 20 are \method's. We did not rejudge.

\subsection{\speobj: object continuity (main test)}
\label{sec:supp_objects}

\speobj asks whether a plan keeps the same object when it returns in a later scene, keeps two different objects of one kind apart, and shows a cause and its effect when they run through an object. It tests carried object state, not causal reasoning in general.

\paragraph{Construction.} \emph{Inputs}: the script sheets, agreed cells only (a prop present in the scene, its holder quote, a character present), both readers' cause lists, and each scene's screenplay text. \emph{Candidates} (code): an object's kind is the last word of its sheet name, with synonyms folded (gun and revolver to pistol, bicycle to bike); set dressing is dropped by a fixed list. A candidate is two consecutive appearances of one kind, 2--40 scenes apart, with at least one character on screen in both scenes. The lower bound puts the two scenes in different windows; the upper bound keeps the object plausibly present. This gives 214 candidates. \emph{Truth}: two blind readings by Claude Opus 5.5, with the same instructions, see only the candidate and both scenes' text; each says same, different or unclear, with short object phrases, an owner and verbatim quotes of at most 25 words. A candidate is kept only when both readings give the same verdict, same or different, and all four quotes are found in the scene text (whitespace and case folded). The readings agree on 210 of 214: \R{spo.n.same} same, \R{spo.n.diff} different, 30 unclear (dropped). Every agreed object is kept, with no cap. \emph{Cause-effect items} (code): one reader's cause row at a late scene is used when the other reader names the same cause scene, the two scenes are 2--40 apart, and the cause or effect names an object kind the sheet marks present; at most one link per screenplay, longest gap first, giving \R{spo.n.links} links and \R{spo.n.cause} items. \emph{The set}: \R{spo.pairs} scene pairs over the nine screenplays, \R{spo.gap.min}--\R{spo.gap.max} scenes apart (median \R{spo.gap.median}). A pair can hold more than one scored item, so the \R{spo.pairs} pairs hold \R{spo.n.all} items: \R{spo.pairs.one} pairs hold one identity question, \R{spo.pairs.twoobj} hold two objects (two identity questions) and \R{spo.pairs.cause} hold one cause-effect link (a cause and an effect question).

\paragraph{Questions.} Each kept object adds one identity question and two presence gates (\cref{tab:obj_kinds}). The identity question names only the object kind, with no owner, description or quote, since the readers' phrases (``Clay's phone'' against ``Whitney's phone'') would let a judge answer from the question instead of the plan. Cause and effect items ask whether the early scene shows the cause and the late scene the effect, in the sheet readers' words; both truths are yes. The set has \R{spo.n.all} scored questions and \R{spo.n.gates} presence gates.

\begin{table}[h]
  \centering
  \caption{\speobj question kinds (\R{spo.n.all} scored questions, nine screenplays). Presence rows are unscored gates. Wording from the frozen files; \{early\} and \{late\} are filled with the scene labels the judge sees.}
  \label{tab:obj_kinds}
  \scriptsize
  \setlength{\tabcolsep}{2pt}
  \begin{tabular}{@{}>{\raggedright\arraybackslash}p{1.6cm}r>{\raggedright\arraybackslash}p{4.9cm}c@{}}
    \toprule
    Kind & $n$ & Example & Truth \\
    \midrule
    Same object & \R{spo.n.same} & \emph{Kin} 5, 9: ``Is the bike seen in the \{late\} scene the very same bike as the one seen in the \{early\} scene (same make, colour, size and markings)?'' & yes \\
    Different objects (control) & \R{spo.n.diff} & \emph{Step Back, Doors Closing} 23, 28: the same wording about a phone, where the readers found two different phones & no \\
    Cause & \R{spo.n.links} & \emph{Kin} 14, 34: ``Does the \{early\} scene show this: `Eli dropped Hal's maglight in the chamber.'?'' & yes \\
    Effect & \R{spo.n.links} & same pair: ``Does the \{late\} scene show this: `Eli retrieves the maglight where it fell.'?'' & yes \\
    Presence (gate) & \R{spo.n.gates} & ``Is a bike on screen in the \{early\} scene?'' & -- \\
    \bottomrule
  \end{tabular}
\end{table}

\paragraph{Gates and scoring.} An identity answer counts as correct only when the judge also answered yes to both presence gates, so a plan that never shows the object cannot score by default. The different-object controls (truth no) keep a planner that gives every phone one identity from being rewarded. A pair with no shots on either side, or a cell with no usable answer, counts as wrong. We report Same (\R{spo.n.same}), Different (\R{spo.n.diff}), Identity (their mean over the \R{spo.n.ident} identity questions), Cause (\R{spo.n.cause}) and All (\R{spo.n.all}, pooled). Intervals for differences are paired 95\% bootstrap intervals with 10{,}000 draws, in two units: over the nine films (films resampled, then pairs within each drawn film), as in the main text, and over pairs alone. The smallest detectable difference (80\% power, two-sided 5\%) is 2.8 times the film-bootstrap standard error. Where an interval spans zero, the difference is not distinguishable at this $n$.

\paragraph{Provenance.} A person designed the set and verified every scored answer against the screenplay. Its pairs and truths come from the script sheets and the screenplay text; no plan, \ledger, request or answer was read in building it, and it was frozen before any final plan was judged. It is a targeted test of carried object state, not a general measure.

\paragraph{Limits.} Truth for identity comes from two readings by one LLM, which one person then checked; no panel of people rated it, so no agreement between people is measured. The cause-effect items are the sheet readers' paraphrases, and some describe things a silent clip cannot show; they test whether the requests carry the cause and the effect, so we call them cause-effect coverage. The \R{spo.n.cause} items give a coarse rate: one item moves it by about six points. Results are in \cref{tab:speobj}, per judge in \cref{sec:supp_judges}, by gap in \cref{tab:supp_gapbins} and with the presence gates' pass rates in \cref{tab:supp_gates}.

\subsection{\spe on scene pairs (second test)}
\label{sec:supp_spe}

\begin{table}[h]
  \centering
  \caption{The four scored dynamics of \spe and their scored questions (\R{spe.n.all}). The frozen files also hold 44 causal questions, frozen and not scored (\cref{sec:spe}).}
  \label{tab:factors}
  \small
  \setlength{\tabcolsep}{3pt}
  \begin{tabular}{@{}lp{4.6cm}r@{}}
    \toprule
    Dynamic & Asks about & $n$ \\
    \midrule
    Character   & how two people act toward each other; change & \R{spe.n.char} \\
    Emotion     & strength of a person's emotion; which scene is stronger & \R{spe.n.emo} \\
    Colour      & warm / cool / neutral; light differs between scenes & \R{spe.n.col} \\
    Story state & identity, wardrobe, place state, props & \R{spe.n.state} \\
    \midrule
    All         & & \R{spe.n.all} \\
    \bottomrule
  \end{tabular}
\end{table}

\paragraph{Pair rule.} Code chooses pairs from the script sheets alone, with no model call, and reads no plan. Candidates are two probe scenes of one screenplay at least 20 scenes apart that share a principal character on screen, a prop in frame or a place. Each candidate is scored by building its questions; a question is cross-scene when it is about the pair as a whole or depends on a question about the other scene. Up to four scene-disjoint pairs per screenplay maximise, in lexicographic order: dynamics with a cross-scene question; cause-effect links; the smallest cross-scene count over dynamics; change questions whose answer is a change; pairs covering all five dynamics of the original design; the smallest scored count over dynamics; cross-scene questions; mean gap; scored questions; ties to the lowest scene numbers. The search is exact (branch and bound). The selection counted the causal dynamic; the pairs stay as frozen, and dropping causal from scoring changes no pair. Result on the nine screenplays: \R{spe.pairs} pairs, four per screenplay except three for \emph{Detail of a Life} and two for \emph{Smalltime}.

\paragraph{Questions.} Code writes every question from the sheet; truths come only from resolved values and only from what a silent clip can show. Presence questions gate the questions that presuppose them; every question offers ``can't tell''; a pair has at most 28 questions. Examples ([\,] = sheet value):
\begin{itemize}[nosep,leftmargin=*]
\item \emph{Character}: ``In the later scene, how do [A] and [B] act toward each other?'' (hostile, neutral, warm); also whether their treatment changes.
\item \emph{Emotion}: ``How strong is [person]'s emotion in the earlier scene?''; also which scene is stronger.
\item \emph{Colour}: ``Is the later scene's overall colour warm, cool or neutral?''; also whether light or colour clearly differs.
\item \emph{Story state}: ``Is [person] the same person in both scenes, judging only face and build?''; also wardrobe, place state and ``does [holder] hold the [prop]?''.
\item \emph{Distractor} (not scored): ``Does the later scene show this: [an event of another probe scene]?'' (truth no).
\end{itemize}

\paragraph{A worked pair.} \emph{Jockey}, scenes 6 and 66 (gap 60), both in the jockeys' room; 26 questions in the file, 4 of them distractors and 2 causal. Early: how do Jackson and Scott act toward each other (warm). Late: does Gabriel wear Ruth's racing silks (yes); is Jackson the same person in both scenes (yes). Distractor: does the late scene show Gabriel refusing the money (no).

\subsection{\Planaudit}
\label{sec:supp_audit}

\paragraph{Whole-plan reading.} Argon reads the whole screenplay and the whole blind plan in one call and scores nine criteria, 1--10: relationships, intensity, colour, causality, verification, fidelity, coverage, filmability and consistency, each with up to three evidence quotes. Code verifies every quote verbatim (whitespace and case folded, failing closed); a lost or invalid reading gets one resend. Each plan of \method is read three times and the mean is reported (mean standard deviation of the three readings \R{aud.read.sd}); each baseline plan is read once.

\paragraph{Plan checks.} The checks run over every scene against the inventory layer with one alias-aware name rule. Code computes character presence (recall of the sheet's cast per scene), relationships kept (after a recorded change, later scenes with both people that the plan also has both in) and causal links carried (an entity of the link in both scenes). Argon gives script events shown (shown rate on real events minus that on distractors, each ``shown'' with a verbatim quote). Values are in \cref{tab:supp_checks,tab:supp_reading}.

\section{\Ceiling details}
\label{sec:supp_ceiling}

A \ceiling is a plan that code writes from a set's answer key, with no LLM call, and that is judged by the same judges, prompts, order, seed and scoring as every system. Its score is what the instruments recover when every scored answer is stated in the requests; one minus its score is the judges' miss rate when every answer is stated explicitly. It does not estimate false positives on realistic plans, where answers are implicit or absent. It is a control, not a system. Each set has its own.

\paragraph{\spe \ceiling.} From the merged script sheet, code writes one entity per sheet entity, with looks and states dated from every scene the sheet covers (probe layer first, else the inventory layer), lasting damage, injuries and haircuts kept until undone, and shots for the probe scenes. Into each shot's text it writes every answer the questions ask: each character's emotion and its strength, the scene's intensity, each relationship with its turn and what one knows or hides, each prop's state and placement, the labelled time of day, light and colour, one look sentence per person, place and prop, and one sentence that every named person, place and object is the same throughout. It states none of the distractor events. Names become cast tokens by the same rule, and stills come from the same still rule and image model, as for every system. An earlier build put looks and states in entity stills rather than shot text and so was not a ceiling; it was rebuilt this way.

\paragraph{\speobj \ceiling.} Built from \speobj's answer key in the same way: each object pair's shot text shows the object on screen in both scenes, one registered object where the truth is same and two where it is different, and the cause and the effect lines in their scenes. It is judged with the same gates.

\paragraph{Audit of answers stated in text.} Before judging, code checks for every scored question whether the \ceiling's shot text states its truth (\cref{tab:ceiling_audit}).

\begin{table}[h]
  \centering
  \caption{\Ceiling audit: scored questions whose truth the control's shot text states, by kind and by dynamic.}
  \label{tab:ceiling_audit}
  \footnotesize
  \setlength{\tabcolsep}{3pt}
  \begin{tabular}{@{}lccccc@{}}
    \toprule
    \speobj & Same & Different & Cause & All & \\
     & \R{spo.n.same} & \R{spo.n.diff} & \R{spo.n.cause} & \R{spo.n.all} & \\
    \midrule
    Stated & \R{ceil.spo.stated.same} & \R{ceil.spo.stated.diff} & \R{ceil.spo.stated.cause} & \R{ceil.spo.stated} & \\
    \midrule
    \spe & Char. & Emo. & Col. & Story & All \\
     & \R{spe.n.char} & \R{spe.n.emo} & \R{spe.n.col} & \R{spe.n.state} & \R{spe.n.all} \\
    \midrule
    Stated & \R{ceil.spe.stated.char} & \R{ceil.spe.stated.emo} & \R{ceil.spe.stated.col} & \R{ceil.spe.stated.state} & \R{ceil.spe.stated} \\
    \bottomrule
  \end{tabular}
\end{table}

\paragraph{Scores and residual misses.} The \ceiling scores \R{spo.all.ceil.jury} / \R{spo.all.ceil.argon} on \speobj and \R{spe.all.ceil.jury} / \R{spe.all.ceil.argon} on \spe (jury / Argon; per kind and dynamic in \cref{tab:main}). The audit finds every cause-effect answer stated (\cref{tab:ceiling_audit}), yet the jury scores the \ceiling \R{spo.cause.ceil.jury} on those \R{spo.n.cause} items (Argon \R{spo.cause.ceil.argon}): the jury misses \R{ceil.spo.miss.cause.jury} stated answers, an instrument error of that size on this kind, which can lower any system's cause-effect rate as well. Every question it misses is sorted by cause. The kinds seen are: decoy-like rows where the judge says yes to something the text does not state; ``can't tell'' on questions the text answers; ``same person'' questions, which ask for face and build and so need stills that match across scenes, not text; a still that contradicts the text; and refused calls.

\section{Baselines}
\label{sec:supp_baselines}

Every planner reads the same screenplay text and scene index, with one neutral study brief, and writes a shot plan; only the plan is scored. Each baseline ran from its released code through an adapter of ours: the released agent classes run from source, and only the chat model is replaced by our caller, on Gemini 3.8 Flash. Each (system, screenplay) got one attempt; a stopped run could be resumed, never restarted. A per-system exporter maps each native plan to the shared plan schema, and reference stills are drawn by one image model from each system's own descriptions. The four baselines plan the same \R{run.scenes} scenes.

FilmWorld~\cite{filmworld} has no released code, and MovieAgent~\cite{movieagent} takes a synopsis and a character bank, not a screenplay, so neither is run.

\begin{table*}[t]
  \centering
  \caption{\textbf{Each adaptation, its reason and its likely effect on our scores.} Why: \emph{plan-only}, only plans are scored, so stages that make image or video calls do not run; \emph{limit}, of the released code; \emph{parity}, one setting for every system; \emph{ours}, our adapter's choice. ViMax and VideoMemory in \cref{tab:supp_adapt2}. Effects are not measured unless a number is given (jury, \cref{sec:supp_why,tab:main}; \speobj same-object unless stated); effects on the plan audit (\cref{sec:supp_audit}) are not estimated.}
  \label{tab:supp_adapt}
  \scriptsize
  \setlength{\tabcolsep}{3pt}
  \renewcommand{\arraystretch}{1.0}
  \begin{tabular}{@{}>{\raggedright\arraybackslash}p{6.0cm}>{\raggedright\arraybackslash}p{3.4cm}>{\raggedright\arraybackslash}p{7.6cm}@{}}
    \toprule
    Change & Why & Likely effect on our scores \\
    \midrule
    \multicolumn{3}{@{}l}{\emph{All four baselines}} \\
    Gemini~3.8~Flash replaces each system's own LLM (StoryEngine's paper: GPT-5.5; ViMax's paper: Gemini-3-Flash; VideoMemory's code: GPT-5.2; CineCrew's code: GPT-5); medium thinking for every baseline, our setting (\method's drafter: low) & parity & Unknown; no run on the published LLMs. \\
    One neutral brief in each system's own instruction input; a scene label before each scene, shot ids tied to it & parity; labels place shots on scenes & Neutral by design. \\
    One attempt per screenplay; one resend of a lost or invalid call replaces native retry loops, except StoryEngine's correction loop & parity & Slightly hurts CineCrew and ViMax, whose released code would re-ask some failed calls (rows below). \\
    Reference stills drawn by one image model from each plan's own descriptions; names replaced by neutral tokens & plan-only & A system that writes no description of an entity gets no still of it (VideoMemory's people and props, ViMax's props and places). \\
    \midrule
    \multicolumn{3}{@{}l}{\emph{StoryEngine}} \\
    Screenplay cut at scene boundaries into independent requests (windows) within the released caps, 20{,}000 characters and 64 shots; no state between windows & limit (one request, no state between requests) & Hurts identity across windows: right on \R{why.se.samewin.right} within one window, \R{why.se.otherwin.right} across (object named in both scenes). Carrying state between windows did not raise its scores on two screenplays (\cref{sec:supp_secarried}). \\
    Film target of 1{,}080 five-second shots, split across windows by text length; it makes the 64-shot cap bind first, so windows are far shorter than 20{,}000 characters & ours (the brief's 5{,}400-s runtime in five-second shots) & Probably hurts: more windows, so more pairs fall across one; size not measured. \\
    One outline call of ours over the whole screenplay (synopsis, events, principal characters and places; no prop list), appended to every window; outline caps raised for the three screenplays it first overran & ours (whole-film context the released code lacks) & Helps keep names and places across windows; little help on props. Not measured. \\
    Released three-attempt correction loop kept per window; a window failing all three is dropped and later windows run & limit (released code stops the whole request) & Neutral against released code; failed windows leave \R{why.empty.se} of \R{run.scenes} scenes empty, \R{why.se.absentempty} of its wrong answers. \\
    Planning prompt still states a video-reference limit of 0 (entities visible from the first frame); drafts compile at the released default, 16 & ours (first-run setting, kept so the rerun replays saved answers) & Probably slight harm: entities may be left out of later beats. Not measured. \\
    Requests written by our shared assembly: heading, action, camera, light, people, place, props; no dialogue & plan-only (its video prompt is written at execution) & Unknown direction; on the scenes it covers, colour \R{spe.covscore.col.se.jury} against \method's \R{spe.covscore.col.cont.vsse.jury} (\spe). \\
    \midrule
    \multicolumn{3}{@{}l}{\emph{CineCrew}} \\
    Split at scene boundaries into parts of at most 20{,}000 characters, each with its own art department and prop library, and segments of at most 10{,}000 & limit (released code truncates to 20{,}000 and 10{,}000); split, not truncation, is ours & Helps coverage (\R{why.empty.cc} empty scenes; truncation would leave most empty); hurts identity across parts: \R{why.cc.samepart.right} within one part, \R{why.cc.otherpart.right} across. \\
    Technical director off; our shared assembly writes its requests from its art, story and staging outputs & ours (about one LLM call a shot); recorded only as plan-only, but it is an LLM text stage the released plan-only default runs & Hurts CineCrew: with all four stages on it scored higher on both sets on two screenplays, still below \method (\cref{sec:supp_ccfull}). \\
    DSL validator off: deterministic checks of staging and speaker ids against the library, with an LLM repair of broken staging & ours (its repair calls the LLM) & None on ids: no staged id is unknown to its library (our recount). In our adapter its repair would have been asked to restage the shots failed staging calls left unstaged (about a fifth; released code fills placeholder staging or stops); shots leaving the object out give \R{why.cc.absentshots} of its wrong answers, a bound; in the two-screenplay check it restaged all of them (\cref{sec:supp_ccfull}). \\
    Voice-over director and dailies reviewer off (emotion and dialogue fill; critique of the technical director's prompts) & ours (calls per shot) & None on request scores while the technical director is off: our assembly reads no dialogue or emotion, and the reviewer has no prompt to review. \\
    Native retries (two a call) and invented fallback staging off & parity; nothing invented & Slight harm: failed calls got one resend, not two retries; size not measured. \\
    \bottomrule
  \end{tabular}
\end{table*}
\begin{table*}[t]
  \centering
  \caption{\textbf{Each adaptation, continued} (\cref{tab:supp_adapt}).}
  \label{tab:supp_adapt2}
  \scriptsize
  \setlength{\tabcolsep}{3pt}
  \renewcommand{\arraystretch}{1.0}
  \begin{tabular}{@{}>{\raggedright\arraybackslash}p{6.0cm}>{\raggedright\arraybackslash}p{3.4cm}>{\raggedright\arraybackslash}p{7.6cm}@{}}
    \toprule
    Change & Why & Likely effect on our scores \\
    \midrule
    \multicolumn{3}{@{}l}{\emph{ViMax}} \\
    Its multi-scene route: characters extracted once from the whole screenplay, then each scene planned alone with that list; novel pipeline not used & limit (its screenplay entry point plans one scene; the novel pipeline loses scene numbers) & Close to released code; one look per character for the film is ViMax's own design. \\
    Released three-attempt retries and camera-tree retry replaced by one resend (none after its character-index check) & parity & Slight harm: about half of its \R{why.empty.vimax} empty scenes come from invalid answers the released code would re-ask; empty scenes give \R{why.vimax.absentempty} of its wrong answers, which bounds the effect. \\
    Requests: its own video-prompt text (motion and sound lines); its first- and last-frame descriptions, from which its released route draws frames the video model also gets, are not read; no prop or place fields & plan-only (the frames are image calls); not reading the frame text is ours & Probably hurts on presence: those descriptions carry the setting and placement. Not measured. The missing prop fields are the system's own. \\
    \midrule
    \multicolumn{3}{@{}l}{\emph{VideoMemory}} \\
    Memory stage not run: no appearance text, no prop or scene descriptions, no stills of people or props & plan-only (one image call per asset); a run with the image tool stubbed (the code ships a mock) was possible, not attempted & Probably hurts, most on identity; not measured. Its scores, as context: same-object \R{spo.same.vmem.jury} against ViMax \R{spo.same.vimax.jury}, StoryEngine \R{spo.same.se.jury}, CineCrew \R{spo.same.cc.jury}; character \R{spe.char.vmem.jury} (\spe). Its memory reads only the storyboard and holds props fixed: effect on different-object items unknown. \\
    Visualisation stage not run, with its LLM rewrite of each shot into the video prompt (a person label, the film's style) & plan-only (image and video calls); its text could have run with the shipped mock tools, not attempted & Probably hurts on identity and colour. Not measured. \\
    Request is its video-shot text only; its keyframe text, key props and film style are not read & ours (pre-registered route) & Hurts on presence: its keyframe text names key props more often than the text we score; shots that leave the object out give \R{why.vmem.absentshots} of its wrong answers. \\
    Four shots a scene requested (code default: 8 a script) & ours (so every scene can get shots) & Helps coverage: at eight a script most scenes would have no shots. It plans fewer shots a scene than StoryEngine and ViMax, which may lower presence; not measured. \\
    Script split at scene boundaries into chunks run without state, when planned output at an assumed shot size would pass half the output cap & ours (follows from four shots a scene; released code makes one call) & Small next to the missing memory; more chunk boundaries than needed. Not measured. \\
    Shots placed on scenes by the place words of their headings; shots our matcher cannot place dropped; names merged across chunks & ours (it writes no scene ids) & Hurts where the matcher misses (about one shot in twelve dropped); its \R{why.empty.vmem} empty scenes, from misses or scenes it gave no shots, give \R{why.vmem.absentempty} of its wrong answers; a merged prop id, no still behind it, is right on \R{spo.carry.vmem.right.jury} of \R{spo.carry.vmem.n} items. \\
    Place stills drawn from heading place and time & ours (memory off) & Slightly helps on places against no still. \\
    \bottomrule
  \end{tabular}
\end{table*}

\paragraph{The rerun.} The first run lost output to faults of ours, listed per baseline below. All four baselines were rerun for those faults only, and every baseline number in this paper comes from the rerun plans. Every saved call was replayed, and only the calls lost to our faults and the calls that follow from them were sent; identical reference stills were reused. Faults that are a baseline's own were not fixed: schema or compile rejections after its last attempt (StoryEngine's third; for the others the study's one resend, though CineCrew's and ViMax's released code would retry), answers that looped, and prompts or answers the LLM blocked. The rerun was not planned in advance: we ran it after a breakdown of the first run's results showed how much output the baselines had lost to our faults.

\paragraph{StoryEngine~\cite{storyengine}.} Adaptations in \cref{tab:supp_adapt}. The released planner takes one request of at most 20{,}000 characters and 64 shots and keeps no state between requests, so our adapter cuts the screenplay at scene boundaries and drafts each window as an independent StoryEngine story; under our 1{,}080-shot film target the 64-shot cap binds first, so its windows are far shorter than 20{,}000 characters. The window size, the outline and the absence of state between windows are our adapter's choices, so the cross-window losses of \cref{sec:supp_why} describe StoryEngine as we ran it, not as published. First-run faults of ours: on three screenplays the added outline call overran a length cap we set, before StoryEngine's own planner started; our adapter compiled drafts with a video-reference limit of 0 against the released default of 16; one screenplay never got its second and third correction attempts. Rerun: those screenplays re-planned with the cap raised, saved drafts compiled with the default limit, the missing attempts given; the planning prompt still states the limit of 0, so saved answers replay unchanged.

\paragraph{CineCrew~\cite{cinecrew}.} Adaptations in \cref{tab:supp_adapt}. Nothing crosses a part boundary, because the art department takes no earlier library; this is our adapter's split, not CineCrew as published. Its technical director (which writes its image and video prompts), DSL validator, voice-over director and dailies reviewer are LLM text stages that make no image or video call, and its released default, itself plan-only, runs all four; we turned them off, so its requests are written by our shared prompt assembly, not by CineCrew, and its scores are those of CineCrew without its prompt writer. The validator checks the staging and speaker ids against the asset library and repairs broken staging; on our plans none of those ids is unknown to the library, so it would have changed only the unstaged shots (\cref{sec:supp_why}). First-run faults of ours: calls lost to a full disk, and a per-stage call cap we set too low, which let a resend use up a later unit's slot. Rerun: lost calls sent, cap doubled.

\paragraph{ViMax~\cite{vimax}.} Plan text only; adaptations in \cref{tab:supp_adapt,tab:supp_adapt2}. First-run fault of ours: we set its output cap to 16{,}384 tokens (its released agent runtime's default is 4{,}096), below the caps of the other systems, and thinking used most of it in the answers that stopped there. Rerun: cap raised to the model's maximum, 65{,}536, with those answers and the calls lost in transport sent again.

\paragraph{VideoMemory~\cite{videomemory}.} Storyboard stage only; adaptations in \cref{tab:supp_adapt,tab:supp_adapt2}. Its memory stage is an agent loop that writes a description of each character, prop and scene and draws a still of each; the descriptions are LLM text, and running the loop with its image tool stubbed could have given them, which we did not do. Without them its plans give people and props no appearance text, and no still of either can be drawn. First-run fault of ours: VideoMemory rewrites scene headings, and our export, which placed a shot only when it read the same place from both headings, dropped many of its shots. Rerun: shots placed by the place words of their headings, in order, with no model call; shots our matcher cannot place stay out. We make no claim against VideoMemory as published.

\paragraph{Comparability.} No number here compares with those in the baselines' papers. The LLM differs from the published one; the systems ran plan-only with their image and video stages off and, by our choice, some LLM text stages off too (\cref{tab:supp_adapt,tab:supp_adapt2}); the input is a whole feature screenplay, longer than the inputs they were built for; and our adapters, exporters and prompt assembly sit between each native plan and the measures. The rerun fixed only faults of ours; a different configuration might still do better.

\paragraph{Screenplays.} The nine are \emph{Smalltime}, \emph{Jockey}, \emph{Step Back, Doors Closing}, \emph{Detail of a Life}, \emph{Big Time Adolescence}, \emph{The Civil Dead}, \emph{Wildlife}, \emph{Kin} and \emph{Fast Color} (53--196 scenes; \cref{tab:internals}).

\section{Where the baselines lose questions}
\label{sec:supp_why}

This section says why each released planner loses \speobj questions that \method answers, and which part of \method's planner avoids each failure (\cref{sec:carry}).
A question can be lost in three ways.
\emph{Coverage}: a scene it asks about has no shots in the plan.
\emph{Not shown}: the scene has shots, but the judges do not find the object in them.
\emph{Identity}: the object is shown in both scenes, but the judges see two objects where the screenplay has one, or cannot tell.
\method's planner meets these with three of its parts: it commits every window, listing what it could not fix, so every scene has shots; each shot names the people and props it shows; and every draft gets the id and name of each person and prop registered earlier in the film, so an object that returns can keep one registered entry and description however far apart the scenes are.
\Cref{tab:why_cov} gives coverage, \cref{tab:why_split} splits each system's wrong \speobj answers by these causes, and \cref{tab:why_content} compares content on the questions whose scenes have shots in the baseline's plan, against \method on exactly those questions.
The analysis is free and reads only answers, requests and plans already on file; three independent verifiers recomputed it from the raw files.

\begin{table}[h]
  \centering
  \caption{Coverage, rerun plans of the baselines and \method's final plans. Scenes with no shots, of \R{run.scenes}; scored questions whose scenes have shots in the plan, of \R{spo.n.all} (\speobj) and \R{spe.n.all} (\spe).}
  \label{tab:why_cov}
  \scriptsize
  \setlength{\tabcolsep}{3pt}
  \begin{tabular}{@{}lccc@{}}
    \toprule
    & Scenes with & \multicolumn{2}{c}{Questions covered} \\
    \cmidrule(l){3-4}
    System & no shots & \speobj & \spe \\
    \midrule
    StoryEngine & \R{why.empty.se} & \R{spo.cov.se} & \R{spe.cov.se} \\
    CineCrew    & \R{why.empty.cc} & \R{spo.cov.cc} & \R{spe.cov.cc} \\
    ViMax       & \R{why.empty.vimax} & \R{spo.cov.vimax} & \R{spe.cov.vimax} \\
    VideoMemory & \R{why.empty.vmem} & \R{spo.cov.vmem} & \R{spe.cov.vmem} \\
    \midrule
    \method     & \R{why.empty.cont} & \R{spo.n.all} & \R{spe.n.all} \\
    \bottomrule
  \end{tabular}
\end{table}

\begin{table*}[t]
  \centering
  \caption{Where the wrong \speobj answers go (jury). Wrong: wrong answers of \R{spo.n.all}; then shares of them. Empty: a scene the answer is about has no shots. Not shown: the scene has shots and the judges do not find the object. Two objects: shown in both scenes and judged the opposite identity. Can't tell: shown in both, identity undetermined. Named: share of the scenes of an object pair whose requests name the object. One id: same-object items whose two scenes carry one prop id ($n$), right with it / without it.}
  \label{tab:why_split}
  \scriptsize
  \setlength{\tabcolsep}{2pt}
  \begin{tabular}{@{}lccccccc@{}}
    \toprule
    System & Wrong & Empty & Not shown & Two objects & Can't tell & Named & One id ($n$): with / without \\
    \midrule
    StoryEngine & \R{why.spo.all.se.wrong.jury} & \R{why.se.absentempty} & \R{why.se.absentshots} & \R{why.se.contra} & \R{why.se.undet} & \R{why.se.sidesnamed} & \R{spo.carry.se.n}: \R{spo.carry.se.right.jury} / \R{spo.carry.se.noid.right.jury} \\
    CineCrew & \R{why.spo.all.cc.wrong.jury} & \R{why.cc.absentempty} & \R{why.cc.absentshots} & \R{why.cc.contra} & \R{why.cc.undet} & \R{why.cc.sidesnamed} & \R{spo.carry.cc.n}: \R{spo.carry.cc.right.jury} / \R{spo.carry.cc.noid.right.jury} \\
    ViMax & \R{why.spo.all.vimax.wrong.jury} & \R{why.vimax.absentempty} & \R{why.vimax.absentshots} & \R{why.vimax.contra} & \R{why.vimax.undet} & \R{why.vimax.sidesnamed} & \R{spo.carry.vimax.n}: -- / \R{spo.carry.vimax.noid.right.jury} \\
    VideoMemory & \R{why.spo.all.vmem.wrong.jury} & \R{why.vmem.absentempty} & \R{why.vmem.absentshots} & \R{why.vmem.contra} & \R{why.vmem.undet} & \R{why.vmem.sidesnamed} & \R{spo.carry.vmem.n}: \R{spo.carry.vmem.right.jury} / \R{spo.carry.vmem.noid.right.jury} \\
    \midrule
    \method & \R{why.spo.all.cont.wrong.jury} & \R{why.cont.absentempty} & \R{why.cont.absentshots} & \R{why.cont.contra} & \R{why.cont.undet} & \R{why.cont.sidesnamed} & \R{spo.carry.cont.n}: \R{spo.carry.cont.right.jury} / \R{spo.carry.cont.noid.right.jury} \\
    \bottomrule
  \end{tabular}
\end{table*}

\begin{table*}[t]
  \centering
  \caption{Content per baseline: on the questions whose scenes have shots in the baseline's plan, baseline / \method on the same questions, jury and Argon in separate blocks.}
  \label{tab:why_content}
  \scriptsize
  \setlength{\tabcolsep}{2.5pt}
  \begin{tabular}{@{}llcccccccc@{}}
    \toprule
    & & \multicolumn{3}{c}{\speobj} & \multicolumn{5}{c}{\spe} \\
    \cmidrule(lr){3-5}\cmidrule(l){6-10}
    Judge & Baseline & Same & Different & All & Character & Emotion & Colour & Story state & All \\
    \midrule
    \multirow{4}{*}{Jury}
    & StoryEngine & \R{spo.covscore.same.se.jury} / \R{spo.covscore.same.cont.vsse.jury} & \R{spo.covscore.diff.se.jury} / \R{spo.covscore.diff.cont.vsse.jury} & \R{spo.covscore.all.se.jury} / \R{spo.covscore.all.cont.vsse.jury} & \R{spe.covscore.char.se.jury} / \R{spe.covscore.char.cont.vsse.jury} & \R{spe.covscore.emo.se.jury} / \R{spe.covscore.emo.cont.vsse.jury} & \R{spe.covscore.col.se.jury} / \R{spe.covscore.col.cont.vsse.jury} & \R{spe.covscore.state.se.jury} / \R{spe.covscore.state.cont.vsse.jury} & \R{spe.covscore.se.jury} / \R{spe.covscore.cont.vsse.jury} \\
    & CineCrew & \R{spo.covscore.same.cc.jury} / \R{spo.covscore.same.cont.vscc.jury} & \R{spo.covscore.diff.cc.jury} / \R{spo.covscore.diff.cont.vscc.jury} & \R{spo.covscore.all.cc.jury} / \R{spo.covscore.all.cont.vscc.jury} & \R{spe.covscore.char.cc.jury} / \R{spe.covscore.char.cont.vscc.jury} & \R{spe.covscore.emo.cc.jury} / \R{spe.covscore.emo.cont.vscc.jury} & \R{spe.covscore.col.cc.jury} / \R{spe.covscore.col.cont.vscc.jury} & \R{spe.covscore.state.cc.jury} / \R{spe.covscore.state.cont.vscc.jury} & \R{spe.covscore.cc.jury} / \R{spe.covscore.cont.vscc.jury} \\
    & ViMax & \R{spo.covscore.same.vimax.jury} / \R{spo.covscore.same.cont.vsvimax.jury} & \R{spo.covscore.diff.vimax.jury} / \R{spo.covscore.diff.cont.vsvimax.jury} & \R{spo.covscore.all.vimax.jury} / \R{spo.covscore.all.cont.vsvimax.jury} & \R{spe.covscore.char.vimax.jury} / \R{spe.covscore.char.cont.vsvimax.jury} & \R{spe.covscore.emo.vimax.jury} / \R{spe.covscore.emo.cont.vsvimax.jury} & \R{spe.covscore.col.vimax.jury} / \R{spe.covscore.col.cont.vsvimax.jury} & \R{spe.covscore.state.vimax.jury} / \R{spe.covscore.state.cont.vsvimax.jury} & \R{spe.covscore.vimax.jury} / \R{spe.covscore.cont.vsvimax.jury} \\
    & VideoMemory & \R{spo.covscore.same.vmem.jury} / \R{spo.covscore.same.cont.vsvmem.jury} & \R{spo.covscore.diff.vmem.jury} / \R{spo.covscore.diff.cont.vsvmem.jury} & \R{spo.covscore.all.vmem.jury} / \R{spo.covscore.all.cont.vsvmem.jury} & \R{spe.covscore.char.vmem.jury} / \R{spe.covscore.char.cont.vsvmem.jury} & \R{spe.covscore.emo.vmem.jury} / \R{spe.covscore.emo.cont.vsvmem.jury} & \R{spe.covscore.col.vmem.jury} / \R{spe.covscore.col.cont.vsvmem.jury} & \R{spe.covscore.state.vmem.jury} / \R{spe.covscore.state.cont.vsvmem.jury} & \R{spe.covscore.vmem.jury} / \R{spe.covscore.cont.vsvmem.jury} \\
    \midrule
    \multirow{4}{*}{Argon}
    & StoryEngine & \R{spo.covscore.same.se.argon} / \R{spo.covscore.same.cont.vsse.argon} & \R{spo.covscore.diff.se.argon} / \R{spo.covscore.diff.cont.vsse.argon} & \R{spo.covscore.all.se.argon} / \R{spo.covscore.all.cont.vsse.argon} & \R{spe.covscore.char.se.argon} / \R{spe.covscore.char.cont.vsse.argon} & \R{spe.covscore.emo.se.argon} / \R{spe.covscore.emo.cont.vsse.argon} & \R{spe.covscore.col.se.argon} / \R{spe.covscore.col.cont.vsse.argon} & \R{spe.covscore.state.se.argon} / \R{spe.covscore.state.cont.vsse.argon} & \R{spe.covscore.se.argon} / \R{spe.covscore.cont.vsse.argon} \\
    & CineCrew & \R{spo.covscore.same.cc.argon} / \R{spo.covscore.same.cont.vscc.argon} & \R{spo.covscore.diff.cc.argon} / \R{spo.covscore.diff.cont.vscc.argon} & \R{spo.covscore.all.cc.argon} / \R{spo.covscore.all.cont.vscc.argon} & \R{spe.covscore.char.cc.argon} / \R{spe.covscore.char.cont.vscc.argon} & \R{spe.covscore.emo.cc.argon} / \R{spe.covscore.emo.cont.vscc.argon} & \R{spe.covscore.col.cc.argon} / \R{spe.covscore.col.cont.vscc.argon} & \R{spe.covscore.state.cc.argon} / \R{spe.covscore.state.cont.vscc.argon} & \R{spe.covscore.cc.argon} / \R{spe.covscore.cont.vscc.argon} \\
    & ViMax & \R{spo.covscore.same.vimax.argon} / \R{spo.covscore.same.cont.vsvimax.argon} & \R{spo.covscore.diff.vimax.argon} / \R{spo.covscore.diff.cont.vsvimax.argon} & \R{spo.covscore.all.vimax.argon} / \R{spo.covscore.all.cont.vsvimax.argon} & \R{spe.covscore.char.vimax.argon} / \R{spe.covscore.char.cont.vsvimax.argon} & \R{spe.covscore.emo.vimax.argon} / \R{spe.covscore.emo.cont.vsvimax.argon} & \R{spe.covscore.col.vimax.argon} / \R{spe.covscore.col.cont.vsvimax.argon} & \R{spe.covscore.state.vimax.argon} / \R{spe.covscore.state.cont.vsvimax.argon} & \R{spe.covscore.vimax.argon} / \R{spe.covscore.cont.vsvimax.argon} \\
    & VideoMemory & \R{spo.covscore.same.vmem.argon} / \R{spo.covscore.same.cont.vsvmem.argon} & \R{spo.covscore.diff.vmem.argon} / \R{spo.covscore.diff.cont.vsvmem.argon} & \R{spo.covscore.all.vmem.argon} / \R{spo.covscore.all.cont.vsvmem.argon} & \R{spe.covscore.char.vmem.argon} / \R{spe.covscore.char.cont.vsvmem.argon} & \R{spe.covscore.emo.vmem.argon} / \R{spe.covscore.emo.cont.vsvmem.argon} & \R{spe.covscore.col.vmem.argon} / \R{spe.covscore.col.cont.vsvmem.argon} & \R{spe.covscore.state.vmem.argon} / \R{spe.covscore.state.cont.vsvmem.argon} & \R{spe.covscore.vmem.argon} / \R{spe.covscore.cont.vsvmem.argon} \\
    \bottomrule
  \end{tabular}
\end{table*}

\paragraph{StoryEngine: identity lives inside one draft window, and failed windows leave scenes empty.}
Our adapter has StoryEngine draft the screenplay in independent windows of at most 20{,}000 characters and 64 shots, each with one outline of the whole screenplay appended and no other state passed between windows (\cref{sec:supp_baselines}); its own entity ids are therefore per window, and the study's name linker joins two windows only where they happen to name a prop the same way. A window that fails its compile gate after the released code's three correction attempts is discarded, so its scenes have no shots (\R{why.empty.se} of \R{run.scenes} scenes; \R{why.se.absentempty} of its wrong answers).
On the same-object items whose requests name the object in both scenes (a subset of the \R{spo.unit.se.same.n} within-window and \R{spo.unit.se.other.n} cross-window items of \cref{sec:carry}, jury), it is right on \R{why.se.samewin.right} when both scenes fall in one window (\R{why.se.samewin.n} items) and on \R{why.se.otherwin.right} when they fall in different windows (\R{why.se.otherwin.n}), where \R{why.se.otherwin.contra} are judged two different objects; the split holds at matched gap (gap of 4 or less: \R{why.se.samewin4.right}, $n$ \R{why.se.samewin4.n}, against \R{why.se.otherwin4.right}, $n$ \R{why.se.otherwin4.n}; 5 or more: \R{why.se.samewin5.right}, $n$ \R{why.se.samewin5.n}, against \R{why.se.otherwin5.right}, $n$ \R{why.se.otherwin5.n}).
On all same-object items, named or not (\R{spo.unit.se.same.n} and \R{spo.unit.se.other.n}), \method scores \R{spo.unit.se.same.cont.jury} within one StoryEngine window and \R{spo.unit.se.other.cont.jury} across windows, against StoryEngine's \R{spo.unit.se.same.se.jury} and \R{spo.unit.se.other.se.jury} (jury; differences \R{spo.unit.se.same.lead.jury} [\R{spo.unit.se.same.lead.jury.cl.lo}, \R{spo.unit.se.same.lead.jury.cl.hi}] within and \R{spo.unit.se.other.lead.jury} [\R{spo.unit.se.other.lead.jury.cl.lo}, \R{spo.unit.se.other.lead.jury.cl.hi}] / \R{spo.unit.se.other.lead.argon} [\R{spo.unit.se.other.lead.argon.cl.lo}, \R{spo.unit.se.other.lead.argon.cl.hi}] across, jury / Argon).
\emph{What avoids it in \method:} the film-wide registry hands every window the ids of everything registered earlier, and a window that fails its checks is committed with its open issues listed instead of being dropped.
On the \spe questions its shots cover, StoryEngine scores \R{spe.covscore.se.jury} against \method's \R{spe.covscore.cont.vsse.jury}; on the covered \speobj same-object questions, \R{spo.covscore.same.se.jury} / \R{spo.covscore.same.se.argon} against \R{spo.covscore.same.cont.vsse.jury} / \R{spo.covscore.same.cont.vsse.argon} (\cref{tab:why_content}).

\paragraph{CineCrew: one prop library per part, and unstaged shots.}
In our adapter CineCrew builds its asset library per screenplay part of at most 20{,}000 characters, and nothing crosses a part boundary (\cref{sec:supp_baselines}). Its plans cover nearly every scene (\R{why.empty.cc} empty), so its losses are content: \R{why.cc.absentshots} of its wrong answers are scenes whose shots leave the object out (about a fifth of its plan shots were left unstaged and list no prop, place or cast), and \R{why.cc.undet} are objects shown in both scenes whose identity the judges cannot tell.
On the same-object items whose requests name the object in both scenes (a subset of the \R{spo.unit.cc.same.n} within-part and \R{spo.unit.cc.other.n} cross-part items of \cref{sec:carry}, jury), it is right on \R{why.cc.samepart.right} within one part (\R{why.cc.samepart.n}) and on \R{why.cc.otherpart.right} across parts (\R{why.cc.otherpart.n}; few items, so a coarse estimate). When both scenes do carry one of its prop ids it is right on \R{spo.carry.cc.right.jury} (\R{spo.carry.cc.n} items), but that happens at 16--39 scenes on only \R{spo.carry.cc.16to39.n} of \R{spo.gap.16to39.n.same} items.
On all same-object items, \method scores \R{spo.unit.cc.other.cont.jury} across CineCrew's parts against CineCrew's \R{spo.unit.cc.other.cc.jury} (jury; difference \R{spo.unit.cc.other.lead.jury} [\R{spo.unit.cc.other.lead.jury.cl.lo}, \R{spo.unit.cc.other.lead.jury.cl.hi}], Argon \R{spo.unit.cc.other.lead.argon} [\R{spo.unit.cc.other.lead.argon.cl.lo}, \R{spo.unit.cc.other.lead.argon.cl.hi}]); within one part, CineCrew \R{spo.unit.cc.same.cc.jury} against \R{spo.unit.cc.same.cont.jury} (\R{spo.unit.cc.same.n} items).
\emph{What avoids it in \method:} one registry for the whole film, and shots that name the props they show.
On colour it scores \R{spe.col.cc.jury} / \R{spe.col.cc.argon}, against \method's \R{spe.col.cont.jury} / \R{spe.col.cont.argon}.

\paragraph{ViMax: no prop fields.}
ViMax extracts the characters once for the whole screenplay and plans every scene on its own from that list; it writes no prop or place fields, so no prop has an id, a description or a still, and an object exists in a request only if a shot's motion or sound line mentions it.
Its requests name the object in \R{why.vimax.sidesnamed} of the scenes of an object pair, and it shows the object in both scenes on \R{spo.gate.same.vimax.jury} of the same-object items, but it is right on only \R{spo.gateacc.same.vimax.jury} of those: on \R{why.vimax.undet} of its wrong answers the judges see the object in both scenes and cannot tell whether it is the same one.
\emph{What avoids it in \method:} every prop is registered with an id, a description and a still, and the drafter is told to reuse a returning prop's earlier id.

\paragraph{VideoMemory: camera plans without props, and empty scenes.}
We ran only VideoMemory's storyboard stage, once per screenplay chunk with nothing carried between chunks; its storyboard has no prop descriptions, and the request we score, its video-shot text, is a timed camera plan that seldom names props (its keyframe text, which we do not score, names them more often; \cref{tab:supp_adapt2}). Of its wrong answers, \R{why.vmem.absentempty} come from scenes with no shots (\R{why.empty.vmem} empty scenes) and \R{why.vmem.absentshots} from shots that leave the object out; its requests name the object in \R{why.vmem.sidesnamed} of the scenes of an object pair. Our export merges its key-prop names across chunks, so a prop id it shares across scenes is a name with no description or still behind it, and it is right on \R{spo.carry.vmem.right.jury} of the \R{spo.carry.vmem.n} items that share one. Its plans also carry no appearance text, so no character still can be drawn for it, a disadvantage of the stage we ran, not of the published system.
\emph{What avoids it in \method:} every scene is planned, and every shot names the people and props it shows by their registered entries.

\subsection{StoryEngine with state carried between windows}
\label{sec:supp_secarried}

Our adapter passes no state between StoryEngine's windows (\cref{tab:supp_adapt}), so after the main results we asked whether carrying state between them changes its scores.
The variant is StoryEngine exactly as in the main run (same windows, outline, prompts, Gemini~3.8~Flash at medium thinking, released three-attempt correction loop and exporter), except that every window's draft request also gets the previous compiled window's last five shots and the world state at its end, marked as context, not new plot text, as in an earlier feature-length adapter of ours; windows run in order, and after a window that fails all three attempts the next continues from the last compiled window's state.
By a rule fixed before any output of this check, it ran on the two of the seven screenplays not used in development with the most same-object items whose two scenes fall in different windows of the main StoryEngine run, \emph{Fast Color} (16 of its 29 same-object items) and \emph{The Civil Dead} (12 of 17), and was judged exactly as the main run: the same frozen questions, reference stills drawn by the same image model from this plan's own descriptions, the same four judges and the same scoring, with missing scenes counting as wrong.
StoryEngine's own checks rejected 5 of the 38 windows after three attempts (main run on these screenplays: 2 of 38), leaving 31 of 196 and 16 of 128 scenes without shots (main run: 19 and 0), although the share of attempts rejected was about the same (48\% against 46\%).
Carrying state did not raise StoryEngine's scores (\cref{tab:supp_secarried}; carried minus main run, jury: $-4.5$ [$-15.2$, $+7.2$] points on \speobj and $-6.2$ [$-31.2$, $+18.4$] on \spe), and its same-object accuracy across windows stayed near the main run's and far below \method's, also on the pairs both runs planned.
With two screenplays and wide intervals this cannot rule out a gain on others; on the four \spe pairs both runs planned the carried run scored higher, and it lost more windows to StoryEngine's checks at about the same rate per attempt.

\begin{table}[t]
  \centering
  \caption{StoryEngine with state carried between windows (\cref{sec:supp_secarried}), pooled over \emph{Fast Color} and \emph{The Civil Dead}. Cells jury / Argon; $n$: scored questions. Across / within: same-object items whose two scenes fall in different windows / one window of the main StoryEngine run. Both planned: pairs whose two scenes have shots in both StoryEngine runs, every system scored on that set. Missing scenes count as wrong. Intervals in the text: 95\%, paired bootstrap over pairs resampled within each film.}
  \label{tab:supp_secarried}
  \footnotesize
  \setlength{\tabcolsep}{3pt}
  \begin{tabular}{@{}lrccc@{}}
    \toprule
    & & \multicolumn{2}{c}{StoryEngine} & \\
    \cmidrule(lr){3-4}
    & $n$ & Main run & Carried & \method \\
    \midrule
    \speobj, all & 67 & .52 / .55 & .48 / .49 & .63 / .69 \\
    \quad Same, across windows & 28 & .14 / .14 & .11 / .18 & .64 / .79 \\
    \quad Same, within a window & 18 & .78 / .78 & .78 / .78 & .56 / .56 \\
    \spe, all (8 pairs) & 128 & .75 / .75 & .69 / .72 & .88 / .91 \\
    \midrule
    \multicolumn{5}{@{}l}{\emph{Pairs both StoryEngine runs planned}} \\
    \speobj, all & 51 & .65 / .69 & .61 / .61 & .55 / .63 \\
    \quad Same, across windows & 16 & .25 / .25 & .19 / .25 & .44 / .69 \\
    \spe, all (4 pairs) & 62 & .84 / .85 & .90 / .95 & .87 / .87 \\
    \bottomrule
  \end{tabular}
\end{table}

\subsection{CineCrew with its text stages on}
\label{sec:supp_ccfull}

Our adapter turns off CineCrew's technical director, DSL validator, voice-over director and dailies reviewer, so our assembly writes its requests (\cref{tab:supp_adapt}); after the main results we asked what this costs it.
The variant is CineCrew as released in plan-only mode, with all four stages on in the released order and everything else as in the main run (parts, segments, brief, Gemini~3.8~Flash at medium thinking); the earlier stages' saved answers are replayed unchanged, and the new stages of different segments run in parallel, which changes no request.
The judges score its own technical director's request on every probe shot, and its validator restaged all 427 shots that its staging calls had left empty.
By a rule fixed before any output of this check, it ran on the two screenplays with the most CineCrew probe-scene shots, \emph{Wildlife} and \emph{Step Back, Doors Closing}, neither used in development, and was judged exactly as the main run.
With its stages on CineCrew scored higher (\cref{tab:supp_ccfull}; stages on minus main run, points: on \speobj $+5.2$ [$-5.4$, $+16.7$] under the jury and $+13.8$ [$+1.7$, $+26.3$] under Argon, mostly on different-object items; on \spe $+7.8$ under both, [$+2.9$, $+11.1$] and [$+2.5$, $+12.4$]), so turning the stages off cost it a few points on these screenplays.
\method stayed ahead: by $+27.6$ points on \speobj under both judges, and on \spe by $+7.8$ [$+1.3$, $+15.2$] (jury) and $+11.7$ [$+1.7$, $+24.8$] (Argon).
With two screenplays and a check run after the main results, this bounds the cost of the adaptation; CineCrew's rows elsewhere are the main run's.

\begin{table}[t]
  \centering
  \caption{CineCrew with its four text stages on (\cref{sec:supp_ccfull}), pooled over \emph{Wildlife} and \emph{Step Back, Doors Closing}. Cells jury / Argon; $n$: scored questions. Missing scenes count as wrong. Intervals in the text: 95\%, paired bootstrap over pairs resampled within each film.}
  \label{tab:supp_ccfull}
  \footnotesize
  \setlength{\tabcolsep}{3pt}
  \begin{tabular}{@{}lrccc@{}}
    \toprule
    & & \multicolumn{2}{c}{CineCrew} & \\
    \cmidrule(lr){3-4}
    & $n$ & Main run & Stages on & \method \\
    \midrule
    \speobj, all & 58 & .34 / .29 & .40 / .43 & .67 / .71 \\
    \quad Same object & 25 & .36 / .32 & .36 / .36 & .52 / .64 \\
    \quad Different object & 29 & .24 / .17 & .34 / .41 & .79 / .76 \\
    \spe, all (8 pairs) & 154 & .68 / .68 & .76 / .75 & .84 / .87 \\
    \bottomrule
  \end{tabular}
\end{table}

\section{Full results}
\label{sec:supp_results}

\method's rows come from the final engine's nine-film run; every baseline's rows come from its rerun plans (\cref{sec:supp_baselines}). Missing plans, unanswered cells and ``can't tell'' count as wrong. Cells are jury / Argon unless a caption says otherwise.

\begin{table*}[t]
  \centering
  \caption{\speobj per screenplay: All (gated), jury / Argon. Pairs, items: scene pairs in the frozen set and their scored items. $^\dagger$\emph{Smalltime} has \R{spo.film.smalltime.pairs} pairs and \R{spo.film.smalltime.n} items, so one item moves a cell by a third and even the \ceiling scores \R{spo.film.smalltime.ceil.jury} under the jury; its cells say nothing about the systems.}
  \label{tab:supp_perfilm_obj}
  \scriptsize
  \setlength{\tabcolsep}{3pt}
  \begin{tabular}{@{}lcccccc@{}}
    \toprule
    Screenplay (pairs, items) & StoryEngine & VideoMemory & ViMax & CineCrew & \method & \textit{\Ceiling} \\
    \midrule
    Step Back, Doors Closing (\R{spo.film.stepback.pairs}, \R{spo.film.stepback.n}) & \R{spo.film.stepback.se.jury} / \R{spo.film.stepback.se.argon} & \R{spo.film.stepback.vmem.jury} / \R{spo.film.stepback.vmem.argon} & \R{spo.film.stepback.vimax.jury} / \R{spo.film.stepback.vimax.argon} & \R{spo.film.stepback.cc.jury} / \R{spo.film.stepback.cc.argon} & \R{spo.film.stepback.cont.jury} / \R{spo.film.stepback.cont.argon} & \R{spo.film.stepback.ceil.jury} / \R{spo.film.stepback.ceil.argon} \\
    The Civil Dead (\R{spo.film.civildead.pairs}, \R{spo.film.civildead.n}) & \R{spo.film.civildead.se.jury} / \R{spo.film.civildead.se.argon} & \R{spo.film.civildead.vmem.jury} / \R{spo.film.civildead.vmem.argon} & \R{spo.film.civildead.vimax.jury} / \R{spo.film.civildead.vimax.argon} & \R{spo.film.civildead.cc.jury} / \R{spo.film.civildead.cc.argon} & \R{spo.film.civildead.cont.jury} / \R{spo.film.civildead.cont.argon} & \R{spo.film.civildead.ceil.jury} / \R{spo.film.civildead.ceil.argon} \\
    Jockey (\R{spo.film.jockey.pairs}, \R{spo.film.jockey.n}) & \R{spo.film.jockey.se.jury} / \R{spo.film.jockey.se.argon} & \R{spo.film.jockey.vmem.jury} / \R{spo.film.jockey.vmem.argon} & \R{spo.film.jockey.vimax.jury} / \R{spo.film.jockey.vimax.argon} & \R{spo.film.jockey.cc.jury} / \R{spo.film.jockey.cc.argon} & \R{spo.film.jockey.cont.jury} / \R{spo.film.jockey.cont.argon} & \R{spo.film.jockey.ceil.jury} / \R{spo.film.jockey.ceil.argon} \\
    Fast Color (\R{spo.film.fastcolor.pairs}, \R{spo.film.fastcolor.n}) & \R{spo.film.fastcolor.se.jury} / \R{spo.film.fastcolor.se.argon} & \R{spo.film.fastcolor.vmem.jury} / \R{spo.film.fastcolor.vmem.argon} & \R{spo.film.fastcolor.vimax.jury} / \R{spo.film.fastcolor.vimax.argon} & \R{spo.film.fastcolor.cc.jury} / \R{spo.film.fastcolor.cc.argon} & \R{spo.film.fastcolor.cont.jury} / \R{spo.film.fastcolor.cont.argon} & \R{spo.film.fastcolor.ceil.jury} / \R{spo.film.fastcolor.ceil.argon} \\
    Kin (\R{spo.film.kin.pairs}, \R{spo.film.kin.n}) & \R{spo.film.kin.se.jury} / \R{spo.film.kin.se.argon} & \R{spo.film.kin.vmem.jury} / \R{spo.film.kin.vmem.argon} & \R{spo.film.kin.vimax.jury} / \R{spo.film.kin.vimax.argon} & \R{spo.film.kin.cc.jury} / \R{spo.film.kin.cc.argon} & \R{spo.film.kin.cont.jury} / \R{spo.film.kin.cont.argon} & \R{spo.film.kin.ceil.jury} / \R{spo.film.kin.ceil.argon} \\
    Big Time Adolescence (\R{spo.film.bta.pairs}, \R{spo.film.bta.n}) & \R{spo.film.bta.se.jury} / \R{spo.film.bta.se.argon} & \R{spo.film.bta.vmem.jury} / \R{spo.film.bta.vmem.argon} & \R{spo.film.bta.vimax.jury} / \R{spo.film.bta.vimax.argon} & \R{spo.film.bta.cc.jury} / \R{spo.film.bta.cc.argon} & \R{spo.film.bta.cont.jury} / \R{spo.film.bta.cont.argon} & \R{spo.film.bta.ceil.jury} / \R{spo.film.bta.ceil.argon} \\
    Wildlife (\R{spo.film.wildlife.pairs}, \R{spo.film.wildlife.n}) & \R{spo.film.wildlife.se.jury} / \R{spo.film.wildlife.se.argon} & \R{spo.film.wildlife.vmem.jury} / \R{spo.film.wildlife.vmem.argon} & \R{spo.film.wildlife.vimax.jury} / \R{spo.film.wildlife.vimax.argon} & \R{spo.film.wildlife.cc.jury} / \R{spo.film.wildlife.cc.argon} & \R{spo.film.wildlife.cont.jury} / \R{spo.film.wildlife.cont.argon} & \R{spo.film.wildlife.ceil.jury} / \R{spo.film.wildlife.ceil.argon} \\
    Detail of a Life (\R{spo.film.detail.pairs}, \R{spo.film.detail.n}) & \R{spo.film.detail.se.jury} / \R{spo.film.detail.se.argon} & \R{spo.film.detail.vmem.jury} / \R{spo.film.detail.vmem.argon} & \R{spo.film.detail.vimax.jury} / \R{spo.film.detail.vimax.argon} & \R{spo.film.detail.cc.jury} / \R{spo.film.detail.cc.argon} & \R{spo.film.detail.cont.jury} / \R{spo.film.detail.cont.argon} & \R{spo.film.detail.ceil.jury} / \R{spo.film.detail.ceil.argon} \\
    Smalltime$^\dagger$ (\R{spo.film.smalltime.pairs}, \R{spo.film.smalltime.n}) & \R{spo.film.smalltime.se.jury} / \R{spo.film.smalltime.se.argon} & \R{spo.film.smalltime.vmem.jury} / \R{spo.film.smalltime.vmem.argon} & \R{spo.film.smalltime.vimax.jury} / \R{spo.film.smalltime.vimax.argon} & \R{spo.film.smalltime.cc.jury} / \R{spo.film.smalltime.cc.argon} & \R{spo.film.smalltime.cont.jury} / \R{spo.film.smalltime.cont.argon} & \R{spo.film.smalltime.ceil.jury} / \R{spo.film.smalltime.ceil.argon} \\
    \bottomrule
  \end{tabular}
\end{table*}

\begin{table*}[t]
  \centering
  \caption{\spe per screenplay (\R{spe.n.all} questions), jury / Argon. $n$: scored questions of the screenplay. \method{} scores highest on \R{spe.first.jury} of the nine screenplays under the jury and \R{spe.first.argon} under Argon.}
  \label{tab:supp_perscript}
  \scriptsize
  \setlength{\tabcolsep}{3pt}
  \begin{tabular}{@{}lcccccc@{}}
    \toprule
    Screenplay (pairs, $n$) & StoryEngine & VideoMemory & ViMax & CineCrew & \method & \textit{\Ceiling} \\
    \midrule
    Step Back, Doors Closing (4, \R{spe.film.stepback.n}) & \R{spe.film.stepback.se.jury} / \R{spe.film.stepback.se.argon} & \R{spe.film.stepback.vmem.jury} / \R{spe.film.stepback.vmem.argon} & \R{spe.film.stepback.vimax.jury} / \R{spe.film.stepback.vimax.argon} & \R{spe.film.stepback.cc.jury} / \R{spe.film.stepback.cc.argon} & \R{spe.film.stepback.cont.jury} / \R{spe.film.stepback.cont.argon} & \R{spe.film.stepback.ceil.jury} / \R{spe.film.stepback.ceil.argon} \\
    The Civil Dead (4, \R{spe.film.civildead.n}) & \R{spe.film.civildead.se.jury} / \R{spe.film.civildead.se.argon} & \R{spe.film.civildead.vmem.jury} / \R{spe.film.civildead.vmem.argon} & \R{spe.film.civildead.vimax.jury} / \R{spe.film.civildead.vimax.argon} & \R{spe.film.civildead.cc.jury} / \R{spe.film.civildead.cc.argon} & \R{spe.film.civildead.cont.jury} / \R{spe.film.civildead.cont.argon} & \R{spe.film.civildead.ceil.jury} / \R{spe.film.civildead.ceil.argon} \\
    Jockey (4, \R{spe.film.jockey.n}) & \R{spe.film.jockey.se.jury} / \R{spe.film.jockey.se.argon} & \R{spe.film.jockey.vmem.jury} / \R{spe.film.jockey.vmem.argon} & \R{spe.film.jockey.vimax.jury} / \R{spe.film.jockey.vimax.argon} & \R{spe.film.jockey.cc.jury} / \R{spe.film.jockey.cc.argon} & \R{spe.film.jockey.cont.jury} / \R{spe.film.jockey.cont.argon} & \R{spe.film.jockey.ceil.jury} / \R{spe.film.jockey.ceil.argon} \\
    Fast Color (4, \R{spe.film.fastcolor.n}) & \R{spe.film.fastcolor.se.jury} / \R{spe.film.fastcolor.se.argon} & \R{spe.film.fastcolor.vmem.jury} / \R{spe.film.fastcolor.vmem.argon} & \R{spe.film.fastcolor.vimax.jury} / \R{spe.film.fastcolor.vimax.argon} & \R{spe.film.fastcolor.cc.jury} / \R{spe.film.fastcolor.cc.argon} & \R{spe.film.fastcolor.cont.jury} / \R{spe.film.fastcolor.cont.argon} & \R{spe.film.fastcolor.ceil.jury} / \R{spe.film.fastcolor.ceil.argon} \\
    Kin (4, \R{spe.film.kin.n}) & \R{spe.film.kin.se.jury} / \R{spe.film.kin.se.argon} & \R{spe.film.kin.vmem.jury} / \R{spe.film.kin.vmem.argon} & \R{spe.film.kin.vimax.jury} / \R{spe.film.kin.vimax.argon} & \R{spe.film.kin.cc.jury} / \R{spe.film.kin.cc.argon} & \R{spe.film.kin.cont.jury} / \R{spe.film.kin.cont.argon} & \R{spe.film.kin.ceil.jury} / \R{spe.film.kin.ceil.argon} \\
    Big Time Adolescence (4, \R{spe.film.bta.n}) & \R{spe.film.bta.se.jury} / \R{spe.film.bta.se.argon} & \R{spe.film.bta.vmem.jury} / \R{spe.film.bta.vmem.argon} & \R{spe.film.bta.vimax.jury} / \R{spe.film.bta.vimax.argon} & \R{spe.film.bta.cc.jury} / \R{spe.film.bta.cc.argon} & \R{spe.film.bta.cont.jury} / \R{spe.film.bta.cont.argon} & \R{spe.film.bta.ceil.jury} / \R{spe.film.bta.ceil.argon} \\
    Wildlife (4, \R{spe.film.wildlife.n}) & \R{spe.film.wildlife.se.jury} / \R{spe.film.wildlife.se.argon} & \R{spe.film.wildlife.vmem.jury} / \R{spe.film.wildlife.vmem.argon} & \R{spe.film.wildlife.vimax.jury} / \R{spe.film.wildlife.vimax.argon} & \R{spe.film.wildlife.cc.jury} / \R{spe.film.wildlife.cc.argon} & \R{spe.film.wildlife.cont.jury} / \R{spe.film.wildlife.cont.argon} & \R{spe.film.wildlife.ceil.jury} / \R{spe.film.wildlife.ceil.argon} \\
    Detail of a Life (3, \R{spe.film.detail.n}) & \R{spe.film.detail.se.jury} / \R{spe.film.detail.se.argon} & \R{spe.film.detail.vmem.jury} / \R{spe.film.detail.vmem.argon} & \R{spe.film.detail.vimax.jury} / \R{spe.film.detail.vimax.argon} & \R{spe.film.detail.cc.jury} / \R{spe.film.detail.cc.argon} & \R{spe.film.detail.cont.jury} / \R{spe.film.detail.cont.argon} & \R{spe.film.detail.ceil.jury} / \R{spe.film.detail.ceil.argon} \\
    Smalltime (2, \R{spe.film.smalltime.n}) & \R{spe.film.smalltime.se.jury} / \R{spe.film.smalltime.se.argon} & \R{spe.film.smalltime.vmem.jury} / \R{spe.film.smalltime.vmem.argon} & \R{spe.film.smalltime.vimax.jury} / \R{spe.film.smalltime.vimax.argon} & \R{spe.film.smalltime.cc.jury} / \R{spe.film.smalltime.cc.argon} & \R{spe.film.smalltime.cont.jury} / \R{spe.film.smalltime.cont.argon} & \R{spe.film.smalltime.ceil.jury} / \R{spe.film.smalltime.ceil.argon} \\
    \bottomrule
  \end{tabular}
\end{table*}

\begin{table}[t]
  \centering
  \caption{\Planaudit checks, mean over the nine screenplays (a missing plan scores 0). Presence, relationships kept and causal links carried are computed by code; script events shown come from Argon.}
  \label{tab:supp_checks}
  \footnotesize
  \setlength{\tabcolsep}{3pt}
  \begin{tabular}{@{}lcccc@{}}
    \toprule
    System & Presence & Relations & Causal links & Events \\
    \midrule
    StoryEngine & \R{aud.pres.se} & \R{aud.rel.se} & \R{aud.caus.se} & \R{aud.events.se} \\
    CineCrew    & \R{aud.pres.cc} & \R{aud.rel.cc} & \R{aud.caus.cc} & \R{aud.events.cc} \\
    VideoMemory & \R{aud.pres.vmem} & \R{aud.rel.vmem} & \R{aud.caus.vmem} & \R{aud.events.vmem} \\
    ViMax       & \R{aud.pres.vimax} & \R{aud.rel.vimax} & \R{aud.caus.vimax} & \R{aud.events.vimax} \\
    \method     & \R{aud.pres.cont} & \R{aud.rel.cont} & \R{aud.caus.cont} & \R{aud.events.cont} \\
    \bottomrule
  \end{tabular}
\end{table}

\begin{table*}[t]
  \centering
  \caption{Whole-plan reading by Argon, 1--10; \method: mean of three readings per screenplay; baselines: one reading. Top: per criterion, mean over the nine screenplays (the whole-plan reading). Bottom: mean of the nine criteria per screenplay.}
  \label{tab:supp_reading}
  \scriptsize
  \setlength{\tabcolsep}{3pt}
  \begin{tabular}{@{}lcccccccccc@{}}
    \toprule
    System & Relations & Intensity & Colour & Causality & Verification & Fidelity & Coverage & Filmability & Consistency & Mean \\
    \midrule
    StoryEngine & \R{aud.crit.rel.se} & \R{aud.crit.int.se} & \R{aud.crit.col.se} & \R{aud.crit.caus.se} & \R{aud.crit.ver.se} & \R{aud.crit.fid.se} & \R{aud.crit.cov.se} & \R{aud.crit.film.se} & \R{aud.crit.cons.se} & \R{aud.read.se} \\
    CineCrew & \R{aud.crit.rel.cc} & \R{aud.crit.int.cc} & \R{aud.crit.col.cc} & \R{aud.crit.caus.cc} & \R{aud.crit.ver.cc} & \R{aud.crit.fid.cc} & \R{aud.crit.cov.cc} & \R{aud.crit.film.cc} & \R{aud.crit.cons.cc} & \R{aud.read.cc} \\
    VideoMemory & \R{aud.crit.rel.vmem} & \R{aud.crit.int.vmem} & \R{aud.crit.col.vmem} & \R{aud.crit.caus.vmem} & \R{aud.crit.ver.vmem} & \R{aud.crit.fid.vmem} & \R{aud.crit.cov.vmem} & \R{aud.crit.film.vmem} & \R{aud.crit.cons.vmem} & \R{aud.read.vmem} \\
    ViMax & \R{aud.crit.rel.vimax} & \R{aud.crit.int.vimax} & \R{aud.crit.col.vimax} & \R{aud.crit.caus.vimax} & \R{aud.crit.ver.vimax} & \R{aud.crit.fid.vimax} & \R{aud.crit.cov.vimax} & \R{aud.crit.film.vimax} & \R{aud.crit.cons.vimax} & \R{aud.read.vimax} \\
    \method & \R{aud.crit.rel.cont} & \R{aud.crit.int.cont} & \R{aud.crit.col.cont} & \R{aud.crit.caus.cont} & \R{aud.crit.ver.cont} & \R{aud.crit.fid.cont} & \R{aud.crit.cov.cont} & \R{aud.crit.film.cont} & \R{aud.crit.cons.cont} & \R{aud.read.cont} \\
    \bottomrule
  \end{tabular}

  \vspace{4pt}
  \begin{tabular}{@{}lccccc@{}}
    \toprule
    Screenplay & StoryEngine & VideoMemory & ViMax & CineCrew & \method \\
    \midrule
    Step Back, Doors Closing & \R{aud.film.stepback.read.se} & \R{aud.film.stepback.read.vmem} & \R{aud.film.stepback.read.vimax} & \R{aud.film.stepback.read.cc} & \R{aud.film.stepback.read.cont} \\
    The Civil Dead & \R{aud.film.civildead.read.se} & \R{aud.film.civildead.read.vmem} & \R{aud.film.civildead.read.vimax} & \R{aud.film.civildead.read.cc} & \R{aud.film.civildead.read.cont} \\
    Jockey & \R{aud.film.jockey.read.se} & \R{aud.film.jockey.read.vmem} & \R{aud.film.jockey.read.vimax} & \R{aud.film.jockey.read.cc} & \R{aud.film.jockey.read.cont} \\
    Fast Color & \R{aud.film.fastcolor.read.se} & \R{aud.film.fastcolor.read.vmem} & \R{aud.film.fastcolor.read.vimax} & \R{aud.film.fastcolor.read.cc} & \R{aud.film.fastcolor.read.cont} \\
    Kin & \R{aud.film.kin.read.se} & \R{aud.film.kin.read.vmem} & \R{aud.film.kin.read.vimax} & \R{aud.film.kin.read.cc} & \R{aud.film.kin.read.cont} \\
    Big Time Adolescence & \R{aud.film.bta.read.se} & \R{aud.film.bta.read.vmem} & \R{aud.film.bta.read.vimax} & \R{aud.film.bta.read.cc} & \R{aud.film.bta.read.cont} \\
    Wildlife & \R{aud.film.wildlife.read.se} & \R{aud.film.wildlife.read.vmem} & \R{aud.film.wildlife.read.vimax} & \R{aud.film.wildlife.read.cc} & \R{aud.film.wildlife.read.cont} \\
    Detail of a Life & \R{aud.film.detail.read.se} & \R{aud.film.detail.read.vmem} & \R{aud.film.detail.read.vimax} & \R{aud.film.detail.read.cc} & \R{aud.film.detail.read.cont} \\
    Smalltime & \R{aud.film.smalltime.read.se} & \R{aud.film.smalltime.read.vmem} & \R{aud.film.smalltime.read.vimax} & \R{aud.film.smalltime.read.cc} & \R{aud.film.smalltime.read.cont} \\
    \bottomrule
  \end{tabular}
\end{table*}

\paragraph{By gap, presence and development screenplays.} \Cref{tab:supp_gapbins} splits \speobj by the gap of the object pair, \cref{tab:supp_gates} gives how often each system's requests pass both presence gates and how accurate the identity answers are on those objects, and \cref{tab:supp_nodev} repeats the main results without the two development screenplays (\cref{sec:supp_disc}).

\begin{table*}[t]
  \centering
  \caption{\speobj by the gap of the object pair, jury / Argon. Column heads give the scored questions in each bin; Lead: \method minus the best baseline; same-object items, points, with the 95\% film interval (bootstrap over the nine films, pairs resampled within each drawn film).}
  \label{tab:supp_gapbins}
  \scriptsize
  \setlength{\tabcolsep}{3pt}
  \begin{tabular}{@{}lcccccc@{}}
    \toprule
    & \multicolumn{2}{c}{2--5 scenes (\R{spo.gap.2to5.pairs} pairs)} & \multicolumn{2}{c}{6--15 scenes (\R{spo.gap.6to15.pairs} pairs)} & \multicolumn{2}{c}{16--39 scenes (\R{spo.gap.16to39.pairs} pairs)} \\
    \cmidrule(lr){2-3}\cmidrule(lr){4-5}\cmidrule(l){6-7}
    System & Same (\R{spo.gap.2to5.n.same}) & All (\R{spo.gap.2to5.n.all}) & Same (\R{spo.gap.6to15.n.same}) & All (\R{spo.gap.6to15.n.all}) & Same (\R{spo.gap.16to39.n.same}) & All (\R{spo.gap.16to39.n.all}) \\
    \midrule
    StoryEngine & \R{spo.gap.2to5.same.se.jury} / \R{spo.gap.2to5.same.se.argon} & \R{spo.gap.2to5.all.se.jury} / \R{spo.gap.2to5.all.se.argon} & \R{spo.gap.6to15.same.se.jury} / \R{spo.gap.6to15.same.se.argon} & \R{spo.gap.6to15.all.se.jury} / \R{spo.gap.6to15.all.se.argon} & \R{spo.gap.16to39.same.se.jury} / \R{spo.gap.16to39.same.se.argon} & \R{spo.gap.16to39.all.se.jury} / \R{spo.gap.16to39.all.se.argon} \\
    VideoMemory & \R{spo.gap.2to5.same.vmem.jury} / \R{spo.gap.2to5.same.vmem.argon} & \R{spo.gap.2to5.all.vmem.jury} / \R{spo.gap.2to5.all.vmem.argon} & \R{spo.gap.6to15.same.vmem.jury} / \R{spo.gap.6to15.same.vmem.argon} & \R{spo.gap.6to15.all.vmem.jury} / \R{spo.gap.6to15.all.vmem.argon} & \R{spo.gap.16to39.same.vmem.jury} / \R{spo.gap.16to39.same.vmem.argon} & \R{spo.gap.16to39.all.vmem.jury} / \R{spo.gap.16to39.all.vmem.argon} \\
    ViMax & \R{spo.gap.2to5.same.vimax.jury} / \R{spo.gap.2to5.same.vimax.argon} & \R{spo.gap.2to5.all.vimax.jury} / \R{spo.gap.2to5.all.vimax.argon} & \R{spo.gap.6to15.same.vimax.jury} / \R{spo.gap.6to15.same.vimax.argon} & \R{spo.gap.6to15.all.vimax.jury} / \R{spo.gap.6to15.all.vimax.argon} & \R{spo.gap.16to39.same.vimax.jury} / \R{spo.gap.16to39.same.vimax.argon} & \R{spo.gap.16to39.all.vimax.jury} / \R{spo.gap.16to39.all.vimax.argon} \\
    CineCrew & \R{spo.gap.2to5.same.cc.jury} / \R{spo.gap.2to5.same.cc.argon} & \R{spo.gap.2to5.all.cc.jury} / \R{spo.gap.2to5.all.cc.argon} & \R{spo.gap.6to15.same.cc.jury} / \R{spo.gap.6to15.same.cc.argon} & \R{spo.gap.6to15.all.cc.jury} / \R{spo.gap.6to15.all.cc.argon} & \R{spo.gap.16to39.same.cc.jury} / \R{spo.gap.16to39.same.cc.argon} & \R{spo.gap.16to39.all.cc.jury} / \R{spo.gap.16to39.all.cc.argon} \\
    \method & \R{spo.gap.2to5.same.cont.jury} / \R{spo.gap.2to5.same.cont.argon} & \R{spo.gap.2to5.all.cont.jury} / \R{spo.gap.2to5.all.cont.argon} & \R{spo.gap.6to15.same.cont.jury} / \R{spo.gap.6to15.same.cont.argon} & \R{spo.gap.6to15.all.cont.jury} / \R{spo.gap.6to15.all.cont.argon} & \R{spo.gap.16to39.same.cont.jury} / \R{spo.gap.16to39.same.cont.argon} & \R{spo.gap.16to39.all.cont.jury} / \R{spo.gap.16to39.all.cont.argon} \\
    \midrule
    \textit{\Ceiling} & \R{spo.gap.2to5.same.ceil.jury} / \R{spo.gap.2to5.same.ceil.argon} & \R{spo.gap.2to5.all.ceil.jury} / \R{spo.gap.2to5.all.ceil.argon} & \R{spo.gap.6to15.same.ceil.jury} / \R{spo.gap.6to15.same.ceil.argon} & \R{spo.gap.6to15.all.ceil.jury} / \R{spo.gap.6to15.all.ceil.argon} & \R{spo.gap.16to39.same.ceil.jury} / \R{spo.gap.16to39.same.ceil.argon} & \R{spo.gap.16to39.all.ceil.jury} / \R{spo.gap.16to39.all.ceil.argon} \\
    \midrule
    Lead, Same, jury & \multicolumn{2}{c}{\R{spo.gap.2to5.lead.same.jury} [\R{spo.gap.2to5.lead.same.jury.cl.lo}, \R{spo.gap.2to5.lead.same.jury.cl.hi}]} & \multicolumn{2}{c}{\R{spo.gap.6to15.lead.same.jury} [\R{spo.gap.6to15.lead.same.jury.cl.lo}, \R{spo.gap.6to15.lead.same.jury.cl.hi}]} & \multicolumn{2}{c}{\R{spo.gap.16to39.lead.same.jury} [\R{spo.gap.16to39.lead.same.jury.cl.lo}, \R{spo.gap.16to39.lead.same.jury.cl.hi}]} \\
    Lead, Same, Argon & \multicolumn{2}{c}{\R{spo.gap.2to5.lead.same.argon} [\R{spo.gap.2to5.lead.same.argon.cl.lo}, \R{spo.gap.2to5.lead.same.argon.cl.hi}]} & \multicolumn{2}{c}{\R{spo.gap.6to15.lead.same.argon} [\R{spo.gap.6to15.lead.same.argon.cl.lo}, \R{spo.gap.6to15.lead.same.argon.cl.hi}]} & \multicolumn{2}{c}{\R{spo.gap.16to39.lead.same.argon} [\R{spo.gap.16to39.lead.same.argon.cl.lo}, \R{spo.gap.16to39.lead.same.argon.cl.hi}]} \\
    \bottomrule
  \end{tabular}
\end{table*}

\begin{table*}[t]
  \centering
  \caption{\speobj presence gates per system, so readers can see how much of each score is presence. Gates passed: share of identity objects whose two presence gates the judge answers yes. Accuracy on passed: identity accuracy on those objects. Jury / Argon.}
  \label{tab:supp_gates}
  \scriptsize
  \setlength{\tabcolsep}{3pt}
  \begin{tabular}{@{}lcccccc@{}}
    \toprule
    & \multicolumn{3}{c}{Gates passed} & \multicolumn{3}{c}{Accuracy on passed} \\
    \cmidrule(lr){2-4}\cmidrule(l){5-7}
    System & Same & Different & Identity & Same & Different & Identity \\
    \midrule
    StoryEngine & \R{spo.gate.same.se.jury} / \R{spo.gate.same.se.argon} & \R{spo.gate.diff.se.jury} / \R{spo.gate.diff.se.argon} & \R{spo.gate.ident.se.jury} / \R{spo.gate.ident.se.argon} & \R{spo.gateacc.same.se.jury} / \R{spo.gateacc.same.se.argon} & \R{spo.gateacc.diff.se.jury} / \R{spo.gateacc.diff.se.argon} & \R{spo.gateacc.ident.se.jury} / \R{spo.gateacc.ident.se.argon} \\
    VideoMemory & \R{spo.gate.same.vmem.jury} / \R{spo.gate.same.vmem.argon} & \R{spo.gate.diff.vmem.jury} / \R{spo.gate.diff.vmem.argon} & \R{spo.gate.ident.vmem.jury} / \R{spo.gate.ident.vmem.argon} & \R{spo.gateacc.same.vmem.jury} / \R{spo.gateacc.same.vmem.argon} & \R{spo.gateacc.diff.vmem.jury} / \R{spo.gateacc.diff.vmem.argon} & \R{spo.gateacc.ident.vmem.jury} / \R{spo.gateacc.ident.vmem.argon} \\
    ViMax & \R{spo.gate.same.vimax.jury} / \R{spo.gate.same.vimax.argon} & \R{spo.gate.diff.vimax.jury} / \R{spo.gate.diff.vimax.argon} & \R{spo.gate.ident.vimax.jury} / \R{spo.gate.ident.vimax.argon} & \R{spo.gateacc.same.vimax.jury} / \R{spo.gateacc.same.vimax.argon} & \R{spo.gateacc.diff.vimax.jury} / \R{spo.gateacc.diff.vimax.argon} & \R{spo.gateacc.ident.vimax.jury} / \R{spo.gateacc.ident.vimax.argon} \\
    CineCrew & \R{spo.gate.same.cc.jury} / \R{spo.gate.same.cc.argon} & \R{spo.gate.diff.cc.jury} / \R{spo.gate.diff.cc.argon} & \R{spo.gate.ident.cc.jury} / \R{spo.gate.ident.cc.argon} & \R{spo.gateacc.same.cc.jury} / \R{spo.gateacc.same.cc.argon} & \R{spo.gateacc.diff.cc.jury} / \R{spo.gateacc.diff.cc.argon} & \R{spo.gateacc.ident.cc.jury} / \R{spo.gateacc.ident.cc.argon} \\
    \method & \R{spo.gate.same.cont.jury} / \R{spo.gate.same.cont.argon} & \R{spo.gate.diff.cont.jury} / \R{spo.gate.diff.cont.argon} & \R{spo.gate.ident.cont.jury} / \R{spo.gate.ident.cont.argon} & \R{spo.gateacc.same.cont.jury} / \R{spo.gateacc.same.cont.argon} & \R{spo.gateacc.diff.cont.jury} / \R{spo.gateacc.diff.cont.argon} & \R{spo.gateacc.ident.cont.jury} / \R{spo.gateacc.ident.cont.argon} \\
    \midrule
    \textit{\Ceiling} & \R{spo.gate.same.ceil.jury} / \R{spo.gate.same.ceil.argon} & \R{spo.gate.diff.ceil.jury} / \R{spo.gate.diff.ceil.argon} & \R{spo.gate.ident.ceil.jury} / \R{spo.gate.ident.ceil.argon} & \R{spo.gateacc.same.ceil.jury} / \R{spo.gateacc.same.ceil.argon} & \R{spo.gateacc.diff.ceil.jury} / \R{spo.gateacc.diff.ceil.argon} & \R{spo.gateacc.ident.ceil.jury} / \R{spo.gateacc.ident.ceil.argon} \\
    \bottomrule
  \end{tabular}
\end{table*}

\begin{table*}[t]
  \centering
  \caption{Results without the two development screenplays (\emph{Jockey}, \emph{Smalltime}): seven screenplays, \R{spo.nodev.pairs} scene pairs on \speobj and \R{spe.nodev.pairs} pairs on \spe. Cells jury / Argon; column heads give scored questions. Lead: \method minus the best baseline; points, with the 95\% film interval (films resampled, then pairs within each drawn film).}
  \label{tab:supp_nodev}
  \scriptsize
  \setlength{\tabcolsep}{3pt}
  \begin{tabular}{@{}lccccc@{}}
    \toprule
    \speobj & Same (\R{spo.nodev.n.same}) & Different (\R{spo.nodev.n.diff}) & Identity (\R{spo.nodev.n.ident}) & Cause (\R{spo.nodev.n.cause}) & All (\R{spo.nodev.n.all}) \\
    \midrule
    StoryEngine & \R{spo.nodev.same.se.jury} / \R{spo.nodev.same.se.argon} & \R{spo.nodev.diff.se.jury} / \R{spo.nodev.diff.se.argon} & \R{spo.nodev.ident.se.jury} / \R{spo.nodev.ident.se.argon} & \R{spo.nodev.cause.se.jury} / \R{spo.nodev.cause.se.argon} & \R{spo.nodev.all.se.jury} / \R{spo.nodev.all.se.argon} \\
    VideoMemory & \R{spo.nodev.same.vmem.jury} / \R{spo.nodev.same.vmem.argon} & \R{spo.nodev.diff.vmem.jury} / \R{spo.nodev.diff.vmem.argon} & \R{spo.nodev.ident.vmem.jury} / \R{spo.nodev.ident.vmem.argon} & \R{spo.nodev.cause.vmem.jury} / \R{spo.nodev.cause.vmem.argon} & \R{spo.nodev.all.vmem.jury} / \R{spo.nodev.all.vmem.argon} \\
    ViMax & \R{spo.nodev.same.vimax.jury} / \R{spo.nodev.same.vimax.argon} & \R{spo.nodev.diff.vimax.jury} / \R{spo.nodev.diff.vimax.argon} & \R{spo.nodev.ident.vimax.jury} / \R{spo.nodev.ident.vimax.argon} & \R{spo.nodev.cause.vimax.jury} / \R{spo.nodev.cause.vimax.argon} & \R{spo.nodev.all.vimax.jury} / \R{spo.nodev.all.vimax.argon} \\
    CineCrew & \R{spo.nodev.same.cc.jury} / \R{spo.nodev.same.cc.argon} & \R{spo.nodev.diff.cc.jury} / \R{spo.nodev.diff.cc.argon} & \R{spo.nodev.ident.cc.jury} / \R{spo.nodev.ident.cc.argon} & \R{spo.nodev.cause.cc.jury} / \R{spo.nodev.cause.cc.argon} & \R{spo.nodev.all.cc.jury} / \R{spo.nodev.all.cc.argon} \\
    \method & \R{spo.nodev.same.cont.jury} / \R{spo.nodev.same.cont.argon} & \R{spo.nodev.diff.cont.jury} / \R{spo.nodev.diff.cont.argon} & \R{spo.nodev.ident.cont.jury} / \R{spo.nodev.ident.cont.argon} & \R{spo.nodev.cause.cont.jury} / \R{spo.nodev.cause.cont.argon} & \R{spo.nodev.all.cont.jury} / \R{spo.nodev.all.cont.argon} \\
    \midrule
    Lead, jury & \R{spo.nodev.lead.same.jury} [\R{spo.nodev.lead.same.jury.cl.lo}, \R{spo.nodev.lead.same.jury.cl.hi}] & \R{spo.nodev.lead.diff.jury} [\R{spo.nodev.lead.diff.jury.cl.lo}, \R{spo.nodev.lead.diff.jury.cl.hi}] & \R{spo.nodev.lead.ident.jury} [\R{spo.nodev.lead.ident.jury.cl.lo}, \R{spo.nodev.lead.ident.jury.cl.hi}] & \R{spo.nodev.lead.cause.jury} [\R{spo.nodev.lead.cause.jury.cl.lo}, \R{spo.nodev.lead.cause.jury.cl.hi}] & \R{spo.nodev.lead.all.jury} [\R{spo.nodev.lead.all.jury.cl.lo}, \R{spo.nodev.lead.all.jury.cl.hi}] \\
    Lead, Argon & \R{spo.nodev.lead.same.argon} [\R{spo.nodev.lead.same.argon.cl.lo}, \R{spo.nodev.lead.same.argon.cl.hi}] & \R{spo.nodev.lead.diff.argon} [\R{spo.nodev.lead.diff.argon.cl.lo}, \R{spo.nodev.lead.diff.argon.cl.hi}] & \R{spo.nodev.lead.ident.argon} [\R{spo.nodev.lead.ident.argon.cl.lo}, \R{spo.nodev.lead.ident.argon.cl.hi}] & \R{spo.nodev.lead.cause.argon} [\R{spo.nodev.lead.cause.argon.cl.lo}, \R{spo.nodev.lead.cause.argon.cl.hi}] & \R{spo.nodev.lead.all.argon} [\R{spo.nodev.lead.all.argon.cl.lo}, \R{spo.nodev.lead.all.argon.cl.hi}] \\
    \midrule
    \spe & Character & Emotion & Colour & Story state & All (\R{spe.nodev.n}) \\
    \midrule
    StoryEngine & \R{spe.nodev.char.se.jury} / \R{spe.nodev.char.se.argon} & \R{spe.nodev.emo.se.jury} / \R{spe.nodev.emo.se.argon} & \R{spe.nodev.col.se.jury} / \R{spe.nodev.col.se.argon} & \R{spe.nodev.state.se.jury} / \R{spe.nodev.state.se.argon} & \R{spe.nodev.all.se.jury} / \R{spe.nodev.all.se.argon} \\
    VideoMemory & \R{spe.nodev.char.vmem.jury} / \R{spe.nodev.char.vmem.argon} & \R{spe.nodev.emo.vmem.jury} / \R{spe.nodev.emo.vmem.argon} & \R{spe.nodev.col.vmem.jury} / \R{spe.nodev.col.vmem.argon} & \R{spe.nodev.state.vmem.jury} / \R{spe.nodev.state.vmem.argon} & \R{spe.nodev.all.vmem.jury} / \R{spe.nodev.all.vmem.argon} \\
    ViMax & \R{spe.nodev.char.vimax.jury} / \R{spe.nodev.char.vimax.argon} & \R{spe.nodev.emo.vimax.jury} / \R{spe.nodev.emo.vimax.argon} & \R{spe.nodev.col.vimax.jury} / \R{spe.nodev.col.vimax.argon} & \R{spe.nodev.state.vimax.jury} / \R{spe.nodev.state.vimax.argon} & \R{spe.nodev.all.vimax.jury} / \R{spe.nodev.all.vimax.argon} \\
    CineCrew & \R{spe.nodev.char.cc.jury} / \R{spe.nodev.char.cc.argon} & \R{spe.nodev.emo.cc.jury} / \R{spe.nodev.emo.cc.argon} & \R{spe.nodev.col.cc.jury} / \R{spe.nodev.col.cc.argon} & \R{spe.nodev.state.cc.jury} / \R{spe.nodev.state.cc.argon} & \R{spe.nodev.all.cc.jury} / \R{spe.nodev.all.cc.argon} \\
    \method & \R{spe.nodev.char.cont.jury} / \R{spe.nodev.char.cont.argon} & \R{spe.nodev.emo.cont.jury} / \R{spe.nodev.emo.cont.argon} & \R{spe.nodev.col.cont.jury} / \R{spe.nodev.col.cont.argon} & \R{spe.nodev.state.cont.jury} / \R{spe.nodev.state.cont.argon} & \R{spe.nodev.all.cont.jury} / \R{spe.nodev.all.cont.argon} \\
    \midrule
    Lead, jury & \R{spe.nodev.lead.char.jury} [\R{spe.nodev.lead.char.jury.cl.lo}, \R{spe.nodev.lead.char.jury.cl.hi}] & \R{spe.nodev.lead.emo.jury} [\R{spe.nodev.lead.emo.jury.cl.lo}, \R{spe.nodev.lead.emo.jury.cl.hi}] & \R{spe.nodev.lead.col.jury} [\R{spe.nodev.lead.col.jury.cl.lo}, \R{spe.nodev.lead.col.jury.cl.hi}] & \R{spe.nodev.lead.state.jury} [\R{spe.nodev.lead.state.jury.cl.lo}, \R{spe.nodev.lead.state.jury.cl.hi}] & \R{spe.nodev.lead.all.jury} [\R{spe.nodev.lead.all.jury.cl.lo}, \R{spe.nodev.lead.all.jury.cl.hi}] \\
    Lead, Argon & \R{spe.nodev.lead.char.argon} [\R{spe.nodev.lead.char.argon.cl.lo}, \R{spe.nodev.lead.char.argon.cl.hi}] & \R{spe.nodev.lead.emo.argon} [\R{spe.nodev.lead.emo.argon.cl.lo}, \R{spe.nodev.lead.emo.argon.cl.hi}] & \R{spe.nodev.lead.col.argon} [\R{spe.nodev.lead.col.argon.cl.lo}, \R{spe.nodev.lead.col.argon.cl.hi}] & \R{spe.nodev.lead.state.argon} [\R{spe.nodev.lead.state.argon.cl.lo}, \R{spe.nodev.lead.state.argon.cl.hi}] & \R{spe.nodev.lead.all.argon} [\R{spe.nodev.lead.all.argon.cl.lo}, \R{spe.nodev.lead.all.argon.cl.hi}] \\
    \bottomrule
  \end{tabular}
\end{table*}

\section{Judges}
\label{sec:supp_judges}

Every judge reads the same requests, stills and questions with the same prompt and seeds (\cref{sec:supp_eval}). \Cref{tab:supp_judges_obj,tab:supp_judges} give each judge's score; \cref{tab:supp_agree} gives how often two judges give the same answer. Opus 5.5 also fixes the truth of \speobj; \cref{tab:supp_noopus} gives the results without it.

\begin{table}[t]
  \centering
  \caption{\speobj per judge: All (\R{spo.n.all}, gated). Opus, Sonnet, Sol: single jurors; Jury: their majority.}
  \label{tab:supp_judges_obj}
  \footnotesize
  \setlength{\tabcolsep}{3pt}
  \begin{tabular}{@{}lccccc@{}}
    \toprule
    System & Opus & Sonnet & Sol & Jury & Argon \\
    \midrule
    StoryEngine & \R{spo.all.se.opus} & \R{spo.all.se.sonnet} & \R{spo.all.se.sol} & \R{spo.all.se.jury} & \R{spo.all.se.argon} \\
    VideoMemory & \R{spo.all.vmem.opus} & \R{spo.all.vmem.sonnet} & \R{spo.all.vmem.sol} & \R{spo.all.vmem.jury} & \R{spo.all.vmem.argon} \\
    ViMax       & \R{spo.all.vimax.opus} & \R{spo.all.vimax.sonnet} & \R{spo.all.vimax.sol} & \R{spo.all.vimax.jury} & \R{spo.all.vimax.argon} \\
    CineCrew    & \R{spo.all.cc.opus} & \R{spo.all.cc.sonnet} & \R{spo.all.cc.sol} & \R{spo.all.cc.jury} & \R{spo.all.cc.argon} \\
    \method     & \R{spo.all.cont.opus} & \R{spo.all.cont.sonnet} & \R{spo.all.cont.sol} & \R{spo.all.cont.jury} & \R{spo.all.cont.argon} \\
    \midrule
    \textit{\Ceiling} & \R{spo.all.ceil.opus} & \R{spo.all.ceil.sonnet} & \R{spo.all.ceil.sol} & \R{spo.all.ceil.jury} & \R{spo.all.ceil.argon} \\
    \bottomrule
  \end{tabular}
\end{table}

\begin{table}[t]
  \centering
  \caption{\spe per judge: All (\R{spe.n.all}). Opus, Sonnet, Sol: single jurors; Jury: their majority.}
  \label{tab:supp_judges}
  \footnotesize
  \setlength{\tabcolsep}{3pt}
  \begin{tabular}{@{}lccccc@{}}
    \toprule
    System & Opus & Sonnet & Sol & Jury & Argon \\
    \midrule
    StoryEngine & \R{spe.all.se.opus} & \R{spe.all.se.sonnet} & \R{spe.all.se.sol} & \R{spe.all.se.jury} & \R{spe.all.se.argon} \\
    VideoMemory & \R{spe.all.vmem.opus} & \R{spe.all.vmem.sonnet} & \R{spe.all.vmem.sol} & \R{spe.all.vmem.jury} & \R{spe.all.vmem.argon} \\
    ViMax       & \R{spe.all.vimax.opus} & \R{spe.all.vimax.sonnet} & \R{spe.all.vimax.sol} & \R{spe.all.vimax.jury} & \R{spe.all.vimax.argon} \\
    CineCrew    & \R{spe.all.cc.opus} & \R{spe.all.cc.sonnet} & \R{spe.all.cc.sol} & \R{spe.all.cc.jury} & \R{spe.all.cc.argon} \\
    \method     & \R{spe.all.cont.opus} & \R{spe.all.cont.sonnet} & \R{spe.all.cont.sol} & \R{spe.all.cont.jury} & \R{spe.all.cont.argon} \\
    \midrule
    \textit{\Ceiling} & \R{spe.all.ceil.opus} & \R{spe.all.ceil.sonnet} & \R{spe.all.ceil.sol} & \R{spe.all.ceil.jury} & \R{spe.all.ceil.argon} \\
    \bottomrule
  \end{tabular}
\end{table}

\begin{table}[t]
  \centering
  \caption{Answer agreement between judges over every system: share of questions with the same answer (\%) / Cohen's $\kappa$.}
  \label{tab:supp_agree}
  \footnotesize
  \setlength{\tabcolsep}{4pt}
  \begin{tabular}{@{}lcc@{}}
    \toprule
    Judges & \speobj & \spe \\
    \midrule
    Opus, Sonnet & \R{agree.spo.opus-sonnet.pct} / \R{agree.spo.opus-sonnet.kappa} & \R{agree.spe.opus-sonnet.pct} / \R{agree.spe.opus-sonnet.kappa} \\
    Opus, Sol & \R{agree.spo.opus-sol.pct} / \R{agree.spo.opus-sol.kappa} & \R{agree.spe.opus-sol.pct} / \R{agree.spe.opus-sol.kappa} \\
    Sonnet, Sol & \R{agree.spo.sonnet-sol.pct} / \R{agree.spo.sonnet-sol.kappa} & \R{agree.spe.sonnet-sol.pct} / \R{agree.spe.sonnet-sol.kappa} \\
    Opus, Argon & \R{agree.spo.opus-argon.pct} / \R{agree.spo.opus-argon.kappa} & \R{agree.spe.opus-argon.pct} / \R{agree.spe.opus-argon.kappa} \\
    Sonnet, Argon & \R{agree.spo.sonnet-argon.pct} / \R{agree.spo.sonnet-argon.kappa} & \R{agree.spe.sonnet-argon.pct} / \R{agree.spe.sonnet-argon.kappa} \\
    Sol, Argon & \R{agree.spo.sol-argon.pct} / \R{agree.spo.sol-argon.kappa} & \R{agree.spe.sol-argon.pct} / \R{agree.spe.sol-argon.kappa} \\
    \midrule
    Jury, Argon & \R{agree.spo.pct} / \R{agree.spo.kappa} & \R{agree.spe.pct} / \R{agree.spe.kappa} \\
    \bottomrule
  \end{tabular}
\end{table}

\paragraph{Agreement.} Among the five compared systems (\method and the four baselines), \R{agree.spo.rank} of the four single judges rank \method first on \speobj, and their system rankings agree with a mean Kendall $\tau$ of \R{agree.spo.tau}; on \spe the counts are \R{agree.spe.rank} of four and \R{agree.spe.tau}.

\paragraph{Jury without Opus.} Claude Opus 5.5 also fixes the truth of \speobj; \cref{tab:supp_noopus} gives the results without it, under the majority of Claude Sonnet 5.5, GPT-6.1 Sol and Gemini 4 Argon per answer (no majority counts as wrong). Its answers agree with the jury's on \R{agree.spo.noopus.pct}\% of \speobj answers and \R{agree.spe.noopus.pct}\% of \spe answers.

\begin{table*}[t]
  \centering
  \caption{Results under the majority of Sonnet, Sol and Argon (no Opus). Lead: \method minus the best baseline; points, with the 95\% film interval (films resampled, then pairs within each drawn film).}
  \label{tab:supp_noopus}
  \scriptsize
  \setlength{\tabcolsep}{3pt}
  \begin{tabular}{@{}lccccc@{}}
    \toprule
    \speobj & Same (\R{spo.n.same}) & Different (\R{spo.n.diff}) & Identity (\R{spo.n.ident}) & Cause (\R{spo.n.cause}) & All (\R{spo.n.all}) \\
    \midrule
    StoryEngine & \R{spo.same.se.noopus} & \R{spo.diff.se.noopus} & \R{spo.ident.se.noopus} & \R{spo.cause.se.noopus} & \R{spo.all.se.noopus} \\
    VideoMemory & \R{spo.same.vmem.noopus} & \R{spo.diff.vmem.noopus} & \R{spo.ident.vmem.noopus} & \R{spo.cause.vmem.noopus} & \R{spo.all.vmem.noopus} \\
    ViMax & \R{spo.same.vimax.noopus} & \R{spo.diff.vimax.noopus} & \R{spo.ident.vimax.noopus} & \R{spo.cause.vimax.noopus} & \R{spo.all.vimax.noopus} \\
    CineCrew & \R{spo.same.cc.noopus} & \R{spo.diff.cc.noopus} & \R{spo.ident.cc.noopus} & \R{spo.cause.cc.noopus} & \R{spo.all.cc.noopus} \\
    \method & \R{spo.same.cont.noopus} & \R{spo.diff.cont.noopus} & \R{spo.ident.cont.noopus} & \R{spo.cause.cont.noopus} & \R{spo.all.cont.noopus} \\
    \midrule
    Lead & \R{spo.lead.same.noopus} [\R{spo.lead.same.noopus.cl.lo}, \R{spo.lead.same.noopus.cl.hi}] & \R{spo.lead.diff.noopus} [\R{spo.lead.diff.noopus.cl.lo}, \R{spo.lead.diff.noopus.cl.hi}] & \R{spo.lead.ident.noopus} [\R{spo.lead.ident.noopus.cl.lo}, \R{spo.lead.ident.noopus.cl.hi}] & \R{spo.lead.cause.noopus} [\R{spo.lead.cause.noopus.cl.lo}, \R{spo.lead.cause.noopus.cl.hi}] & \R{spo.lead.all.noopus} [\R{spo.lead.all.noopus.cl.lo}, \R{spo.lead.all.noopus.cl.hi}] \\
    \midrule
    \spe & Character (\R{spe.n.char}) & Emotion (\R{spe.n.emo}) & Colour (\R{spe.n.col}) & Story state (\R{spe.n.state}) & All (\R{spe.n.all}) \\
    \midrule
    StoryEngine & \R{spe.char.se.noopus} & \R{spe.emo.se.noopus} & \R{spe.col.se.noopus} & \R{spe.state.se.noopus} & \R{spe.all.se.noopus} \\
    VideoMemory & \R{spe.char.vmem.noopus} & \R{spe.emo.vmem.noopus} & \R{spe.col.vmem.noopus} & \R{spe.state.vmem.noopus} & \R{spe.all.vmem.noopus} \\
    ViMax & \R{spe.char.vimax.noopus} & \R{spe.emo.vimax.noopus} & \R{spe.col.vimax.noopus} & \R{spe.state.vimax.noopus} & \R{spe.all.vimax.noopus} \\
    CineCrew & \R{spe.char.cc.noopus} & \R{spe.emo.cc.noopus} & \R{spe.col.cc.noopus} & \R{spe.state.cc.noopus} & \R{spe.all.cc.noopus} \\
    \method & \R{spe.char.cont.noopus} & \R{spe.emo.cont.noopus} & \R{spe.col.cont.noopus} & \R{spe.state.cont.noopus} & \R{spe.all.cont.noopus} \\
    \midrule
    Lead & \R{spe.lead.char.noopus} [\R{spe.lead.char.noopus.cl.lo}, \R{spe.lead.char.noopus.cl.hi}] & \R{spe.lead.emo.noopus} [\R{spe.lead.emo.noopus.cl.lo}, \R{spe.lead.emo.noopus.cl.hi}] & \R{spe.lead.col.noopus} [\R{spe.lead.col.noopus.cl.lo}, \R{spe.lead.col.noopus.cl.hi}] & \R{spe.lead.state.noopus} [\R{spe.lead.state.noopus.cl.lo}, \R{spe.lead.state.noopus.cl.hi}] & \R{spe.lead.all.noopus} [\R{spe.lead.all.noopus.cl.lo}, \R{spe.lead.all.noopus.cl.hi}] \\
    \bottomrule
  \end{tabular}
\end{table*}

\section{Video-judge noise control}
\label{sec:supp_noise}

The main evaluation judges shot requests, not rendered video. This control asks whether an LLM judge answers known-answer questions consistently and correctly from video, against the same questions answered from the shot requests behind that video.

\paragraph{Design.} Ten items from rendered probe scenes of earlier plans on a separate frozen set of scene pairs, drawn by a seeded rule with no hand picks, at most two per system: five \emph{copies} (the same scene on both sides) and five \emph{cross-film decoys} (Scene 1 from one screenplay, Scene 2 from another, rendered by the same system, so look and renderer match while people and sets differ). Truths are known by construction: in a copy, ``the same person (place) in both scenes'' is yes and ``did the time or intensity change'' is no; in a decoy, ``the same person (place)'' is no. In a copy, each Scene 1 question is also asked of Scene 2 (a mirror), and the two answers must match. Each item is asked ten times from the video and ten times from the requests, with fresh calls, the default temperature and one fixed presentation (question order, option order, labels). The judge is Gemini 4 Argon, with the same settings in both media; in the decoys' request text, the second screenplay's cast tokens are renamed so that no token collision says ``same person''. Items, questions and truths were fixed before any call. The renders come from earlier plans; the control measures the judge, not a planner.

\paragraph{Metrics.} \emph{Answer changes}: the share of question-repeat answers that differ from that question's most common answer. \emph{Mirror match}: the share of mirror pairs with equal answers. \emph{Accuracy}: the share of answers equal to the known truth, pooled over the questions whose truth is known by construction and those whose truth comes from the script sheet (the latter mix judge error with render and plan misses). \emph{Chance}: the accuracy of a judge that picks uniformly among each question's options, over the same questions.

\begin{table}[t]
  \centering
  \caption{Video-judge noise control: ten known-answer items, ten repeats each, Argon. Chance: \R{noise.chance}.}
  \label{tab:noise}
  \footnotesize
  \setlength{\tabcolsep}{4pt}
  \begin{tabular}{@{}lccc@{}}
    \toprule
    Input & Answer changes & Mirror match & Accuracy \\
    \midrule
    Video    & \R{noise.flip.video} & \R{noise.mirror.video} & \R{noise.acc.video} \\
    Requests & \R{noise.flip.text} & \R{noise.mirror.text} & \R{noise.acc.text} \\
    \bottomrule
  \end{tabular}
\end{table}

\paragraph{Results.} \Cref{tab:noise} gives answer changes, mirror matches and accuracy for video and for requests, on \R{noise.items} items. On these \R{noise.items} items, with one judge and renders of earlier plans, answers from video and from requests change across repeats at similar rates (\R{noise.flip.video} against \R{noise.flip.text}); video answers match their mirrors on \R{noise.mirror.video} against \R{noise.mirror.text} and are right on \R{noise.acc.video} against \R{noise.acc.text} (chance \R{noise.chance}). This small control measures the judge's stability on known answers. It does not show that request-space scores predict scores on rendered video, which we have not measured. We judge requests because they separate planning from rendering failures and need no video.

\section{Example pairs}
\label{sec:supp_qual}

\paragraph{The pair of \cref{fig:teaser}.}
The pair was picked by a rule fixed before any pair's stills, requests or answers were looked at.
From the \speobj pairs of the seven screenplays not used in development, we kept the same-object pairs on which the jury marks \method right and each of the four baselines wrong or unable to tell (\exKept{} of the \R{spo.nodev.pairs.same} test pairs whose objects are all the same; the seven test screenplays hold \R{spo.nodev.n.same} same-object items, as a pair can hold two objects, and \R{spo.nodev.pairs.mixed} of them holds a same and a different one), ordered them by gap, longest first, and took the first whose object is visible in \method's reference stills of both scenes: \emph{\exFilm}, scenes \exE{} and \exL, the first in the order.
The second judge agrees on every system.
The figure is a best case of \method against the baselines.
The figure shows the two baselines with the highest \speobj score, every still attached to the two scenes' requests with the car's first, and, per scene, the first request sentence that names the car by its most specific word.

\begin{figure*}[t]
  \centering
  \input{fig/style_v3}
\begin{tikzpicture}[fig, x=1cm, y=1cm]
  \path[use as bounding box] (0,0) rectangle (17.4,5.25);
  \def\qh{2.8}                                    %
  \def\qtop{4.74}                                 %
  \node[anchor=base west, inner sep=0pt] at (0,5.02) {\figlabelb (a) Character\ \ \textit{Big Time Adolescence}\ \ \fignote\color{cInk2}the principal, 94 scenes apart};
  \figimgcrop[anchor=north west, at={(0,\qtop)}]{qaE}{2.1cm}{\qh cm}{height=\qh cm}{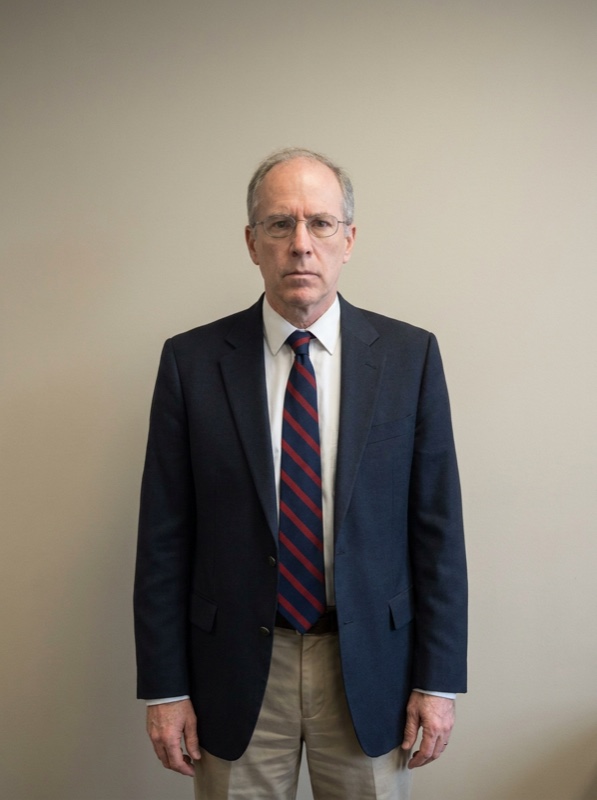}
  \figimgcrop[anchor=north west, at={(2.18,\qtop)}]{qaL}{2.1cm}{\qh cm}{height=\qh cm}{fig/frames/qual/g4_a.jpg}
  \node[tab, anchor=north west] at ([xshift=2pt,yshift=-2pt]qaE.north west) {scene 2};
  \node[tab, anchor=north west] at ([xshift=2pt,yshift=-2pt]qaL.north west) {scene 96};
  \draw[hl, rounded corners=1.5pt] ([xshift=.3cm,yshift=-1pt]qaE.south west) -- ++(0,-.22) -| ([xshift=-.3cm,yshift=-1pt]qaL.south east);
  \node[fact, anchor=center] at ([yshift=-.255cm]$(qaE.south west)!.5!(qaL.south east)$) {principal\_wool\ {\fignote\color{cLedgerInk}one character}};
  \node[anchor=base west, inner sep=0pt] at (0,1.27) {\fignote\color{cInk2}scene 96 reuses his still from scene 2};
  \node[fact, anchor=base west] at (0,1.0) {principal\_wool.faceDescription};
  \node[anchor=north west, inner sep=0pt, text width=4.28cm, align=flush left] at (0,.77)
    {\figquote ``Thinning grey hair, tight lips, rimless spectacles''\ {\fignote\color{cInk2}the plan's cast entry}};
  \node[anchor=base west, inner sep=0pt] at (4.56,4.5) {\figlabelb Frozen question};
  \node[anchor=north west, inner sep=0pt, text width=3.3cm, align=flush left] at (4.56,4.36)
    {\figquote ``Is Principal the same person in both scenes, judging only face and build (not clothes or light)?''};
  \node[anchor=base west, inner sep=0pt] at (4.56,2.98) {\figlabelb Answers\ \ \fignote\color{cInk2}jury and Argon agree};
  \draw[cRule, line width=.3pt] (4.56,2.86) -- (7.86,2.86);
  \newcommand{\qa}[5]{%
    \node[anchor=base west, inner sep=0pt] at (4.56,{2.56-(#1)*.52}) {\fignote #2\ \ #3};
    \node[anchor=base east, inner sep=0pt] at (7.86,{2.56-(#1)*.52}) {\figquote #4};
    \node[anchor=base west, inner sep=0pt] at (4.89,{2.32-(#1)*.52}) {\fignote\color{cInk2}#5};}
  \qa{0}{\figyes}{{\bfseries\method}}{yes}{one character, one still}
  \qa{1}{\figno}{StoryEngine}{no}{two characters, two stills}
  \qa{2}{\figunk}{CineCrew}{can't tell}{no still of him}
  \qa{3}{\figunk}{ViMax}{can't tell}{no still of him}
  \qa{4}{\figunk}{VideoMemory}{can't tell}{no still of him}
  \draw[cRule, line width=.3pt] (8.13,0) -- (8.13,5.25);
  \def\bx{8.4}
  \def\bw{2.6}                                    %
  \node[anchor=base west, inner sep=0pt] at (\bx,5.02) {\figlabelb (b) Cause and effect\ \ \textit{Fast Color}\ \ \fignote\color{cInk2}a gas canister, 25 scenes apart};
  \figimgcrop[anchor=north west, at={(\bx,\qtop)}]{qbE}{\bw cm}{\qh cm}{height=\qh cm}{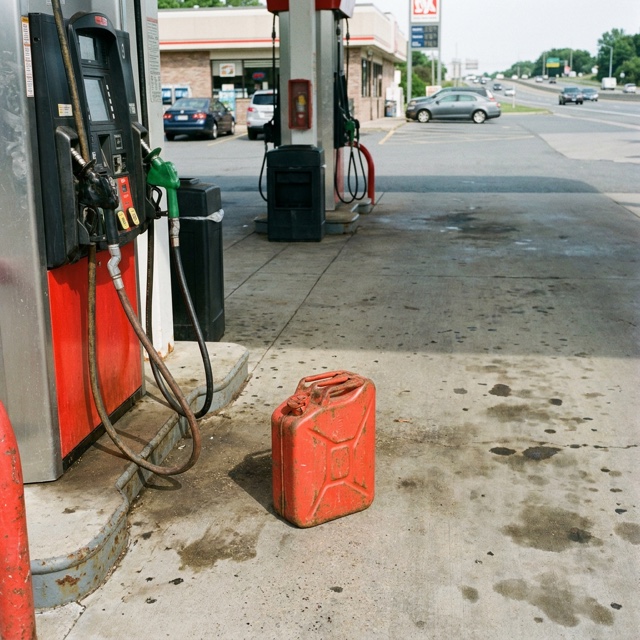}
  \figimgcrop[anchor=north west, at={(\bx+\bw+.08,\qtop)}]{qbL}{\bw cm}{\qh cm}{height=\qh cm}{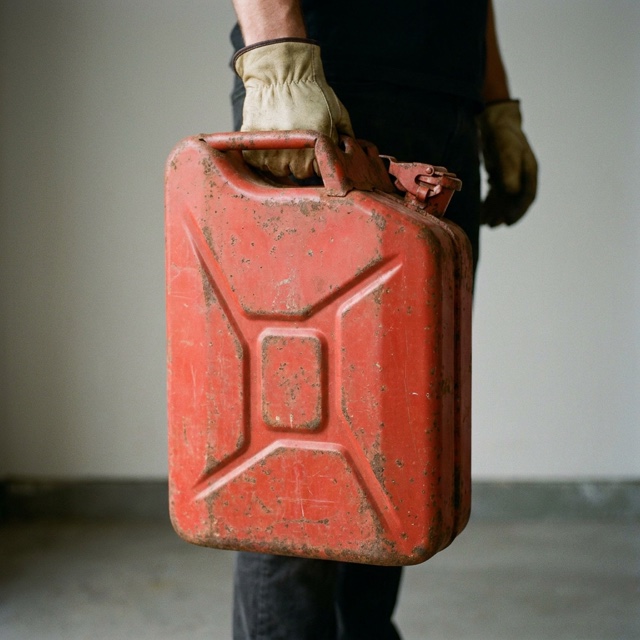}
  \node[tab, anchor=north west] at ([xshift=2pt,yshift=-2pt]qbE.north west) {scene 152: cause};
  \node[tab, anchor=north west] at ([xshift=2pt,yshift=-2pt]qbL.north west) {scene 177: effect};
  \fighl{qbE}{.36}{.54}{.65}{.86}
  \node[fact, anchor=north, align=left] (fE) at ([yshift=-3pt]qbE.south) {gas\_canister.at =\\[-1pt]highway\_gas\_station};
  \node[fact, anchor=north, align=left] (fL) at ([yshift=-3pt]qbL.south) {gas\_canister.heldBy\\[-1pt]= ruth};
  \draw[ledger io] (fE.east) -- (fL.west);
  \node[anchor=north west, inner sep=0pt, text width=\bw cm, align=flush left] at (\bx,1.2)
    {\figquote ``\dots\ drops the red metal canister down \figcarry{onto the concrete curb}.''};
  \node[anchor=north west, inner sep=0pt, text width=\bw cm, align=flush left] at (\bx+\bw+.08,1.2)
    {\figquote ``\dots\ the red metal gas canister \figcarry{off the concrete curb where she left it}.''};
  \def\cx{13.96}
  \node[anchor=base west, inner sep=0pt] at (\cx,4.5) {\figlabelb Frozen questions};
  \node[anchor=north west, inner sep=0pt, text width=3.44cm, align=flush left] at (\cx,4.36)
    {{\fignote\color{cInk2}Does scene 152 show this:}\\{\figquote ``Ruth put down the canister and ran when her hand shook.''}\\[2.5pt]
     {\fignote\color{cInk2}Does scene 177 show this:}\\{\figquote ``She retrieves the canister from where she left it.''}};
  \node[anchor=base, inner sep=0pt] at (\cx+1.915,2.0) {\fignote cause};
  \node[anchor=base, inner sep=0pt] at (\cx+2.94,2.0) {\fignote effect};
  \draw[cRule, line width=.3pt] (\cx+1.5,1.93) -- (\cx+2.42,1.93) (\cx+2.5,1.93) -- (\cx+3.44,1.93);
  \foreach \gx/\gt in {1.66/jury,2.17/Argon,2.69/jury,3.19/Argon}
    \node[anchor=base, inner sep=0pt] at (\cx+\gx,1.77) {\sffamily\fontsize{6}{6}\selectfont\color{cInk2}\gt};
  \draw[cRule, line width=.3pt] (\cx,1.67) -- (\cx+3.44,1.67);
  \newcommand{\qb}[6]{%
    \node[anchor=base west, inner sep=0pt] at (\cx,{1.42-(#1)*.29}) {\fignote #2};
    \node[anchor=base, inner sep=0pt] at (\cx+1.66,{1.42-(#1)*.29}) {\fignote #3};
    \node[anchor=base, inner sep=0pt] at (\cx+2.17,{1.42-(#1)*.29}) {\fignote #4};
    \node[anchor=base, inner sep=0pt] at (\cx+2.69,{1.42-(#1)*.29}) {\fignote #5};
    \node[anchor=base, inner sep=0pt] at (\cx+3.19,{1.42-(#1)*.29}) {\fignote #6};}
  \qb{0}{{\bfseries\method}}{\figyes}{\figyes}{\figyes}{\figyes}
  \qb{1}{StoryEngine}{\figyes}{\figyes}{\figunk}{\figyes}
  \qb{2}{CineCrew}{\figno}{\figno}{\figunk}{\figyes}
  \qb{3}{ViMax}{\figyes}{\figyes}{\figunk}{\figunk}
  \qb{4}{VideoMemory}{\figunk}{\figyes}{\figunk}{\figno}
\end{tikzpicture}%
  \caption{\textbf{A character and a cause, \method's reference stills.} (a)~A same-person question of \spe: \method's plan keeps the principal as one character, and its requests in scenes 2 and 96 attach the same still of him. (b)~A cause--effect pair of \speobj: the plan records where the canister is left in scene 152 (amber box) and that Ruth holds it in scene 177; under each still, the request sentence that places it. Frames are the reference stills that \method's final-engine requests attach in each scene, the images the judges were given; no video is rendered here. Marks are each system's recorded answers, under the jury and under Argon (\figyes{} right, \figno{} wrong, \figunk{} can't tell, scored wrong). Best cases by a fixed rule (\cref{sec:supp_qual}), not a sample; in (b) the jury marks every baseline wrong on the effect, while some baselines answer the cause right and Argon marks StoryEngine and CineCrew right on the effect.}
  \label{fig:qual}
\end{figure*}

\paragraph{The examples of \cref{fig:qual}.}
\Cref{fig:qual} shows two best cases, with \method's reference stills only: (a)~a character, \emph{Big Time Adolescence}, scenes 2 and 96, asked by \spe ``Is Principal the same person in both scenes, judging only face and build (not clothes or light)?'' (truth yes); (b)~a cause and its effect, \emph{Fast Color}, scenes 152 and 177, the \speobj cause-effect pair on a gas canister (cause shown in the early scene, effect in the late scene).
Both rules were fixed before any still was looked at and use the seven test screenplays (the nine without \emph{Jockey} and \emph{Smalltime}).
(a): among the same-person questions, keep those the jury marks right for \method and wrong for every baseline (2 of 41); prefer a film not shown in \cref{fig:teaser}, then the longest gap, and take the first for which \method's requests attach a still of the character in both scenes.
(b): among the cause-effect pairs, keep those the jury marks right for \method on both questions and wrong for every baseline on at least one (1 of 7), and take the longest gap.
In (a) \method's plan has one cast entry for the principal and attaches the same still in both scenes; StoryEngine makes two entities with two different stills, and CineCrew, ViMax and VideoMemory attach no still of him.
In (b) not every baseline is wrong: under the jury every baseline is wrong on the effect, but StoryEngine and ViMax are right on the cause; under Argon StoryEngine and CineCrew are right on the effect.
These are best cases, not a sample.

\section{Disclosures}
\label{sec:supp_disc}
\begin{itemize}[leftmargin=*,itemsep=0pt,topsep=2pt]
  \item Every judge, plan reader and script-sheet reader is an LLM.
  \item One judge, Argon, is from the planners' provider; it is reported apart from the jury (Claude Opus 5.5, Claude Sonnet 5.5, GPT-6.1 Sol), which is from two other providers. Claude Opus 5.5 both helps build the truth (script sheets, \speobj readings) and sits on the jury; \cref{tab:supp_noopus} gives the results without it.
  \item Baselines ran through our adapters, plan-only, on a matched LLM, with some stages off (\cref{tab:supp_adapt}), one attempt each plus the rerun of calls lost to our faults; no number compares with their papers. The rerun was not planned in advance.
  \item \speobj was designed by a person, who verified every scored answer; its builders read no plan, and it was frozen before any final plan was judged.
  \item The \spe pairs and questions were frozen before any plan of the final engine was made. The causal dynamic is not scored.
  \item The screenplays are not held out. The final engine's options were checked on \emph{Smalltime}, and an earlier version was built with \emph{Jockey}; \emph{Fast Color} was run once before the full run only to check that the engine ran without errors, and nothing was tuned on it. Without \emph{Smalltime} and \emph{Jockey}, \method scores \R{spo.nodev.all.cont.jury} against \R{spo.nodev.all.best.jury} for the best baseline on \speobj, and \R{spe.nodev.all.cont.jury} against \R{spe.nodev.all.best.jury} on \spe (jury; \cref{tab:supp_nodev}).
  \item Judge calls lost in transit were resent; a refusal or unusable answer still used up its attempt.
  \item \R{run.openissues.cont} of \R{run.windows.cont} windows were committed with open issues; the plan lists them.
  \item The mean standard deviation of the three whole-plan readings of one \method plan is \R{aud.read.sd}; we read the nine-film mean. Baseline plans were read once.
  \item The noise-control items come from earlier plans, not from \method's final plans.
\end{itemize}

\end{document}